%% file: arxiv-main.tex
\PassOptionsToPackage{dvipsnames}{xcolor}
\documentclass[format=acmsmall, screen=true, nonacm]{acmart}


\usepackage{pifont}  

\usepackage{float}
\usepackage{footnotehyper}
\usepackage{graphicx}
\usepackage{subcaption}

\usepackage[main=english]{babel}
\usepackage[autostyle]{csquotes}
\usepackage{microtype}
\usepackage{enumitem}
\usepackage{lipsum}
\usepackage[most]{tcolorbox}
\definecolor{SoftYellow}{RGB}{241, 168, 5}
\definecolor{SoftBlue}{RGB}{75, 115, 185}
\definecolor{ForestGreen}{RGB}{70, 135, 95}

\tcbset{
    colback=blue!5!white,
    colframe=blue!50!black,
    boxrule=0.5pt,
    arc=2mm,
    left=2pt,
    right=2pt,
    top=2pt,
    bottom=2pt,
    fonttitle=\bfseries,
    title style={left color=blue!20!white, right color=blue!20!white}
}

\usepackage{tabularx}
\usepackage{booktabs}
\usepackage{multirow}
\usepackage{makecell}
\usepackage{longtable}

\makeatletter
\newcommand{\customlabel}[2]{%
 \@bsphack\begingroup
 \def\@currentlabel{#2}%
 \label{#1}%
 \endgroup\@esphack
}
\makeatother

\newcommand{\sectionlabel}[1]{\customlabel{#1}{Section \arabic{section}}}
\newcommand{\subsectionlabel}[1]{\customlabel{#1}{Section \arabic{section}.\arabic{subsection}}}

\newcommand{\lllm}{\texttt{LLLMs}}

\usepackage[size=tiny]{todonotes}
\newcommand{\ak}[1]{{\color{black}#1}} 
\newcommand{\se}[1]{\textcolor{black}{#1}} 


\title{LLLMs: A Data-Driven Survey of Evolving Research on Limitations of Large Language Models}

\author{Aida Kostikova}
\affiliation{%
  \institution{University of Bielefeld}
  \city{Bielefeld}
  \country{Germany}}
\email{aida.kostikova@uni-bielefeld.de}

\author{Zhipin Wang}
\affiliation{%
  \institution{University of Technology Nuremberg}
  \city{Nuremberg}
  \country{Germany}}
  
\author{Deidamea Bajri}
\affiliation{%
  \institution{University of Mannheim}
  \city{Mannheim}
  \country{Germany}}
  
\author{Ole Pütz}
\affiliation{%
  \institution{University of Bielefeld}
  \city{Bielefeld}
  \country{Germany}}

\author{Benjamin Paaßen}
\affiliation{%
  \institution{University of Bielefeld}
  \city{Bielefeld}
  \country{Germany}}

\author{Steffen Eger}
\affiliation{%
  \institution{University of Technology Nuremberg}
  \city{Nuremberg}
  \country{Germany}}

\renewcommand{\shortauthors}{Kostikova et al.}

\begin{document}

\begin{abstract}
    Large language model (LLM) research has grown rapidly, along with increasing concern about their limitations. 
    In this survey, we conduct a data-driven, semi-automated review of research on limitations of LLMs (\lllm) from 2022 to \ak{early 2025} using a bottom-up approach. From a corpus of 250,000 ACL and arXiv papers, we identify 14,648 relevant papers using keyword filtering, LLM-based classification, validated against expert labels, and topic clustering (via two approaches, HDBSCAN+BERTopic and LlooM). We find that the \ak{share of LLM-related papers} increases over fivefold in ACL and nearly \ak{eightfold} in arXiv between 2022 and \ak{2025}. Since 2022, \lllm\ research grows even faster, reaching over 30\% of LLM papers by \ak{2025}. \textit{Reasoning} remains the most studied limitation, followed by \textit{generalization}, \textit{hallucination}, \textit{bias}, and \textit{security}. The distribution of topics in the ACL dataset stays relatively stable over time, while arXiv shifts toward 
    \textit{security risks}, \textit{alignment}, \textit{hallucinations}, \textit{knowledge editing}, and \textit{multimodality}. We offer a quantitative view of trends in \lllm\ research and release a dataset of annotated abstracts and a validated methodology, available at: \href{https://github.com/a-kostikova/LLLMs-Survey}{github.com/a-kostikova/LLLMs-Survey}.
\end{abstract}

\maketitle

\input{structure/1-introduction}
\input{structure/2-related-work}
\input{structure/3-0-methodology}

\input{structure/4-1-results}
\input{structure/4-2-results-clustering}

\input{structure/4-3-results-comparison}
\input{structure/5-discussion}

\begin{acks}
This research was funded by the Ministry of Culture and Science of the State of North Rhine-Westphalia under the grant no NW21-059A (SAIL). This project originated at the 2024 retreat of the Natural Language Learning \& Generation (NLLG) Lab at TU Nürnberg. We thank Yanran Chen, Christoph Leiter, Ran Zhang, Daniil Larionov, and Jonas Belouadi for valuable early input and discussions that helped shape this work.
\end{acks}

\bibliographystyle{ACM-Reference-Format}
\bibliography{bibliography} 

\input{structure/appendix} 

\end{document}

%% file: structure/1-introduction.tex
\section{Introduction}\sectionlabel{sec:introduction}

With the explosive growth of large language model (LLM) research and deployment \cite{wei2022emergent}, questions of the limitations of LLMs (\lllm) have also gathered increased interest, ranging from reasoning failures \cite{park2024picturing}, social bias \cite{lin2022gendered}, hallucinations \cite{xu2024hallucination}, difficulty in handling long contexts \cite{li2023compressing}, and many more. Understanding where LLMs fail is essential for knowing how and whether they can be safely and effectively used in real-world settings, especially as LLMs are increasingly deployed in safety-sensitive domains such as healthcare, education, finance, and law \citep{chen2024survey}. Moreover, tracking how these failure modes evolve over time helps reveal whether the fast-paced research landscape is addressing them, overlooking them, or exposing new ones, offering a clearer picture of where further research is most needed.

However, given the sheer size of LLM research, with thousands of published research papers every year (even when limited to highly rated outlets), it is challenging to maintain an up-to-date overview of \lllm\ research using traditional, manual literature review techniques. Accordingly, prior reviews on \lllm\ mostly focus on specific limitations, such as reasoning \cite{li2025system,xu2025towards}, or examine limitations within the broader context of evaluating overall model capabilities \citep{srivastava2022beyond, chang2024survey}. To date, the field 
\se{still misses}
an overview that covers the more recent LLM research 
\se{between 2022 and now} 
and cuts across limitations. \se{Our} review 
is an attempt to provide this high-level overview.

To make our task feasible, we opt for a data-driven, bottom-up approach and build a partially automated, systematic literature review pipeline. Starting from an initial set of almost 250,000 crawled papers from ACL (2022-2024) and arXiv (2022 through early 2025), we extract 14,648 papers that discuss \lllm\ (filtering for keywords first, then classifying the papers' abstracts with an LLM, validated against human expert classifications). Finally, we cluster the papers using two different methods (HDBSCAN+BERTopic and LlooM) to understand which particular limitations are researched. These approaches offer complementary strengths: the former provides single-label, density-based 
clustering, while the latter uses multi-label, LLM-based assignments, allowing us to cross-validate and reduce method-specific bias.
Overall, our methods serve to apply quantitative methods to surveying this vast field.

We observe four main results. i) \lllm\ research has grown rapidly, outpacing even the 
growth of LLM research overall. \ak{The share of LLM-related papers has grown by a factor of over five in ACL and nearly eight in arXiv between 2022-2025}, reaching almost 80\% of all crawled ACL papers and roughly 30\% of all crawled arXiv papers; \lllm\ papers have increased even more sharply, by a factor of over 12 in ACL and 28 in arXiv, accounting for more than 30\% of LLM papers in Q1 of 2025. ii) Within \lllm\ research, \textit{reasoning} limitations are the most prominent, with \textit{generalization}, \textit{hallucination}, \textit{bias}, and \textit{security} as further important concerns. iii) The distribution of limitations appears relatively stable in the ACL dataset, whereas the arXiv dataset shows a rise in concern for topics related to safety and controllability (e.g., \textit{Security Risks}, \textit{Alignment Limitations}, \textit{Knowledge Editing}, \textit{Hallucination}) as well as \textit{Multimodality}. (iv) Despite substantial methodological differences between HDBSCAN and LlooM, we observe topical overlap in several of the biggest clusters (e.g., \textit{Reasoning}, \textit{Hallucination}, \textit{Security Risks}) across both approaches, with broadly similar trend patterns, suggesting that these findings are reliable.

The contributions of this review to the field are i) a large-scale dataset of paper abstracts, tagged with limitation information, for further research,\footnote{\href{https://github.com/a-kostikova/LLLMs-Survey}{\ak{github.com/a-kostikova/LLLMs-Survey}}} ii) an LLM-based paper annotation methodology, validated against human experts, iii) most importantly, quantitative insights into the evolution of \lllm\ research covering the entire period 2022-2024 \se{and early 2025}, providing the first comprehensive overview of \lllm\ research for this period.

%% file: structure/2-related-work.tex
\section{Related Work}

\subsection{Surveys of Large Language Models}

A growing number of surveys have aimed to synthesize the rapid progress 
\se{of LLMs}, 
covering their architectures, training paradigms, applications, and broader impact. Notably, \citet{zhao2023survey} 
\se{have} 
become a widely adopted reference in the field, offering a structured overview along four key dimensions: pre-training, adaptation, utilization, and evaluation. Other comprehensive works expand on this foundation by discussing emerging areas such as multimodal LLMs, robotics, and system efficiency \citep{naveed2023comprehensive,hadi2023large}, as well as reasoning and planning capabilities in large-scale models \citep{matarazzo2025survey}.

In parallel to cross-domain surveys, a number of studies have investigated how LLMs are being adopted and evaluated in specific fields. In the medical domain, surveys examine the effectiveness of LLMs in clinical summarization and diagnostic reasoning \citep{kim2025limitations, van2024adapted}, as well as challenges related to hallucination and factual consistency in medical question answering \citep{pham2024towards}. Other works explore the capabilities and limitations of LLMs in capturing cultural commonsense knowledge \citep{shen2024understanding} and scientific research processes \citep{luo2025llm4sr, eger2025transforming}. In recommendation systems, LLMs have been studied as both retrieval and generation engines \citep{wu2024survey}, while in information retrieval, surveys highlight their use in query expansion, passage ranking, and answer synthesis \citep{zhu2023large}. LLMs have also been applied to software engineering tasks such as code generation and bug fixing, with systematic reviews discussing both their potential and practical limitations \citep{hou2024large}. Beyond text-based applications, LLMs have been integrated with structured resources such as knowledge graphs \citep{pan2024unifying, jin2024large} and studied in the context of autonomous agent design \citep{wang2024survey}.

While these surveys offer valuable perspectives on LLM usage and challenges within specific domains, the literature remains fragmented with respect to how limitations are identified, categorized, and compared. Our work is motivated by the need for a more systematic and scalable approach to mapping research focused explicitly on the limitations of LLMs across domains and tasks.

\subsection{Surveys on Limitations of LLMs}

As 
\se{LLMs} 
are increasingly deployed in real-world applications, a growing body of research has emerged to examine their limitations from different capability-oriented perspectives. One prominent area of concern are hallucinations, where recent surveys investigate underlying causes and mitigation strategies in both text generation \citep{ji2023survey, huang2025survey, tonmoy2024comprehensive} and multimodal contexts \citep{liu2024survey, sahoo2024comprehensive, sahoo-etal-2024-comprehensive}. Another major focus is reasoning, with surveys analyzing the development of novel techniques\cite{plaat2024reasoning, huang2022towards} such as chain-of-thought prompting \citep{chen2025towards}, reinforced reasoning \citep{xu2025towards, li2025system}, and mathematical problem-solving \citep{ahn2024large}. In parallel, the trustworthiness and reliability of LLMs have been studied through the lens of fairness, transparency, and calibration \citep{sun2024trustllm}, while other works concentrate on security and privacy threats including adversarial vulnerabilities and data leakage \citep{yao2024survey, das2025security}. A related thread explores ethical risks in LLM-based agents, particularly concerning safety, misuse, and human interaction \citep{jiao2024navigating, kumar2024ethics, gan2024navigating}.

Despite these valuable contributions, existing reviews typically focus on specific capabilities or domains in isolation, often adopting distinct definitions, evaluation metrics, and analytical frameworks. As a result, the broader landscape of LLM limitations remains fragmented, making it difficult to compare findings or track emerging research trends. This underscores the need for a more systematic and scalable approach to identifying and organizing literature across different limitations of LLMs.

\subsection{LLMs as Analytical Tools for Scientific Literature}

We apply a partially automated pipeline, relying on LLMs to filter the papers included in our survey and providing embeddings for clustering. Such methods have to be applied with care to avoid being misled because of the very limitations this survey is supposed to study. In developing our methodology, we rely on a growing literature of LLMs being used as instruments for analyzing scientific literature \citep{eger2025transforming}. Several recent approaches employ LLMs for topic modeling, semantic clustering, and concept induction, enabling more interpretable organization of large-scale corpora \citep{zhang2023clusterllm, diaz2025k, lam2024concept, feng2024llmedgerefine, pattnaik2024improving}. This has led to a growing number of systematic reviews that use BERTopic and related methods to map research landscapes across areas such as generative AI, information assurance, LLM applications, and research impact evaluation \citep{gana2024leveraging, ding2024unraveling, gupta2024generative, arsalan2025mapping}.

In addition to these analytical techniques, other efforts focus on automating synthesis of results across papers. Systems such as SurveyX and AutoSurvey generate draft surveys from large paper collections \citep{liang2025surveyx, wang2024autosurvey}, while tools like LitLLM \citep{agarwal2024litllm} and PaSa \citep{he2025pasa} support LLM-based retrieval, summarization, and exploration of academic texts. 

\ak{Closer to our work, recent studies use automated methods to analyze research limitations at the paper or corpus level, primarily in support of peer review. \citet{al2025bagels} introduce datasets and methods for extracting and generating study limitations, while \citet{azher2024limtopic} apply LLM-assisted topic modeling to limitation sections. \citet{xu2025limitgen} evaluate whether LLMs can identify critical limitations in AI papers. In contrast, our work studies \lllm\ themselves as the research object.}

To enable a data-driven, bottom-up analysis of the vast field of \lllm\ research, including several ten thousand papers, we opt to apply some of the aforementioned automation techniques. However, given the limitations of LLMs, we aim to validate each step of our method, either by comparing to a human gold standard, or by comparing the outputs of different methods (hence, we use two clustering approaches). As such, we aim to be more conservative in our utilization of LLMs in literature research compared to 
\se{the} prior work \se{pointed out above}.

%% file: structure/3-0-methodology.tex
\section{Methodology}\label{sec:methodology}

\begin{figure}
    \centering
    \includegraphics[width=0.8\textwidth]{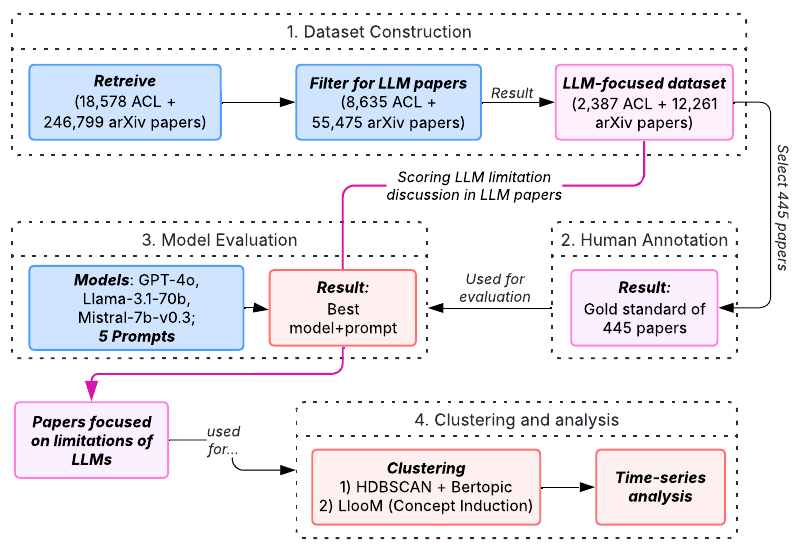}
    \caption{Overview of the pipeline for our systematic literature review.}
    \Description{A flowchart illustrating the pipeline for the systematic literature review. It starts with retrieving papers from ACL and arXiv, filtering them according to keywords, filtering them further based on an LLM classification of their abstracts, and clustering the papers. The LLM classification step is validated against a gold standard of 445 human-annotated papers and only the best-performing LLM in this validation is used. Clustering is performed with BERTopic and LlooM both to check the robustness of the results.}
    \label{fig:experimental_pipeline}
\end{figure}

Fig.~\ref{fig:experimental_pipeline} illustrates the method for our systematic literature review. We begin by retrieving papers from arXiv and ACL (Section~\ref{sec:data}), filter according to keywords (Section~\ref{sec:keywords}), filter papers further by classifying their abstracts with an LLM (Section~\ref{sec:llm-classification}), and finally cluster the papers (Section~\ref{sec:clustering}). At each step of the analysis, we perform validations to ensure the robustness of our results: the keyword list is obtained with an iterative refinement procedure, the LLM classification step is validated against a gold standard of 445 human-annotated papers, and we use two distinct clustering methods for comparison (HDBSCAN+BERTopic and LlooM), \ak{complemented by stratified topic-level human evaluation of cluster assignments.} In the remainder of this section, we describe each step in more detail.

\input{structure/3-1-data}

\input{structure/3-2-llm_classification}
\input{structure/3-3-clustering}

%% file: structure/3-1-data.tex
\subsection{Data Retrieval}\label{sec:data}

Our initial dataset includes academic papers published between January 2022 and March 2025, sourced from \ak{arXiv as a broad, multi-field research corpus, and from the ACL Anthology as a narrower, NLP-focused set.} ArXiv captures preprint research that closely tracks current developments and \ak{often appears months before formal publication, making it well suited for analyzing fast-moving research trends in LLMs}, while ACL venues reflect peer-reviewed work and remain the primary publication outlets for NLP research, where much of the foundational work on LLMs originated and \ak{where consistent multi-year coverage enables longitudinal analysis.\footnote{\ak{In the arXiv dataset, some papers later appear in ACL venues; we retain them to preserve arXiv's role as a source of fast-moving, venue-agnostic research.}}} The time frame was chosen to capture the year preceding the release of ChatGPT as well as all subsequent research on LLMs \citep{zhao2023survey}.

For ACL Anthology, we scrape conference pages for AACL 2022–2023, ACL 2022–2024, EACL 2023–2024, EMNLP 2022–2024, ICLR 2022–2024\footnote{\ak{Although ICLR is not an ACL venue nor part of the ACL Anthology, we include it in this group to represent a major peer-reviewed machine learning conference with substantial LLM research, and retain the \enquote{ACL} label for brevity.}}, NAACL 2022 and 2024, and TACL 2022–2024 \se{as the premier NLP venues}.\footnote{We exclude other tracks and venues such as workshops, system demonstrations, tutorials, shared tasks, and task-specific venues (e.g., SemEval, CoNLL, WMT) to maintain a focus on high-impact research from general-purpose NLP conferences.}
For arXiv, we retrieve papers from the categories of \se{Computation \& Language} 
(cs.CL), Machine Learning (cs.LG), Artificial Intelligence (cs.AI), and Computer Vision (cs.CV) because these communities are closest to LLM research. However, we note that many papers are classified by the authors into multiple arXiv categories, so that research areas beyond these initial ones are covered as well (see Section~\ref{sec:arxiv_categories_analysis} for more details). 
Each entry includes metadata such as title, publication date, author information, download link, and abstract, with arXiv papers also containing all assigned categories. We use titles and abstracts for keyword filtering and clustering, as they capture a paper's main claims and contributions and are \ak{consistently available across venues, making them well suited} for large-scale automated analysis. \ak{Including additional sections (e.g., introductions or related work) could introduce noise, increase preprocessing complexity due to heterogeneous document structures, and raise the cost of large-scale LLM-based annotation.}

\begin{figure}
    \centering
    \includegraphics[width=0.8\textwidth]{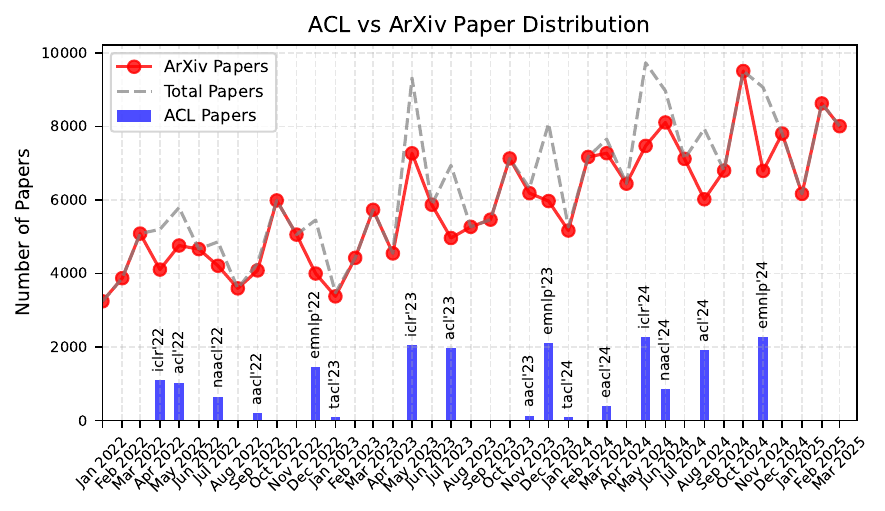}
    \caption{Distribution of papers over time in the crawled dataset, showing ACL papers, arXiv papers, and the total count (ACL + arXiv).} 
    \label{fig:paper_distribution}
\end{figure}

The final crawled dataset includes 245,835 papers (18,578 papers for ACL, 227,257 for arXiv). Fig.~\ref{fig:paper_distribution} shows raw numbers of crawled ACL and arXiv papers over time.

\subsection{Keyword-Based Filtering}
\label{sec:keywords}

In a first filtering stage, we exclude papers if no LLM-related keyword occurs in their title or abstract. This step serves to avoid excessive resource needs in the later, more fine-grained filtering.

To identify keywords related to \lllm\ research, we apply the following
iterative approach:
\begin{enumerate}
    \item We use TNT-KID \citep{martinc2022tnt} to generate initial keywords for each abstract. \ak{TNT-KID is a transformer-based keyword tagger that follows a two-stage training process: language model pretraining on large unlabeled text, followed by fine-tuning on labeled keyword data. 
    As reported by \citet{giarelis2024deep}, TNT-KID is among the top-performing keyphrase extraction methods and provides ready-to-use pretrained models. We use the off-the-shelf TNT-KID model fine-tuned on the KP20k dataset of scientific abstracts and titles.}
    \item Two manually reviewed sets of 50 LLM and 50 non-LLM papers are selected based on predefined seed terms (e.g., \emph{LLM}, \emph{large language model}).
    \item We compute log-likelihood ratio (LLR) scores for TNT-KID-generated keywords in both sets. Keywords with LLR $\geq 25$ are added to the list.
    \item Using the updated list, we expand the dataset by adding 100 more papers to each set, maintaining balance across venues and years while avoiding duplicates.
    \item Steps 3 and 4 repeat until all papers are processed. As more keywords are added, the rate of new informative terms naturally decreases. To avoid including increasingly marginal or noisy keywords in later iterations, we raise the LLR threshold by 5\% whenever fewer than 5\% of the keywords in the current round are new. This keeps the keyword list focused on strongly distinctive terms as the process converges.
\end{enumerate}

This results in a list of 90 keywords (19 unigrams, 44 bigrams, 16 trigrams, and 11 four-grams).
Overall, the final keyword set covers key aspects of LLM research (see the full list in Section~\ref{sec:keyword_list} supplementary material):

\begin{itemize}
    \item terms related to LLMs, including 
    \emph{multimodal LLMs}, \emph{small LMs} and \emph{pre-trained LMs};
    \item training methods: \emph{LoRA}, \emph{PEFT}, \emph{instruction tuning};
    \item capabilities (e.g. \emph{mathematical}, \emph{temporal}, and \emph{commonsense reasoning}); 
    \item limitations and risks, such as security vulnerabilities (\emph{jailbreaking}, \emph{prompt injection}, \emph{data contamination});
    \item model evaluation: \emph{self-evaluation}, \emph{benchmarking};
    \item methods and techniques, such as
    reasoning paradigms (\emph{chain of thought}, \emph{self-reflection}, \emph{tree of thoughts}), prompting strategies (\emph{prompt optimization}, \emph{prompt engineering}), augmentation methods (\emph{retrieval-augmented generation}, \emph{tool learning}).
\end{itemize}

We keep only those papers that contain at least one of the keywords in their title or abstract. This results in 64,110 papers (8,635 for ACL, 55,475 for arXiv). A breakdown of crawled and filtered papers across sources for each year is provided in Table~\ref{tab:crawled_vs_filtered}. While this filtering step does not strictly isolate LLM-focused papers, \ak{it serves as an initial inclusion step; its output is refined by subsequent LLM-based classification}. Even at this stage, we observe that the proportion of papers passing the filter increases over time, which may reflect a growing focus on LLMs in the broader NLP research landscape. We examine this pattern in more detail in \ref{sec:lllm-llm-growth}, where we analyze trends in both LLM and \lllm\ papers after additional filtering.

\begin{table}
    \centering
    \caption{Crawled vs. LLM-filtered paper counts across sources (2022–2025)}.
    \label{tab:crawled_vs_filtered}
    \renewcommand{\arraystretch}{0.95}
    \small
    \begin{tabular}{lrrrr}
        \toprule
        \textbf{Source / Year} & \textbf{2022} & \textbf{2023} & \textbf{2024} & \textbf{2025} \\
        \midrule
        \textit{ACL}   & 1,032 / 294     & 1,977 / 821     & 1,916 / 1,483   & -- / -- \\
        \textit{EACL}  & -- / --         & 478 / 188       & 382 / 204       & -- / -- \\
        \textit{AACL}  & 192 / 59        & 134 / 53        & -- / --         & -- / -- \\
        \textit{TACL}  & 84 / 27         & 98 / 32         & 95 / 60         & -- / -- \\
        \textit{EMNLP} & 1,372 / 520     & 2,107 / 1,177   & 2,273 / 1,844   & -- / -- \\
        \textit{ICLR}  & 1,094 / 143     & 1,573 / 251     & 2,260 / 674     & -- / -- \\
        \textit{NAACL} & 652 / 217       & -- / --         & 859 / 588       & -- / -- \\
        \midrule
        \midrule
        \textit{ArXiv} & 52,642 / 5,726  & 66,179 / 13,361 & 85,645 / 27,700 & 22,791 / 8,688 \\
        \midrule
        \textbf{Yearly Total} & \textbf{57,072 / 6,986} & \textbf{72,546 / 15,883} & \textbf{94,390 / 32,553} & \textbf{22,791 / 8,688} \\
        \textbf{Total (All Years)} & \multicolumn{4}{r}{\textbf{246,799 / 64,110}} \\
        \bottomrule
    \end{tabular}
    
    \vspace{0.5em}
    \small\textit{Note: Each cell is \texttt{Crawled / LLM-filtered}. ACL venues do not include data for 2025 (as of 31.03.2025).}
\end{table}

%% file: structure/3-2-llm_classification.tex
\subsection{LLM-Based Filtering}\label{sec:llm-classification}

In a second filtering stage, we apply an LLM to evaluate every abstract of the 64,110 papers left after the first filtering stage and 1) rate how much \lllm\ are discussed on a scale from 0 to 5, as well as 2) extract text snippets that explicitly discuss limitations for papers rated 2 or higher. The text snippets will later form the basis for clustering.

However, before we apply this filtering, we set up a human-annotated gold standard dataset to check if LLMs are able to perform this filtering in the first place.

\subsubsection{Human Annotation Task}
For human annotation, we randomly select 445 papers from the keyword-filtered dataset, balancing the source (ACL or arXiv, ensuring conference representation within ACL) and publication year. Papers are manually annotated based on their titles and abstracts to assess whether they discuss \lllm. The human annotators rated each paper on a scale from 0-5, reaching from no relation to LLMs (0) to exclusive focus on \lllm\ (5). Refer to Table~\ref{tab:annotation_labels} for the detailed annotation guideline. 

\begin{table}
    \caption{Annotation scheme for LLM limitation discussion, including label descriptions and distribution of papers in the human annotated dataset.}
    \label{tab:annotation_labels}
    \centering
    \renewcommand{\arraystretch}{0.95}
    \small
    \begin{tabular}{clr}
        \toprule
        \textbf{Label} & \textbf{Description} & \textbf{Count} \\
        \midrule
        0 & No mention of LLMs. & 62 \\
        1 & Mentions LLMs but not their limitations. & 106 \\
        2 & Briefly mentions a limitation, e.g., as justification for a new method. & 169 \\
        3 & Discusses one or two limitations in moderate detail but not as the primary focus. & 62 \\
        4 & Extensively discusses multiple limitations, making them a major focus. & 37 \\
        5 & Entirely focused on LLM limitations and challenges. & 9 \\
        \bottomrule
    \end{tabular}
\end{table}

For papers rated 2–5, annotators highlighted textual evidence pointing to the limitation \ak{and, where explicitly expressed in the text, its general type (i.e., the kind of limitation discussed)}, hereafter referred to as \enquote{evidence}. See Table~\ref{tab:annotated_examples} in the supplementary material for representative examples of annotated papers and highlighted evidence.

\paragraph{Limitation rating agreement} 

We measure annotator agreement using: 
\begin{enumerate}
    \item standard Cohen's Kappa for raw agreement;
    \item quadratic weighted Cohen's Kappa, which accounts for the ordinal nature of the 0–5 scale by penalizing larger discrepancies more heavily.
\end{enumerate}

\ak{In our agreement analysis, we include all papers annotated by at least two annotators (up to four). Agreement is computed pairwise for each annotator pair over the subset of papers they both annotated and averaged across all pairs.}

The annotation process included \ak{three} rounds, involving two professors (natural language processing and machine learning), one PhD student (NLP), and one Master's student (computer science), \ak{all of whom are fluent English speakers with at least C1 proficiency. Disagreements were addressed through discussion and refinement of the annotation guidelines between rounds, without per-instance adjudication.} Initial annotations showed moderate inter-annotator agreement (0.27 standard Cohen's Kappa, and 0.62 weighted), which was improved by further rounds (0.57, and 0.75, respectively), indicating substantial agreement in the final version. Overall, the annotation process covered 445 samples (where 195 were annotated solely by a Master's student after all discussion rounds).

\begin{table}
    \caption{Venue-year distribution of papers in the human-annotated dataset}
    \label{tab:venue_year_distribution}
    \centering
    \renewcommand{\arraystretch}{1.0}
    \setlength{\tabcolsep}{4pt}
    \small
    \begin{tabular}{lcccccccccc}
        \toprule
        \textbf{Year} & \textbf{arxiv} & \textbf{acl} & \textbf{aacl} & \textbf{eacl} & \textbf{emnlp} & \textbf{iclr} & \textbf{naacl} & \textbf{tacl} & \textbf{Total} \\
        \midrule
        2022 & 17 & 15 & 4 & 0 & 21 & 2 & 14 & 0 & 73 \\
        2023 & 69 & 22 & 0 & 4 & 39 & 0 & 0  & 82 & 216 \\
        2024 & 62 & 19 & 0 & 6 & 25 & 4 & 13 & 27 & 156 \\
        \midrule
        \textbf{Total} & 148 & 56 & 4 & 10 & 85 & 6 & 27 & 109 & \textbf{445} \\
        \bottomrule
    \end{tabular}
\end{table}

The final rating for each paper is determined by rounding the average of annotators' ratings. The number of papers assigned to each label in the human-annotated dataset is shown in Table~\ref{tab:annotation_labels}. Table~\ref{tab:venue_year_distribution} displays the statistics of labels across years.

\paragraph{Evidence annotation agreement} 
We compare 
the highlighted evidence using BIO-tagged sequences, where tokens are labeled as B-EVID, I-EVID, or O (beginning, inside or outside evidence, respectively). \ak{For each paper annotated by at least two annotators (250 of 455 abstracts), we compute pairwise inter-annotator F1 scores \ak{at the token level}. For each annotator pair and paper, we compare BIO-tagged token sequences, treating B-EVID and I-EVID as positive labels and O as the negative label, and compute F1, excluding annotator pairs for which neither annotator selected any evidence. We report macro-average F1 across all valid annotator pairs.}

On average, evidence agreement across the full jointly annotated dataset (ratings 0–5, 250 papers) is 0.55 in terms of averaged pairwise F1. For papers explicitly discussing limitations (papers with ratings 3–5 as a final label, 48 jointly annotated papers), F1 score increases to 0.71. This score suggests reliable consistency, given the known difficulty of span-level annotation \citep{deyoung2019eraser}.

\subsection{Models and Prompting Evaluation}\label{sec:models-prompting}

We evaluate models on the human-annotated dataset to identify the most effective model-prompt configuration for scoring LLM limitation discussions and extracting supporting evidence. Considering both performance and cost, we select the best-performing one for full-dataset classification and clustering of papers by limitation topics described in \ref{sec:results}.

To represent different families and sizes of models, we evaluated three models and selected the best performing one for full-scale annotation: \texttt{GPT-4o} selected as one of the best-performing models at the time of analysis \citep{chiang2024chatbotarena}, 
\texttt{Mistral-7B-Instruct-v0.3} \citep{jiang2023mistral7b} as a small-scale open-weight model and \texttt{Llama-3.1-70b-Instruct} \citep{grattafiori2024llama} as a large-scale open-weight model.\footnote{For \texttt{Mistral-7B-Instruct-v0.3} and \texttt{Llama-3.1-70b-Instruct}, we set the temperature to 0.6 and top\textunderscore{}k = 0.9; \texttt{Llama-3.1-70b-Instruct} is run with 4-bit quantization. GPT-4o is used in version \texttt{gpt-4o-2024-08-06}.} We also compare the results against a Logistic Regression baseline with SBERT embeddings \citep{reimers2019sentence}, using random sub-sampling validation with three 80/20 train–test splits, and applying SMOTE oversampling \citep{chawla2002smote} to mitigate class imbalance, with results averaged across splits. 

To account for the impact of prompting on model performance, we experiment with different strategies to determine the most effective approach:

\begin{itemize}
    \item \textbf{Prompt 1}: zero-shot baseline (no explicit rating rules).
    \item \textbf{Prompt 2}: zero-shot with defined rating criteria.
    \item \textbf{Prompt 3}: few-shot with defined rating criteria and five examples for each rating which also include explanation for each rating.
\end{itemize}

Similar to human annotators highlighting evidence in abstracts to indicate discussions of \lllm, models are prompted to extract supporting evidence from the text. Each prompt instructs the model: \textit{\enquote{Please respond in the following format, providing a rating and supporting evidence for the discussion of LLM limitations in each abstract. Do not include explanations, only cite the evidence found in the abstract.}} Prompt 3, included as the most comprehensive, is available in Figure~\ref{fig:evaluation_prompt} of the supplementary material.

\paragraph{Metrics and Evaluation}

We compare model ratings to human ratings using weighted Cohen's Kappa for rating prediction. To evaluate evidence extraction, we compare the BIO sequence of human annotators to the BIO sequence of models using averaged pairwise F1. Only papers rated 3–5 are included in the evidence evaluation.

%% file: structure/3-3-clustering.tex
\subsection{Clustering}\label{sec:clustering}

\begin{figure}
    \centering
    \includegraphics[width=0.9\textwidth]{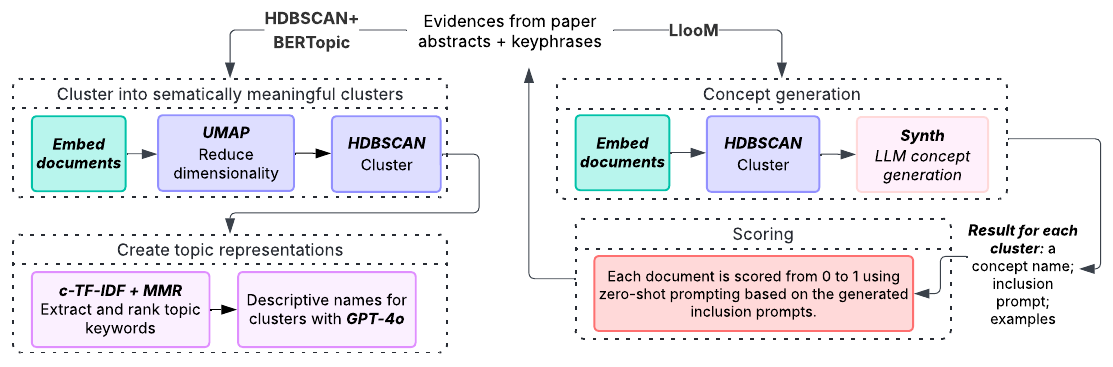}
    \caption{Comparison of clustering steps in HDBSCAN+BERTopic and LLooM. Both methods take on evidence excerpts with appended keyphrases as an input. For LLooM, we omit the \textit{Distill} step, which is typically used to summarize full documents, as our input already consists of concise excerpts.}
    \label{fig:hdbscan_lloom_clustering_steps}
\end{figure}

To identify patterns in \lllm\ discussions, we apply two clustering approaches: (i) HDBSCAN \citep{mcinnes2017hdbscan} + BERTopic \citep{grootendorst2022bertopic} and (ii) LLooM concept induction \citep{lam2024concept}, and compare their results. We select these algorithms in particular because they represent particularly distinct approaches to text clustering: HDBSCAN + BERTopic assigns each paper to at most one cluster and does so based on a hierarchical clustering in the embedding space. By contrast, LLooM derives topics first and then queries an LLM for each paper-topic-combination whether the respective paper belongs to that respective topic, thus permitting papers to belong to multiple clusters. We remain agnostic regarding the choice of clustering algorithm and focus on findings that are consistent across both approaches to enhance the robustness of our literature review. Below, we outline the data preparation process and describe each clustering pipeline in detail.

\paragraph{Data preparation}

As clustering material, we use not the full abstract but only the passages that explicitly describe the LLM limitation the paper is concerned with, i.e.\ the evidence statements of papers rated 3-5 as extracted in Section~\ref{sec:llm-classification}. To enrich the text representation for clustering, we follow the approach of \citet{viswanathan2024large} and generate keyphrases for each statement using GPT-4o. The model is prompted to \enquote{provide a comprehensive set of keyphrases describing the LLM limitations discussed in a paper}, with no constraints on the number generated. For example:

\begin{tcolorbox}[colback=gray!10, colframe=gray!40, sharp corners=south, boxrule=0.5pt]
\textbf{Evidence:}  
\enquote{We find that zero-shot CoT reasoning in sensitive domains significantly increases a model's likelihood to produce harmful or undesirable output [...]}\citep{shaikh2022second}

\vspace{0.5em}

\textbf{Generated Keyphrases:} \enquote{zero-shot CoT reasoning limitations}, \enquote{increased harmful output}, \enquote{sensitive domains challenges}, \enquote{prompt format issues}
\end{tcolorbox}

Each set of keyphrases is appended to the original evidence statement, and this combined text serves as input to both clustering approaches. Figure~\ref{fig:hdbscan_lloom_clustering_steps} illustrates the respective pipelines.


\paragraph{Clustering pipeline 1: HDBSCAN + BERTopic}

We employ a density-based clustering approach using HDBSCAN+BERTopic. We follow the standard BERTopic pipeline \citep{grootendorst2022bertopic}: we use OpenAI's \texttt{text-embedding-3-large} model to embed the combined evidence–keyphrase text, and reduce the embedding using UMAP \citep{mcinnes2018umap}, retaining 10 dimensions for ACL and 5 for arXiv. Such low dimensionality has been shown to be effective for HDBSCAN in prior work \citep{rother2020cmce}.

To ensure meaningful clusters and minimize spurious outliers, we tune UMAP, HDBSCAN and BERTopic parameters separately for ACL and arXiv. \ak{This separation accounts for differences in corpus size and topical heterogeneity.} We further apply a distance-based outlier reassignment strategy after observing that many outliers were not true noise but semantically close to existing clusters, suggesting misclassification by HDBSCAN. Full parameter settings and details of the reassignment procedure are provided in Section~\ref{sec:clustering_details} of the supplementary materials.

Finally, clusters are given descriptive names by GPT-4o, based on the top-ranked keywords extracted by BERTopic for each cluster. 

\paragraph{Clustering pipeline 2: LlooM}

For the second clustering approach, we adapt the LLooM concept induction method with modifications tailored to our use case. LlooM involves a process of summarization, clustering, and LLM-based synthesis. It first summarizes documents into bullet points (distill step), then clusters them using HDBSCAN. An LLM then generates a concept (a short, human-readable label that describes the theme of the cluster) and inclusion prompt for each cluster (synthesize step), which are used to score all documents for each concept via zero-shot prompting on a 0–1 scale (score step). For further implementation details, we refer the reader to the original LlooM paper. \citep{lam2024concept}

In our setup, we skip the distill step (summarization of input text into short bullet points), since our dataset already consists of concise quotes from research papers. We generate two concepts per cluster, conducting two rounds of review to refine the concepts. 
The clustering step is performed using \texttt{text-embedding-3-large}, while \texttt{GPT-4o} is used for concept synthesis and iterative review. For final scoring, we employ \texttt{Llama-3.1-70b-Instruct}, and retain only those papers for which the model assigns a 75\% to 100\% confidence score for a given concept.

\ak{\paragraph{Clustering Validation}

We assess the \textit{robustness} of the identified limitation topics using (i) cross-method comparison, which evaluates topic-level alignment (Jaccard overlap, AMI) and trend-level alignment (Kendall's Tau, Spearman's $\rho$) between HDBSCAN and LlooM, and the \textit{validity} of these topics using (ii) human evaluation, in which two expert annotators independently re-annotate papers with limitation topics predefined by HDBSCAN and LlooM, and agreement with each clustering method is measured, based on a stratified sample of 50 ACL and 60 arXiv papers covering all topics identified by both methods. Detailed procedures and results are reported in \ref{sec:hdbscan-vs-lloom-comparison}.}

%% file: structure/4-1-results.tex
\section{Results}\sectionlabel{sec:results}

In this section, we report the results of the LLM-based filtering stage and the clustering stage of our pipeline (Fig.~\ref{fig:experimental_pipeline}). We begin, however, with the validation results of our LLM-based filtering stage, comparing LLM classifications of abstracts to human annotations.

\subsection{LLM-based filtering evaluation}\label{sec:model_eval_results}

\begin{table}
    \caption{Weighted Cohen's Kappa for limitation ratings and pairwise F1 scores for evidence extraction across models and prompts. The best weighted Kappa for each model is \underline{underlined}, while the best score overall is in \textbf{bold}. Evidence extraction is measured in pairwise F1 between each annotator and the model, reported for the best-performing prompts.}
    \label{tab:prompt_results}
    \centering
    \renewcommand{\arraystretch}{0.95}
    \small
    \begin{tabular}{l l cc} 
        \toprule
        \textbf{Model} & \textbf{Prompt} & \textbf{Weighted Kappa} & \textbf{Evidence F1} \\ 
        \midrule
        \multirow{3}{*}{\texttt{Mistral-7B-Instruct-v0.3}} 
        & Prompt 1 & 0.25 & \\
        & Prompt 2 & \underline{0.60} & \\
        & Prompt 3 & \underline{0.60} & 0.36 \\
        \midrule
        \multirow{3}{*}{\texttt{Llama-3.1-70b-Instruct}} 
        & Prompt 1 & 0.60 & \\
        & Prompt 2 & 0.73 & \\ 
        & Prompt 3 & \textbf{\underline{0.74}} & 0.65 \\
        \midrule
        \multirow{3}{*}{GPT-4.o} 
        & Prompt 1 & 0.49 & \\ 
        & Prompt 2 & 0.68 & \\ 
        & Prompt 3 & \underline{0.72} & 0.64 \\
        \midrule
        SBERT + log. regression & — & 0.43 & — \\
        \bottomrule
    \end{tabular}
\end{table}

Table~\ref{tab:prompt_results} summarizes how well different models with different prompts align with human expert annotations. We report quadratic weighted Cohen's Kappa for limitation ratings and pairwise F1 for evidence extraction, measured between each annotator and the model for the best-performing prompts.

As seen in the table, performance improves as prompts become more detailed across all models, with Prompt 3 consistently resulting in the highest Kappa scores but Prompt 2 performing similarly well across models, suggesting that models benefit from clear definitions but less from examples.

Overall, \texttt{Llama-3.1-70b-Instruct} shows the strongest agreement with human annotations, achieving the highest weighted Kappa (0.74) for rating assignment, as well as the highest evidence extraction F1 (0.65). For reference, human–human agreement reaches 0.75 for ratings and 0.71 for evidence extraction. \texttt{GPT-4o} follows closely, with a weighted Kappa of 0.72 and an evidence extraction F1 of 0.64. While \texttt{Mistral-7B} outperforms the baseline (weighted Kappa: 0.43), it lags behind Llama and \texttt{GPT-4o}, with a weighted Kappa of 0.60 and a much lower evidence extraction F1 of 0.36. Taken together, these results indicate that \texttt{Llama-3.1-70b} and \texttt{GPT-4o} can serve as reasonably reliable annotators in our setting, with agreement levels approaching those of human annotators.

\begin{figure}
    \centering
    \resizebox{0.8\textwidth}{!}{
    \begin{tabular}{cc}
        \begin{subfigure}[b]{0.45\textwidth}
            \centering
            \includegraphics[width=0.7\textwidth]{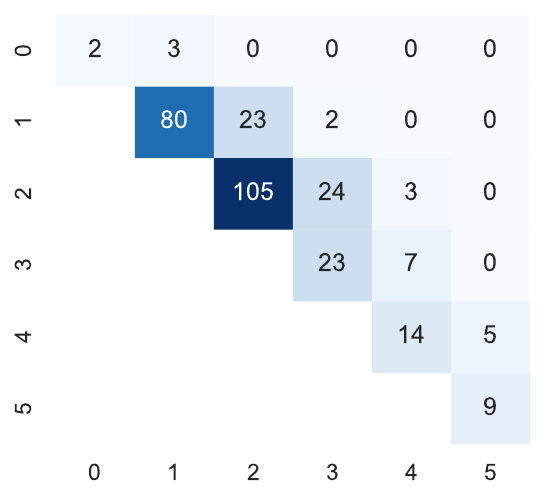}
            \caption{\footnotesize Annotators' agreement}
            \label{fig:annotators-confusion-matrix}
        \end{subfigure} &
        
        \begin{subfigure}[b]{0.45\textwidth}
            \centering
            \includegraphics[width=0.7\textwidth]{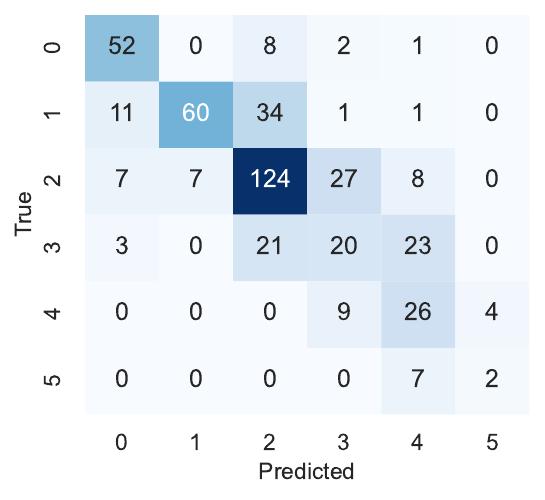}
            \caption{\footnotesize Llama-3.1-70b vs. Final Label}
            \label{fig:llama-confusion-matrix}
        \end{subfigure} 
    \end{tabular}
    }

    \caption{Confusion matrices comparing human agreement (Figure~\ref{fig:annotators-confusion-matrix}) and the predictions of Llama-3.1-70b against the final labels (Figure~\ref{fig:llama-confusion-matrix}). The human agreement matrix is aggregated over all pairwise annotator comparisons, making it symmetric.}
    \label{fig:confusion-matrices}
\end{figure}

\paragraph{Error analysis}

\begin{table*}
    \centering
    \small
    \caption{Comparison of evidence extraction between human annotators and Llama-3.1-70b using Prompt 3. Text highlighted in green indicates parts which both the model and human annotators selected as evidence. Yellow highlights denote evidence selected only by human annotators, while blue highlights indicate evidence selected solely by the model.}
    \label{tab:evidence_extraction_examples}
    \renewcommand{\arraystretch}{1.2}
    \begin{tabularx}{\textwidth}{>{\raggedright}p{5cm} X}
        \toprule
        \textbf{Title} & \textbf{Abstract} \\
        \midrule
        (1) \textit{\enquote{\small Unlocking Adversarial Suffix Optimization Without Affirmative Phrases: Efficient Black-box Jailbreaking via LLM as Optimizer}} \citep{jiang2024unlocking},\\arXiv, August 2024 
        & \enquote{\small \textcolor{ForestGreen}{Despite prior safety alignment efforts, mainstream LLMs can still generate harmful and unethical content when subjected to jailbreaking attacks.} [...] In this paper, we present ECLIPSE, a novel and efficient black-box jailbreaking method utilizing optimizable suffixes. [...] Experimental results demonstrate that ECLIPSE achieves an average attack success rate (ASR) of 0.92 across three open-source LLMs and GPT-3.5-Turbo.} \\
        & {\small \textbf{True label: 3, Predicted: 4}} \\
        \midrule
        (2) \textit{\enquote{\small Can GPT-4V(ision) Serve Medical Applications? Case Studies on GPT-4V for Multimodal Medical Diagnosis}} \citep{wu2023can},\\arXiv, October 2023 
        & \enquote{\small [...] \textcolor{ForestGreen}{Our observation shows that, while GPT-4V demonstrates proficiency in distinguishing between medical image modalities and anatomy, it faces significant challenges in disease diagnosis and generating comprehensive reports.} These findings underscore that while large multimodal models have made significant advancements in computer vision and natural language processing, \textcolor{SoftYellow}{it remains far from being used to effectively support real-world medical applications and clinical decision-making.} [...]} \\
        & {\small \textbf{True label: 4, Predicted: 3}} \\
        \midrule
        (3) \textit{\enquote{\small Still No Lie Detector for Language Models: Probing Empirical and Conceptual Roadblocks}} \citep{levinstein2024still},\\arXiv, June 2023 
        & \enquote{\small \textcolor{SoftBlue}{We consider the questions of whether or not large language models (LLMs) have beliefs, and, if they do, how we might measure them. [...] We provide empirical results that show that these methods fail to generalize in very basic ways.} We then argue that, \textcolor{ForestGreen}{even if LLMs have beliefs, these methods are unlikely to be successful for conceptual reasons.} \textcolor{SoftBlue}{Thus, there is still no lie-detector for LLMs.} [...]} \\
        & {\small \textbf{True label: 4, Predicted: 4}} \\
        \bottomrule
    \end{tabularx}
\end{table*}

\begin{table}
    \centering
    \caption{Distribution of ratings for ACL and arXiv papers. The \textit{ACL Count (\%)} and \textit{arXiv Count (\%)} columns show the number of papers with each rating and their percentage within the ACL and arXiv datasets, respectively. The \textit{Total Count (\%)} column combines both datasets. The last row sums papers with ratings 3–5, referred to as \textit{Limitation Papers}.}
    \label{tab:ratings_distribution}
    \renewcommand{\arraystretch}{0.95}
    \small
    \begin{tabular}{c c c c}  
        \toprule
        \textbf{Rating} & \textbf{ACL Count (\%)} & \textbf{arXiv Count (\%)} & \textbf{Total Count (\%)} \\
        \midrule
        0 & 861 (10.0\%)  & 11,416 (20.6\%) & 12,277 (19.2\%) \\
        1 & 1,463 (17.0\%) & 10,967 (19.8\%) & 12,430 (19.4\%) \\
        2 & 3,911 (45.4\%) & 20,810 (37.5\%) & 24,721 (38.6\%) \\
        3 & 1,274 (14.8\%) & 6,057 (10.9\%)  & 7,331 (11.4\%) \\
        4 & 1,035 (12.0\%) & 5,723 (10.3\%)  & 6,758 (10.5\%) \\
        5 & 78 (0.9\%)     & 481 (0.9\%)     & 559 (0.9\%) \\
        \midrule
        \textbf{Limitation Papers (3–5)} & \textbf{2,387 (27.7\%)} & \textbf{12,261 (22.1\%)} & \textbf{14,648 (22.9\%)} \\
        \bottomrule
    \end{tabular}
\end{table}

To better understand Llama's performance in \emph{limitation rating} beyond Kappa scores, we examine its confusion matrix. Figure~\ref{fig:confusion-matrices} shows that both humans and the model often confuse adjacent categories, such as 2 $\leftrightarrow$ 3, 3 $\leftrightarrow$ 4, which is expected given the ordinal nature of the labels. Overall, the model tends to overestimate rather than underestimate discussions of limitations. In some cases, it misses LLM mentions entirely, predicting label 0 where LLMs are discussed (21 cases). Despite some misclassifications, the model rarely confuses clearly high-rated papers (4–5) with clearly low-rated ones (0–2). This suggests that it can reliably separate papers that meaningfully discuss \lllm\ from those that do not.

Concerning \emph{evidence extraction}, in most cases, the model correctly identifies limitations when they are clearly stated and often fully matches human annotations exactly (Example 1 in Table~\ref{tab:evidence_extraction_examples}). Disagreements primarily stem from the model's tendency to select only 1–2 key sentences, omitting longer arguments that annotators capture (Example 2), or when it chooses full statements instead of specific phrases (Example 3), occasionally even capturing content that humans overlook (for example, the sentence \enquote{...these methods fail to generalize in very basic ways} was missed by a human but selected by the model.) 

As \texttt{Llama-3.1-70b-Instruct} performed best in our evaluation, we select it for the subsequent analysis. In the final classification step, \texttt{Llama-3.1-70b-Instruct} assigns each LLM-focused paper a rating from 0 to 5, with higher scores (3–5) indicating a deeper discussion of limitations. Table~\ref{tab:ratings_distribution} summarizes the results of the large-scale classification by \texttt{Llama-3.1-70b-Instruct} across all ACL and arXiv papers. Most received a score of 2 or lower, with 2,338 ACL papers (27.4\%) and 8,782 arXiv papers (20.9\%) classified as discussing limitations in depth (ratings 3–5). These high-rated papers serve as input for the clustering analysis.

\subsection{LLM and \lllm\ Trends Over Time}\subsectionlabel{sec:lllm-llm-growth}

\vspace{1em}
\begin{center}
\begin{tcolorbox}[title=Key Insights, fontupper=\small, width=0.95
\textwidth]

\begin{itemize}
    \item LLM research is growing rapidly: by late 2024, LLMs account for 75\% of ACL papers and over 30\% of arXiv papers.
    \item Research on \lllm\ grew even more rapidly, with 1 in 3 LLM papers now addressing limitations.
\end{itemize}
\end{tcolorbox}
\end{center}

\begin{figure}
    \centering
    \renewcommand{\thesubfigure}{\roman{subfigure}}

    \begin{subfigure}[t]{0.67\textwidth}
        \centering
        \includegraphics[width=\textwidth]{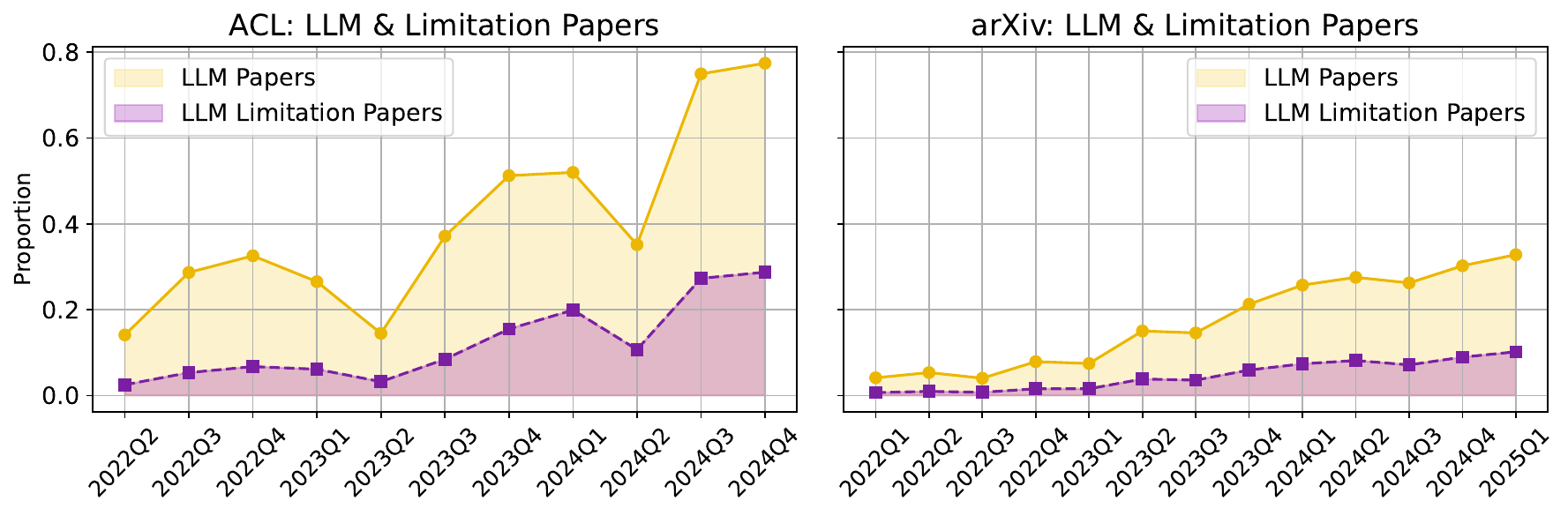}
        \caption{LLM and limitation papers in ACL and arXiv datasets relative to all crawled papers.}
        \label{fig:lllm-llm-crawled-ratio}
    \end{subfigure}
    \hfill
    \begin{subfigure}[t]{0.32\textwidth}
        \centering
        \includegraphics[width=\textwidth]{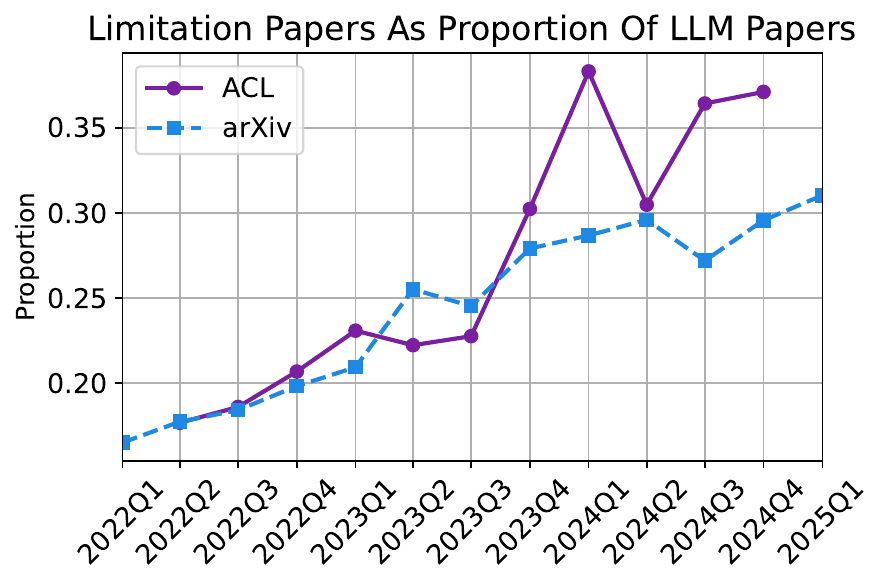}
        \caption{Proportion of LLM limitation papers among all LLM papers.}
        \label{fig:llm-limitation-ratio}
    \end{subfigure}

    \caption{Trends in LLM and LLM limitation research over time. Figure~\ref{fig:lllm-llm-crawled-ratio} shows the share of LLM and limitation papers among all crawled papers, while Figure~\ref{fig:llm-limitation-ratio} illustrates the proportion of limitation papers within LLM research. Note that the limitation trend in (ii) can rise even if it appears flatter in (i), as (ii) reflects growth relative to LLM research, not all papers.}
    \label{fig:llm-lllm-trends}
\end{figure}

Before we turn to clustering, we provide an analysis of the number of LLM-related papers (rating 1 or more) as well as limitations-focused papers (rating 3-5) over time. Figure~\ref{fig:llm-lllm-trends} shows:

\begin{enumerate}[label=(\roman*)]
    \item the proportion of LLM-related and LLM limitation papers among all crawled papers, defined as \( \frac{N_t^{\text{LLM}}}{N_t} \) and \( \frac{N_t^{\text{Lim}}}{N_t} \), where \( N_t \) is the total number of papers \se{at time $t$}, \( N_t^{\text{LLM}} \) the number of LLM-related papers, and \( N_t^{\text{Lim}} \) the number of LLM limitation papers;

    \item the share of limitation papers among LLM-related papers, defined as \( \frac{N_t^{\text{Lim}}}{N_t^{\text{LLM}}} \).
\end{enumerate}

In both corpora, the \emph{(i) overall share of LLM-related papers} has grown substantially since early 2023. This trend is particularly steep in ACL, where, by late 2024, over 75\% of ACL papers are related to LLMs. 

This suggests a notable shift in NLP research, with LLMs becoming central to the field. In arXiv, growth is more moderate but consistent, hitting just above 30\% of papers by the end of the same period. The lower proportion in arXiv might be due to different levels of engagement with LLM research across the categories in our study, as shown in Figure~\ref{fig:llm-lllms-trends-arxiv-categories} in the supplementary material. In cs.CL, LLMs are widely discussed, reaching around 80\% by early 2025, similar to ACL, \ak{while cs.AI shows a sharper but lower rise, peaking around around 50–60\%}. In contrast, in areas like cs.CV and cs.LG, their presence remains below 20\%.

The \emph{(ii) share of limitation papers among LLM-related work} has also grown notably. As shown in Figure~\ref{fig:llm-limitation-ratio}, the proportion of \lllm\ research has steadily increased in both venues. In ACL, this share climbs sharply through early 2024, peaking at nearly 38\% before stabilizing around 35\%. In arXiv, the rise is more gradual, reaching approximately 30\% by the end of 2024. 

Overall, as LLM research accelerates, so does work on their limitations, indicating that the community is not only developing or using new models but also, increasingly, engaging with their risks and shortcomings. 

In the following sections, we refine this analysis and examine these emerging discussions in detail through topic clustering. \ak{We discuss topics identified with HDBSCAN and LlooM, continue with trends identified with LlooM and provide trends based on HDBSCAN trends in the supplementary material.}

%% file: structure/4-2-results-clustering.tex
\subsection{Clustering Results}
\subsectionlabel{sec:clustering_results}

\subsubsection{\ak{Topics identified within ACL and arXiv with HDBSCAN and LlooM}}

\begin{center}
\begin{tcolorbox}[title=Key Insights, fontupper=\small, width=0.95\textwidth]
\begin{itemize}
    \item \ak{Both methods consistently identify shared high-level limitation categories, including \textit{Reasoning}, \textit{Hallucination}, \textit{Security Risks}, \textit{Social Bias}, \textit{Generalization}, and \textit{Long-Context} limitations.
    \item Differences between HDBSCAN and LlooM primarily reflect methodological choices: HDBSCAN concentrates papers into fewer, broader clusters, whereas LlooM's multi-label assignments distribute papers across finer-grained limitation categories.}
\end{itemize}
\end{tcolorbox}
\end{center}

\begin{figure}
    \centering
    \includegraphics[width=\textwidth]{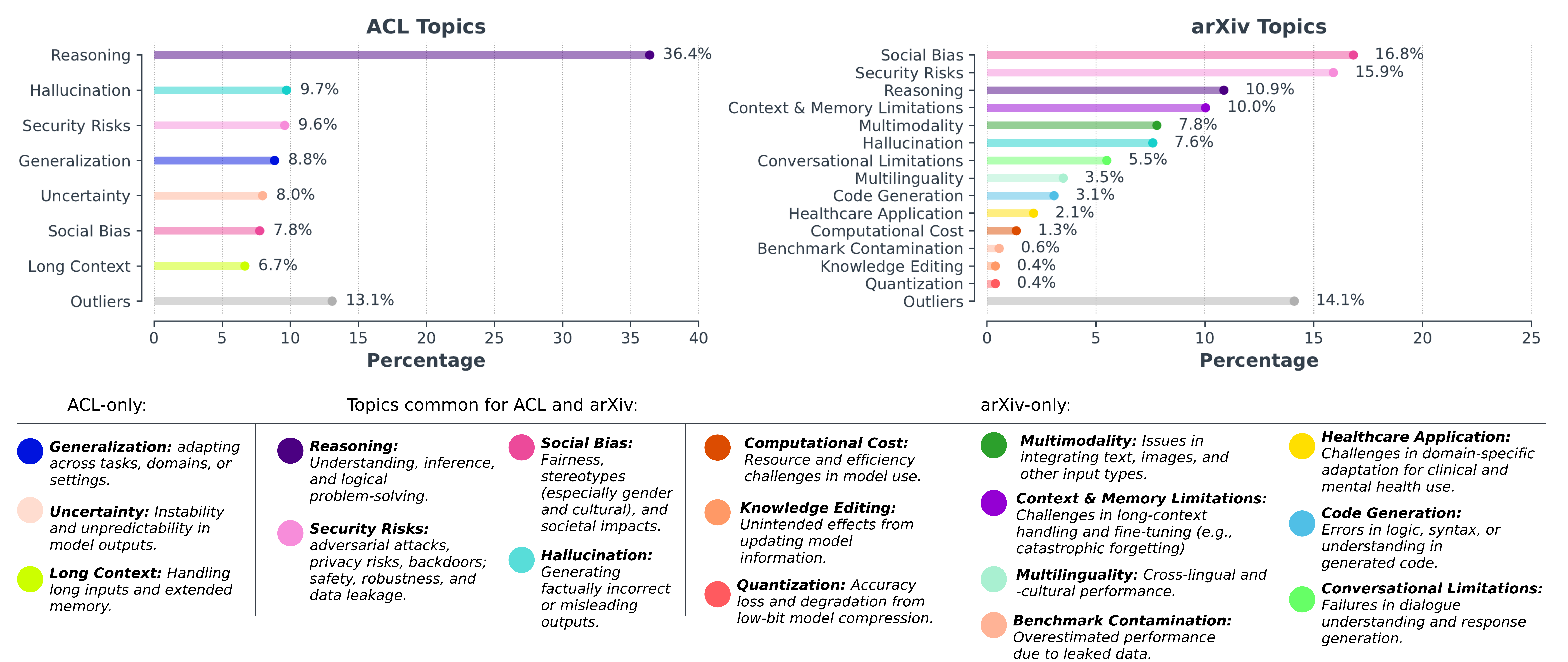}
    \caption{Topics in ACL Anthology and arXiv, clustered using HDBSCAN + BERTopic. Percentages reflect each topic's proportion out of the total LLM limitation papers (2,387 in ACL and 12,261 in arXiv).} 
    \label{fig:acl-arxiv-topics-hdbscan}
\end{figure}

\begin{figure}
    \centering
    \includegraphics[width=\textwidth]{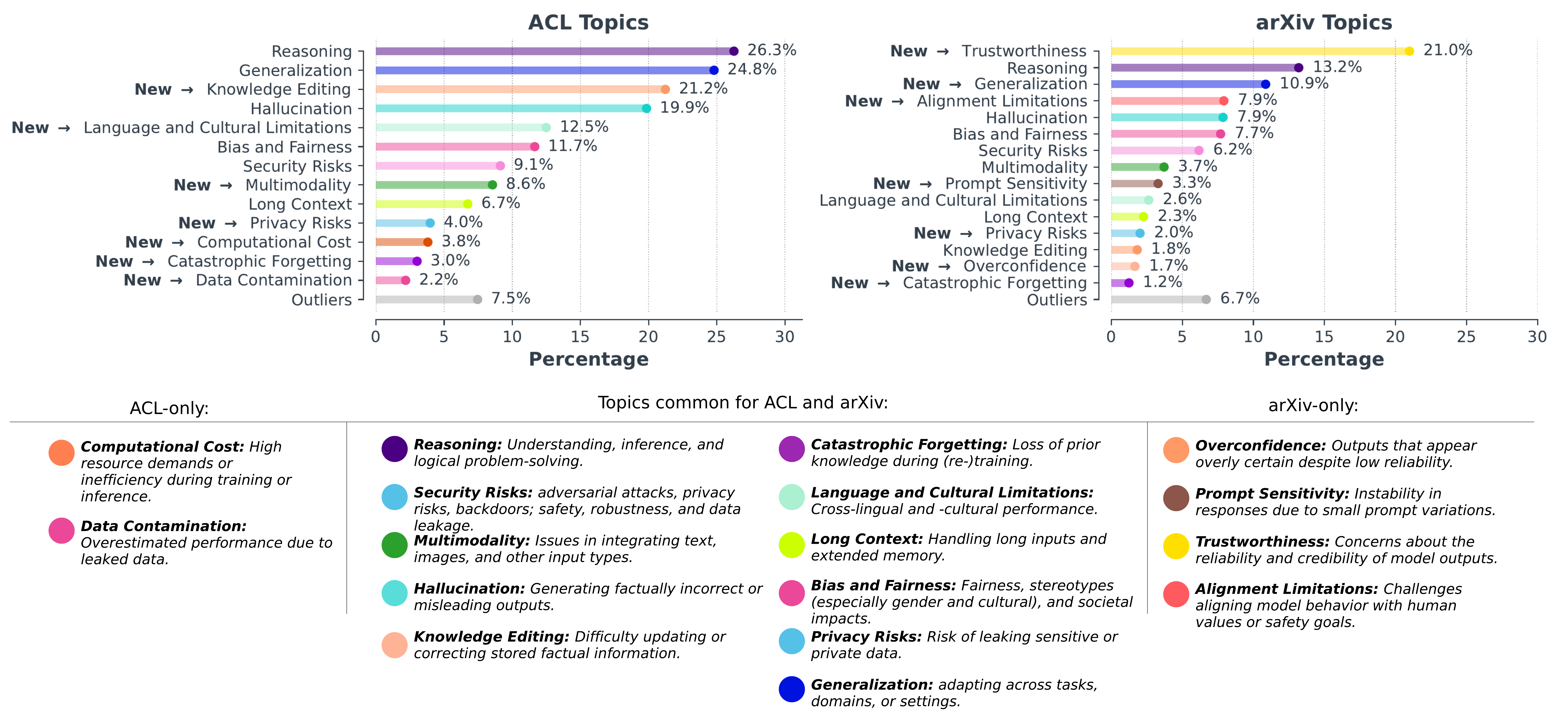}
    \caption{Topics in ACL Anthology and arXiv, clustered using LlooM approach. Percentages reflect each topic's proportion out of the total LLM limitation papers (2,387 in ACL and 12,261 in arXiv). Since papers can be associated with multiple topics, the percentages may exceed 100\% in total.} 
    \label{fig:acl-arxiv-topics-lloom}
\end{figure}

\ak{Figures~\ref{fig:acl-arxiv-topics-hdbscan} and~\ref{fig:acl-arxiv-topics-lloom} show topic distributions for ACL and arXiv under HDBSCAN+BERTopic and LlooM, respectively. HDBSCAN identifies 7 topics in ACL and 15 in arXiv (13.1\% and 14.1\% outliers), while LlooM produces comparable high-level topics with lower outlier rates (7.5\% and 6.7\%).}

\ak{Both clustering methods identify \textit{Reasoning} as the dominant limitation topic in ACL (36.4\% under HDBSCAN vs.\ 26.3\% under LlooM). HDBSCAN concentrates a larger share of papers in this single topic, while LlooM distributes papers more evenly across top limitations, including \textit{Generalization} and \textit{Knowledge Editing}, due to its multi-label assignment. In arXiv, both approaches show a broader topical spread than in ACL, with the most frequent themes being \textit{Social Bias} (16.8\%), \textit{Security Risks} (15.9\%), and \textit{Reasoning} (10.9\%) under HDBSCAN, and \textit{Trustworthiness} (21.0\%), \textit{Reasoning} (13.2\%) and \textit{Generalization} (10.1\%) under LlooM.}

\paragraph{\ak{Topics shared across both methods}}
\ak{Table~\ref{tab:acl-arxiv-topic-papers-hdbscan-lloom} presents representative papers for topics shared between HDBSCAN and LlooM in the ACL and arXiv datasets. Tables~\ref{tab:cluster_keywords_acl_long} and~\ref{tab:cluster_keywords_arxiv_long} in the supplementary material report the dominant terms (top 20 keywords) for each limitation cluster identified with HDBSCAN.}

\ak{Under both clustering methods,} the \textit{Reasoning} cluster captures core cognitive tasks such as natural language understanding (NLU), inference, and logical problem-solving, spanning a broad range of reasoning types (e.g.\, temporal and commonsense). It also includes prompt-related terms such as \textit{chain-of-thought} and \textit{prompt}, highlighting the role of prompt engineering in complex reasoning. \ak{Under HDBSCAN in ACL, however, the \textit{Reasoning} cluster is broader,} absorbing multimodal tasks (example~1 in Table~\ref{tab:acl-arxiv-topic-papers-hdbscan-lloom}), and overlapping with multilinguality and benchmark design. \ak{In contrast, in arXiv under HDBSCAN and in both datasets under LlooM, \textit{Reasoning} is more distinct,} as \textit{Multilinguality} / \textit{Language and Cultural Limitations} (difficulties in handling multilingual input, low-resource languages, or culturally specific content) and \textit{Multimodality} (challenges in integrating and reasoning over inputs from different modalities, such as text and images, example~9) are separated into their own clusters. This difference likely contributes to the higher prevalence of \textit{Reasoning} in ACL under HDBSCAN, though it may also partly reflect clustering artefacts.

\ak{Several shared clusters focus on the correctness and reliability of model outputs in both ACL and arXiv.} The \textit{Hallucination} cluster addresses failures in factual accuracy, with terms related to faithfulness, trust, and correctness (example~2), and also appears in multimodal contexts such as image captioning and generation. The \textit{Security} cluster captures risks including jailbreaks, adversarial prompting, and backdoors (example~3). Under LlooM, security-related concerns further separate into distinct \textit{Security} and \textit{Privacy} topics. The \textit{Social Bias} cluster focuses on fairness, representation, and demographic disparity (example~5), with keywords such as \textit{stereotype}, \textit{gender}, \textit{cultural}, and \textit{demographic}. 

\ak{Other shared clusters reflect technical constraints in model design and deployment.} \textit{Long Context} / \textit{Context \& Memory Limitations} address failures in handling extended inputs and maintaining information over long outputs (example~6 in Table~\ref{tab:acl-arxiv-topic-papers-hdbscan-lloom}), closely related to increased resource demands captured by the \textit{Computational Cost} cluster (example~10). The \textit{Generalization} cluster captures limitations in robustness and domain transfer (example~4). \textit{Knowledge Editing} focuses on the difficulty of updating or correcting model knowledge without retraining, often leading to unintended performance degradation (example~8).

\begin{table*}[!htbp]
\centering
\footnotesize
\caption{\ak{Representative limitation-focused papers for topics that are shared by HDBSCAN and LlooM and that occur in the ACL and/or arXiv corpora. Topics that occur in both corpora are shown once, with dataset indicated by the venue/date preceding the citation. Unless marked with $\dagger$, each paper is assigned to the conceptually corresponding topic by both methods. Topic labels follow HDBSCAN; parentheses show the LlooM label when it differs.} Numbers in the \enquote{Topic} column indicate the example ID referenced in the main text.}
\label{tab:acl-arxiv-topic-papers-hdbscan-lloom}
\renewcommand{\arraystretch}{1.1}
\begin{tabularx}{\textwidth}{>{\centering\arraybackslash}p{1.85cm} >{\raggedright\arraybackslash}X}
\toprule
\textbf{Topic} & \textbf{Evidence} \\
\midrule
\footnotesize (1)\newline Reasoning & \footnotesize \enquote{However, applying this [common-sense] reasoning to multimodal domains, where understanding text and images together is essential, remains a substantial challenge.} (ACL 2024~\citep{park2024picturing}) \\
\midrule
\footnotesize (2) Hallucination & \footnotesize \enquote{Challenges on hallucination and factual inconsistency continue to impede their [LLMs'] wider real-world adoption. [...] However, challenges remain, particularly regarding... generating information not present in the evidence (hallucination).} (EACL 2024~\citep{lv2024coarse}) \\
\midrule
\footnotesize (3)\newline Security Risks & \footnotesize \enquote{The prevalence and strong capability of LLMs present significant safety and ethical risks if exploited by malicious users. [...] Experiments reveal that our attacks effectively compromise the performance of all detectors in the study with plausible generations [...].} (TACL 2024~\citep{shi2024red}) \\
\midrule
\footnotesize (4)\newline Generalization & \footnotesize \enquote{However, in-domain demonstrations are not always readily available in real scenarios, leading to cross-domain in-context learning. Besides, LLMs are still facing challenges in long-tail knowledge in unseen and unfamiliar domains.} (EMNLP 2023~\citep{long2023adapt}) \\
\midrule
\footnotesize (5) Social Bias\newline \textit{(Bias \& Fairness)} & \footnotesize \enquote{We find that masked language models capture societal stigma about gender in mental health: models are consistently more likely to predict female subjects than male in sentences about having a mental health condition (32\% vs. 19\%) [...].} (EMNLP 2022~\citep{lin2022gendered}) \\
\midrule
\footnotesize (6)\newline Long Context & \footnotesize \enquote{However, they face challenges in managing long documents and extended conversations, due to significantly increased computational requirements, both in memory and inference time [...].} (EMNLP 2023~\citep{li2023compressing}) \\
\midrule
\footnotesize (7) Multiling.\newline \textit{(Lang. \& Cult.)} & \footnotesize \enquote{The automated systems [...] are primarily designed for and work far more effectively in English than in the world's other 7,000 languages. [...]} (arXiv, June 2023 \citep{nicholas2023lost}) \\ 
\midrule
\footnotesize (8)\newline Knowledge Editing & \footnotesize \enquote{Existing editing methods lead to inevitable performance deterioration on general benchmarks. [...] When the number of edits is slightly large, the intrinsic knowledge structure of the model is disrupted or even completely damaged.} (arXiv, October 2024 \citep{li2024should}) \\ 
\midrule
\footnotesize (9) Multimodality$^\dagger$ & \footnotesize \enquote{We experimented with state-of-the-art vision and LMs and found that the best (22\%) performed substantially worse than humans (97\%) in understanding figurative language.} (EMNLP 2023 \citep{yosef2023irfl}) \\ 
\midrule
\footnotesize (10) Comput. \newline Cost & \footnotesize \enquote{Training and deploying LLMs are expensive as it requires considerable computing resources and memory...[...]} (arXiv, November 2023 \citep{zhang2023dissecting}) \\ 
\bottomrule
\end{tabularx}
\vspace{0.5em}
\raggedright
\footnotesize
\par\textit{$\dagger$ Assigned to \textit{Reasoning} under LlooM.}
\end{table*}

\paragraph{HDBSCAN-specific clusters}
\ak{In addition to shared topics, HDBSCAN identifies several clusters that do not emerge under LlooM. Representative papers for these clusters are shown in Table~\ref{tab:hdbscan-lloom-specifc-examples} in the supplementary material.}
In ACL, the \textit{Uncertainty} cluster describes behavioral instability, including prompt sensitivity and calibration errors. In arXiv, additional clusters capture a wider range of specialized concerns: \textit{Conversational Limitations} \textit{Code Generation}, \textit{Healthcare Application}, \textit{Benchmark Contamination}, and \textit{Quantization}. These clusters reflect both domain-specific challenges (e.g.\ \textit{Healthcare Application}, which focuses on the use of LLMs in clinical settings and is primarily concerned with their black-box nature, see example~13) and implementation-level trade-offs. For example, \textit{Benchmark Contamination} refers to test data leakage into training sets (example~14), and \textit{Quantization} concerns scaling efficiency (example~15).

\paragraph{\ak{LlooM-specific clusters}}

\ak{A number of topics are specific to LlooM.} These are listed below, with explanations based on the prompts used by LlooM to guide topic assignment (see Table~\ref{tab:acl-cluster-prompts} and Table~\ref{tab:arxiv-cluster-prompts} in the supplementary material). \ak{For each LlooM-specific cluster, we also provide representative paper examples in Table~\ref{tab:hdbscan-lloom-specifc-examples} in the supplementary material}.

\begin{itemize}
    \item \textit{Trustworthiness} (arXiv): the largest cluster in the arXiv set (Figure~\ref{fig:acl-arxiv-topics-lloom}). This category describes concerns about the reliability, transparency, and reproducibility of LLM outputs (see example~19 and 20 which refer to concerns about the reliability of outputs generated by LMs). As a broad category, it often overlaps with related issues like hallucination or alignment, as discussed further in Section~\ref{sec:topic-co-occurrence} of the supplementary material.
    \item \textit{Alignment Limitations} (arXiv): Highlights challenges in aligning LLMs with human values or safety protocols (see example~20 which discusses how models can generate outputs that are untruthful, toxic, or unhelpful despite alignment efforts).
    \item \textit{Prompt Sensitivity} (arXiv): Highlights performance instability when prompts are minimally edited. 
    \item \textit{Overconfidence} (arXiv): Captures cases where LLMs express high certainty despite being incorrect, often due to poor calibration (example~22 shows how persuasive language can mask factual errors). This topic is closely related to the \textit{Uncertainty} cluster identified in ACL under HDBSCAN.
    \item \textit{Privacy Risks} (ACL \& arXiv): previously part of the \textit{Security} cluster in HDBSCAN for both datasets, this now is a distinct category in LlooM. It captures risks of leaking sensitive training data via model outputs or queries. Example~16 (ACL) demonstrates privacy breaches from pretraining on sensitive data.
    \item \textit{Data Contamination} (ACL): inflated evaluation results caused by overlap between training and test datasets. While this topic appeared in arXiv under HDBSCAN (as \textit{Benchmark Contamination}), LlooM identifies it only in ACL (see example~18 which highlights concerns about memorization skewing evaluation).
\end{itemize}

\ak{The comparison of shared and method-specific clusters highlights both similarities and systematic differences between HDBSCAN and LlooM (we quantify these differences in \ref{sec:hdbscan-vs-lloom-comparison}).} While HDBSCAN provides an interpretable, unsupervised clustering of limitations, it assigns each paper to a single cluster and becomes increasingly fragmented as more clusters are added, particularly in the ACL dataset. This makes it difficult to capture overlapping limitations and may complicate the interpretation of trends. By contrast, LlooM allows for broader, non-mutually exclusive categories through multi-label assignment, which reduces fragmentation and provides a more stable basis for longitudinal analysis. Accordingly, we rely on LlooM for the trend analysis in the following section. Trend results derived from HDBSCAN are reported in Section~\ref{sec:hdbscan-trend-analysis} of the supplementary material for completeness.

\subsubsection{Trend Analysis}\subsectionlabel{sec:lloom-trend-analysis}

In this section, we discuss three perspectives on topic dynamics over time:

\begin{enumerate}[label=(\roman*)]
    \item \textbf{LLM-wide share}, measured annually as \( \frac{N^{\text{lim}}_{k,y}}{N^{\text{LLM}}_y} \), to reflect how often limitation topic \( k \) appears in LLM research in year \( y \), relative to the total LLM papers. This shows whether a topic is gaining attention beyond limitations research and becoming part of the general LLM research agenda.
    
    \item \textbf{Limitations share}, measured quarterly as \( \frac{N^{\text{lim}}_{k,q}}{N^{\text{lim}}_q} \), to reflect the share of limitation-focused papers in quarter \( q \) that address topic \( k \). Note the different denominator compared to the LLM-wide share (i): this metric is limited to the subset of limitation-focused papers to show the topic's visibility within the limitations-focused subfield. 
    
    \item Notable shifts in topic trajectories, such as spikes, dips, and periods of stabilization.
    
\end{enumerate}

\noindent
Here, \( N^{\text{lim}}_{k,y} \) is the number of limitation papers on topic \( t \) in year \( y \); \( N^{\text{LLM}}_y \) is the total number of LLM papers in that year; and \( N^{\text{lim}}_{k,q} \), \( N^{\text{lim}}_q \) are the number of limitation papers on topic \( k \) and the total number of limitation papers, respectively, in quarter \( q \).

\paragraph{(i) How are limitation topics represented in the broader growth of LLM research?}

\begin{center}
\begin{tcolorbox}[title=Key Insights, fontupper=\small, width=0.95\textwidth]
\begin{itemize}
    \item The presence of limitation topics in LLM research is increasing across both ACL and arXiv datasets. Topics like \textit{Hallucination}, \textit{Multimodality}, and \textit{Long Context} surge in 2023 and 2024, while longer-standing ones like \textit{Reasoning} grow more gradually and consistently over time.
    \item However, this rise may simply reflect the rapid growth of the limitation field itself.
\end{itemize}
\end{tcolorbox}
\end{center}

\begin{figure}
    \centering
    \caption{Distribution of LLM limitation topics over years for ACL and arXiv, based on clustering results with LlooM. Percentages reflect each topic's proportion out of the total LLM-focused papers (8,635 in ACL and 41,991 in arXiv).}
    \label{fig:lloom-per-year}
    \renewcommand{\thesubfigure}{\roman{subfigure}}
    \resizebox{1.0\textwidth}{!}{
    \begin{tabular}{cc}
        \begin{subfigure}[b]{0.49\textwidth}
            \centering
            \includegraphics[width=\textwidth]{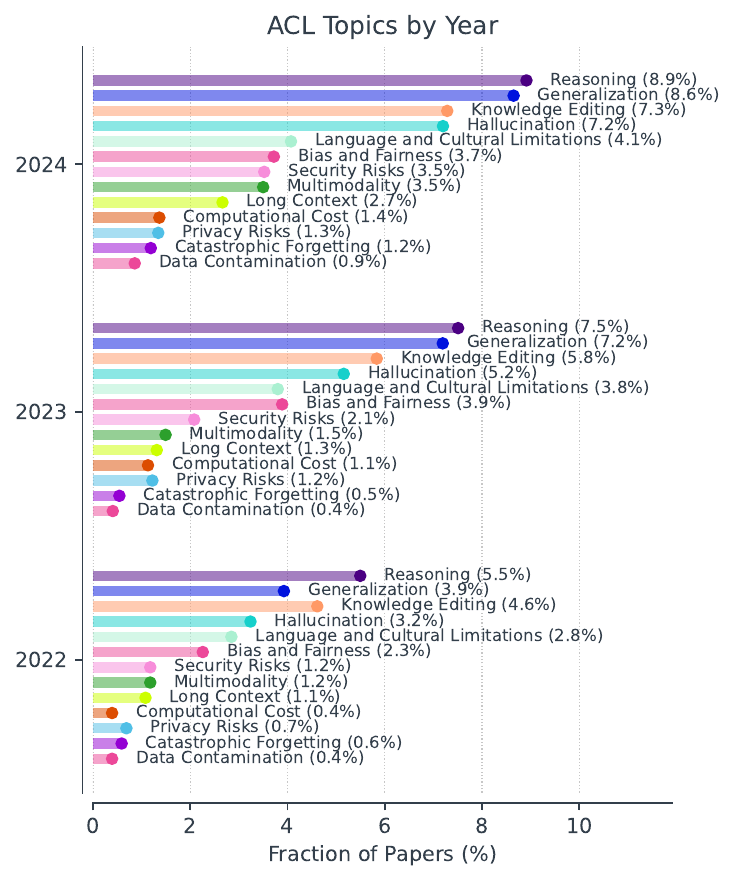}
            \caption{\footnotesize ACL Topic Distribution Per Year (LlooM clustering approach)}
            \label{fig:acl-topic-per-year-lloom}
        \end{subfigure} &
        
        \begin{subfigure}[b]{0.49\textwidth}
            \centering
            \includegraphics[width=\textwidth]{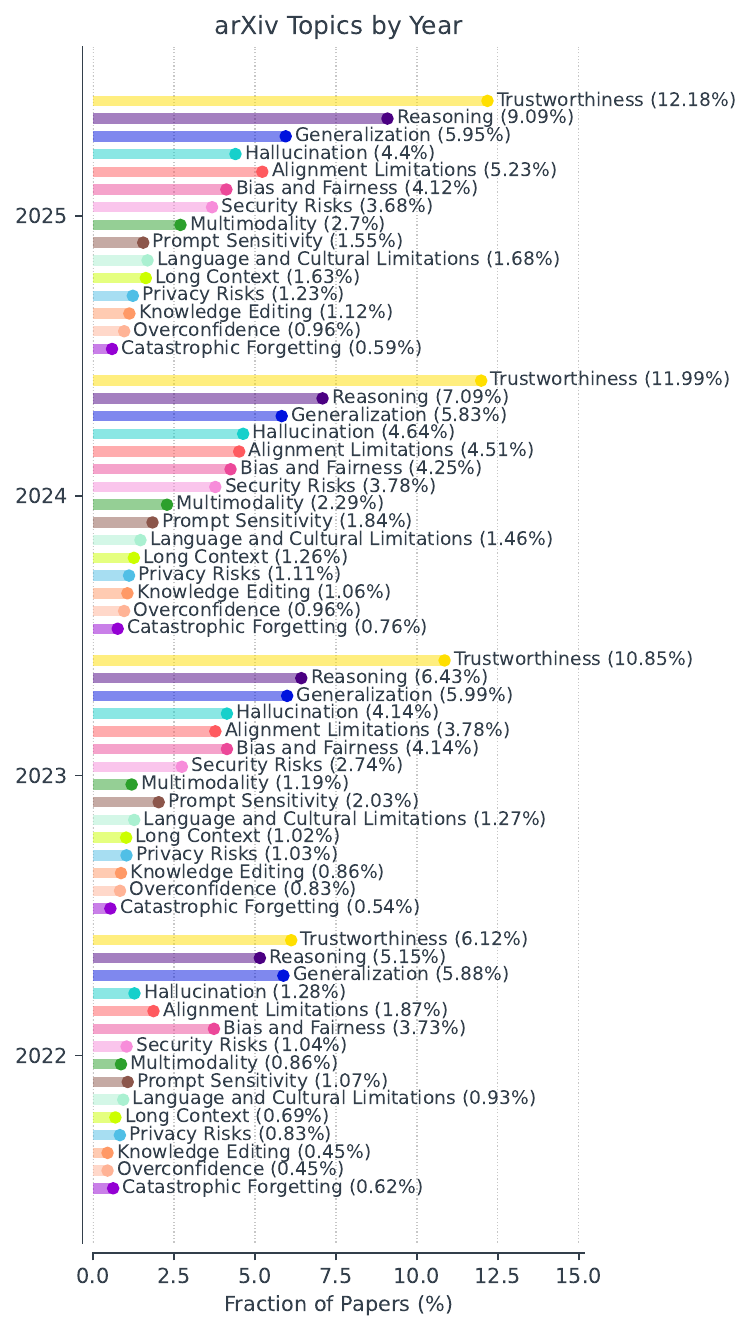}
            \caption{\footnotesize arXiv Topic Distribution Per Year (LlooM clustering approach)}
            \label{fig:arxiv-topic-per-year-lloom}
        \end{subfigure} 
    \end{tabular}
    }
\end{figure}

We begin by examining how visible different limitation topics are across the broader LLM research field. Figure~\ref{fig:lloom-per-year} shows the annual distribution of LLM limitation topics across ACL and arXiv, normalized by all LLM-focused papers. These proportions reflect the \textit{overall visibility} of each topic. Additionally, to capture how visibility changes over time, we compute the \textit{relative percentage change} in LLM-normalized topic share from year \( y \) to \( y+1 \), defined as \( \frac{\text{Share}_{y+1} - \text{Share}_y}{\text{Share}_y} \times 100 \), where \(\text{Share}_y\) refers to the LLM-wide share of a given topic in year \( y \), i.e., the proportion of all LLM papers that address that topic (see Table~\ref{tab:relative_growth_acl_arxiv} in supplementary material).

As seen in Figure~\ref{fig:lloom-per-year}, most limitation topics show an increase in visibility within LLM research over the years. However, topics differ in how their share of LLM research changes over time. Some concerns surged in visibility within LLM research at specific moments (more than doubling in share), such as \textit{Multimodality} (+133\%), \textit{Long Context} (+108\%), \textit{Catastrophic Forgetting} (+140\%) in ACL 2024, and \textit{Hallucination} (+223\%), \textit{Security Risks} (+163\%), and \textit{Alignment Limitations} (+102\%) on arXiv in 2023, reflecting heightened attention to certain types of \lllm\, following widespread LLM deployment. Others, like \textit{Reasoning} and \textit{Knowledge Editing}, show steadier growth across venues. Meanwhile, topics such as \textit{Bias and Fairness} and \textit{Language and Cultural Limitations} peaked in ACL 2023 but declined in 2024 (from +70\% and +36\% to –5\% and +8\%, respectively), while concerns like \textit{Prompt Sensitivity} and \textit{Overconfidence} on arXiv fell steadily after brief spikes. Since the data for 2025 data is incomplete, recent drops should be interpreted with caution. 

Overall, LLM-normalized trends confirm that limitation topics are becoming more prevalent within the broader LLM research. Most topics show growth, some slower, some rapidly. However, a topic's increasing presence in LLM research may simply reflect the overall expansion of limitation research, rather than increased relative focus on that specific topic. Therefore, in the next bullet point, 
we examine whether these trends hold within limitation-focused work.

\paragraph{(ii) What are the trends within LLM limitation research?}

\begin{center}
\begin{tcolorbox}[title=Key Insights, fontupper=\small, width=0.95\textwidth]
\begin{itemize}
    \item Within \lllm\ research, most limitation topics remain stable from 2022 to 2025 in both ACL and arXiv, with only a few showing significant shifts.
    \item \textit{Long Context} increases significantly in ACL, while \textit{Multimodality}, \textit{Security Risks}, and \textit{Alignment Limitations} rise in arXiv.
    \item \textit{Generalization} and \textit{Bias and Fairness} decline significantly in arXiv, but no topics show statistically significant decline in ACL.
\end{itemize}
\end{tcolorbox}
\end{center}

\begin{figure}
    \centering
    \includegraphics[width=0.9\textwidth]{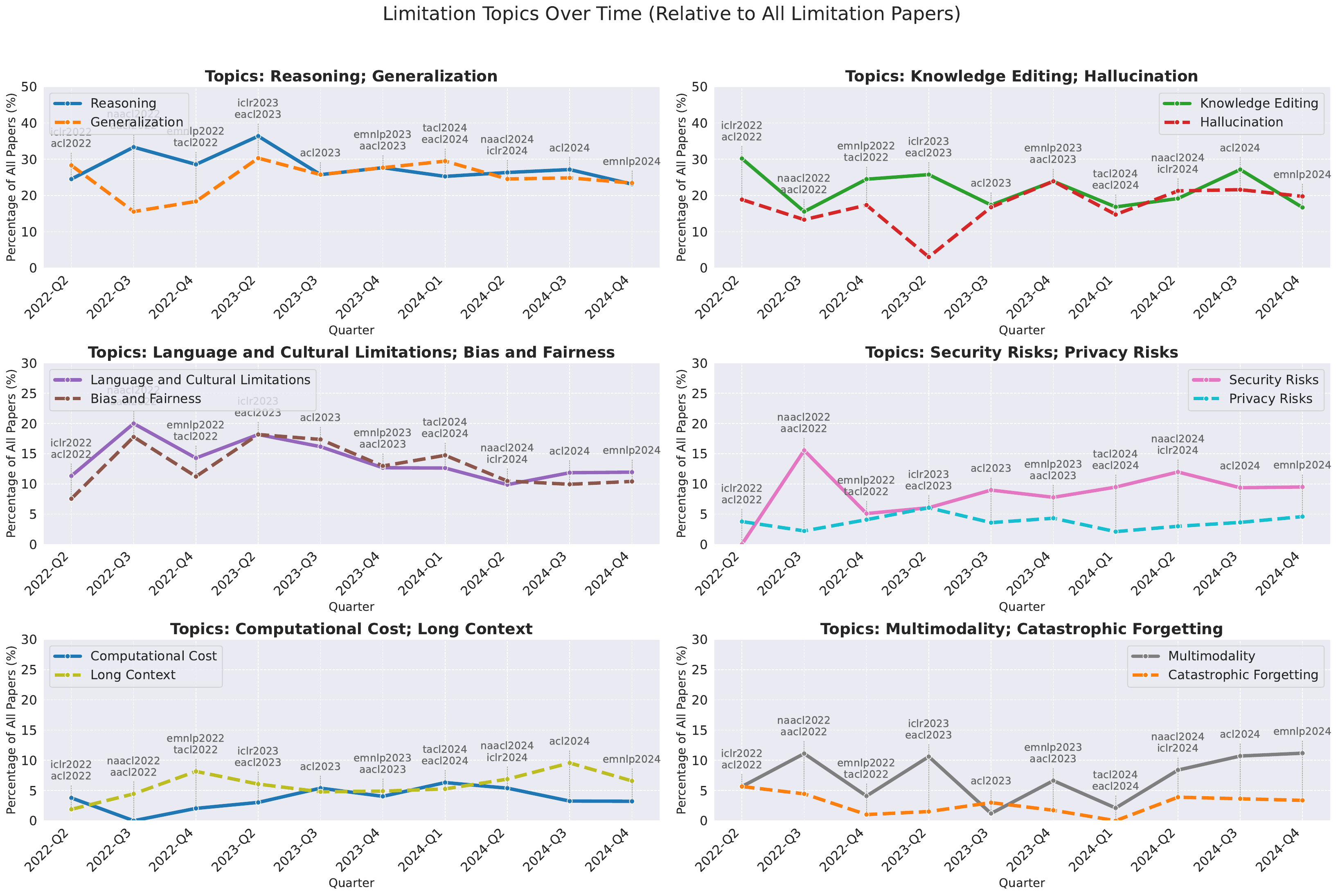}
    \caption{\lllm\ topics trends 
    for the ACL dataset based on LLooM clustering approach. Note that y-axis limits vary across subplots to reflect differences in topic prevalence and improve visualization.}
    \label{fig:acl-lloom-trends}
\end{figure}

Figure~\ref{fig:acl-lloom-trends} shows how the distribution of limitation topics within the ACL dataset has changed over time. We evaluate the significance of these trends using the Mann-Kendall test \citep{mann1945nonparametric, kendall1948rank} for monotonic trend detection.

\begin{itemize}
\item $\uparrow$ \textbf{Increasing:}
\textit{Long Context} shows an upward trend, rising from around 2\% in 2022-Q2 to a peak of 10\% by 2024-Q3. This trend is statistically significant according to the Mann-Kendall test ($\tau$ = 0.51, $p$ = 0.0491). \textit{Security Risks} also shows a similar upward trajectory ($\tau$ = 0.47), though it does not reach significance at the $p < 0.05$ threshold.

\item $\downarrow$ \textbf{Decreasing:}
No topics show statistically significant declines. However, \textit{Bias and Fairness} drops from $\sim$17\% to 10\%, and \textit{Language and Cultural Limitations} from $\sim$20\% to 12\% between 2023-Q2 and 2024-Q4. \textit{Computational Cost}, after a steady rise in the period from late 2022 to late 2023,  decreases from $\sim$5\% in the late 2023 to below 5\% by 2024-Q4. None of these changes are statistically significant.

\item $\rightarrow$ \textbf{Stable or Fluctuating:}
Most topics remain steady over time: \textit{Reasoning}, \textit{Generalization}, and \textit{Hallucination} fluctuate between 10–35\%, while \textit{Knowledge Editing} varies more widely (18–39\%) and \textit{Multimodality} stays between 6–11\%, with a slight increase after early 2024. \textit{Privacy Risks} (3–6\%), \textit{Security Risks} (peaking at 15\% in 2022-Q3 but mostly under 10\%), and \textit{Catastrophic Forgetting} (below 5\%) remain consistently low.

\end{itemize}

\begin{figure}
    \centering
    \includegraphics[width=0.9\textwidth]{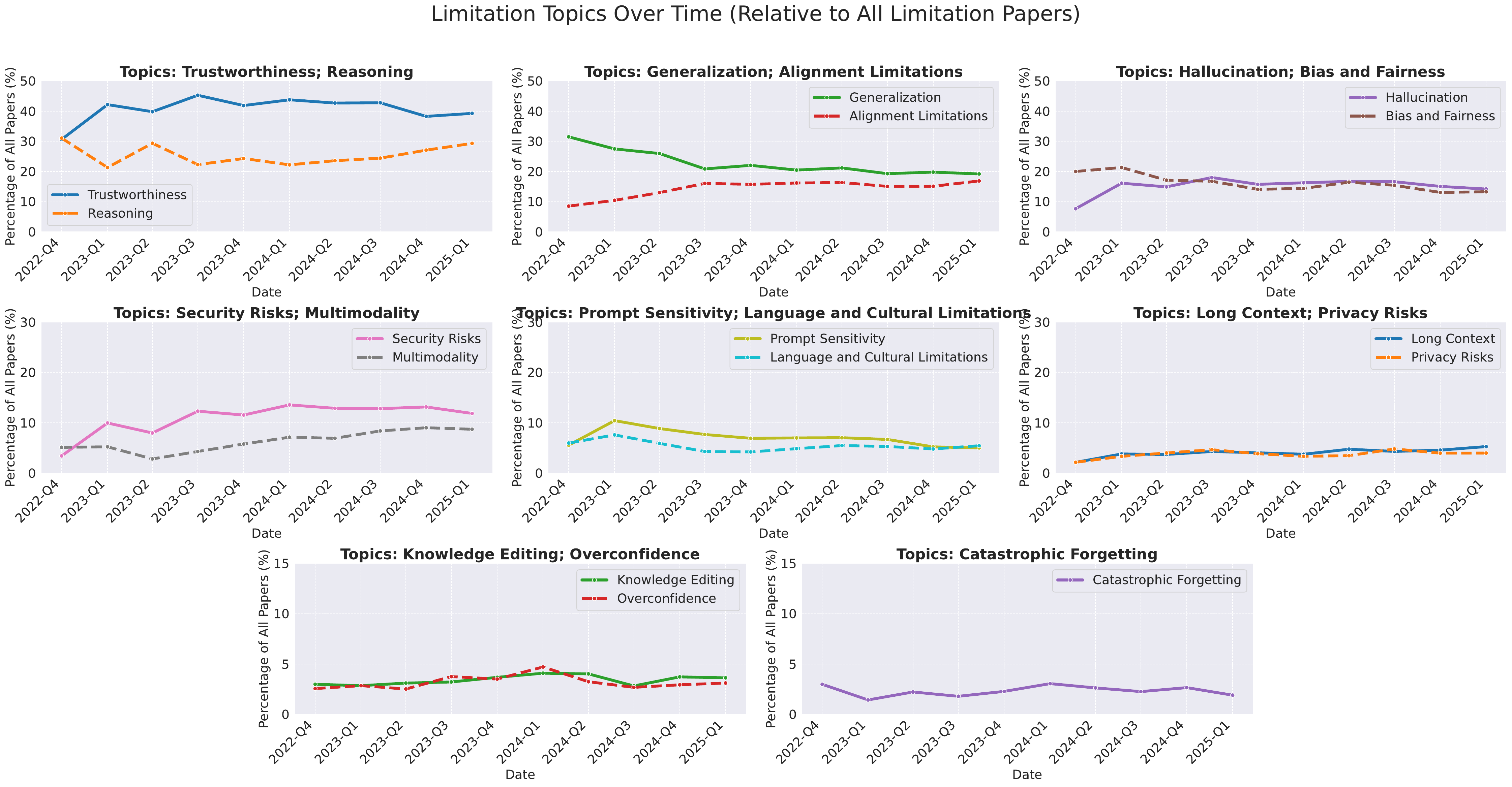}
    \caption{\lllm\ topics trends 
    for the arXiv dataset based on LLooM clustering approach. Note that y-axis limits vary across subplots to reflect differences in topic prevalence and improve visualization.}
    \label{fig:arxiv-lloom-trends}
\end{figure}

For the arXiv dataset, we observe the following trends over time according to LlooM (Figure~\ref{fig:arxiv-lloom-trends}):

\begin{itemize}
    \item $\uparrow$ \textbf{Increasing:}
    \textit{Multimodality}, \textit{Security Risks}, \textit{Alignment Limitations}, and \textit{Knowledge Editing} all show statistically significant upward trends according to the Mann-Kendall test ($p < 0.05$). \textit{Multimodality} rises from approximately 2\% in 2022 to nearly 10\% by late 2024, \textit{Security Risks} increases from around 9\% to 11\%, and \textit{Alignment Limitations} grows from a post-2022 dip to around 18\% by 2025. \textit{Knowledge Editing} increases early on but stabilizes below 5\%.
    Notably, \textit{Hallucination} also shows a positive trend ($\tau$ = 0.38), but this increase is not statistically significant under the Mann-Kendall test.
    
    \item $\downarrow$ \textbf{Decreasing:}
    \textit{Generalization} declines from around 35\% in the second quarter of 2022 to 20\%, and \textit{Bias and Fairness} drops from a peak near 25\% to $\sim$13\%.
    Both of these downward trends are statistically significant ($p < 0.05$).
    
    \item $\rightarrow$ \textbf{Stable or Fluctuating:}
    Most remaining topics show no significant directional movement. \textit{Reasoning} stays around 20–30\%, \textit{Trustworthiness} holds at 40–45\%, and \textit{Overconfidence}, \textit{Long Context}, and \textit{Privacy Risks} remain within narrow ranges. \textit{Prompt Sensitivity} declines from 10\% to $\sim$5\% (2023–2025), while \textit{Hallucination} and \textit{Language and Cultural Limitations} rise modestly and then level off. \textit{Catastrophic Forgetting} stays consistently below 4\%.
    
\end{itemize}

These results confirm that not all LLM-wide trends carry over into increased focus within limitation-specific research. While some topics, such as \textit{Multimodality}, \textit{Security Risks}, \textit{Alignment Limitations}, and \textit{Knowledge Editing}, do show consistent growth within limitation-focused work, the majority remain flat or variable despite gaining visibility across the broader LLM field, as discussed earlier. However, the Mann-Kendall test only captures consistent upward or downward trends, not short-term changes, which may explain why topics like \textit{Security Risks} (ACL) and \textit{Hallucination} (arXiv) show visible growth without being statistically significant. We examine these kinds of short-term spikes and dips in the next paragraph.

\paragraph{(iii) How do topics shift in ways not captured by overall trends?}

\begin{center}
\begin{tcolorbox}[title=Key Insights, fontupper=\small, width=\textwidth]
\begin{itemize}
    \item Limitation topics stabilize around 2023-Q2, either plateauing or beginning steady growth, after earlier volatility. For example, technical concerns (e.g., \textit{Hallucination}, \textit{Alignment Limitations}) rise and plateau, while social topics decline after 2023-Q2.
    \item This shift coincides with rising paper volume and the release of ChatGPT and other major models in early 2023.
\end{itemize}
\end{tcolorbox}
\end{center}

Across both ACL and arXiv, 2023-Q2 marks a shift in the \lllm\ research (Figure~\ref{fig:acl-lloom-trends}, Figure~\ref{fig:arxiv-lloom-trends}). Before this, topic shares are volatile (see e.g. a spike in \textit{Security Risks} in 2022-Q3), particularly in ACL. After early 2023, topics begin to stabilize across both datasets: \textit{Reasoning} levels off, \textit{Generalization} remains flat or slightly declines before stabilizing, and newer concerns like \textit{Security Risks}, \textit{Hallucination}, and \textit{Alignment Limitations} rise sharply and then plateau. \textit{Multimodality} stabilizes somewhat later, starting a steady increase in ACL around early 2024, but earlier in arXiv (around Q2 2023). In contrast, socially focused topics such as \textit{Bias and Fairness} and \textit{Language and Cultural Limitations} decline in share after mid-2023, reflecting the integration of new concerns into the discourse.

The stabilization of topic trends coincides with a sharp rise in raw paper counts beginning in 2023 (see Figure~\ref{fig:acl-arxiv-absolute-counts}), indicating not just increased research volume, but a shift toward a 
more coherent field. Before 2023-Q2, most topics appear in fewer than 25 ACL papers and under 100 in arXiv, making early signals harder to interpret. This growth aligns with the release of ChatGPT in November 2022, as well as the emergence of other major models like GPT-4 \citep{achiam2023gpt}, PaLM \citep{chowdhery2023palm}, and LLaMA \citep{touvron2023llama}
between February and July 2023, which likely contributed to the expansion and differentiation of LLM limitations research during this period.

\subsubsection{LLLMs Topics Distribution Across ArXiv Categories}\label{sec:arxiv_categories_analysis}

Our analysis shows that \lllm\ span a broad range of concerns, from reasoning and generalization to bias, safety, and multimodality. 
ArXiv's category system offers a way to examine how different research communities engage with these topics. We analyze topic distributions across categories to understand where this work is published and which concerns dominate in specific domains.

In our dataset, most \lllm\ papers are concentrated in \texttt{cs.CL} (Computation \& Language; 58.7\%), followed by \texttt{cs.AI} (Artificial Intelligence; 8.7\%), \texttt{cs.CV} (Computer Vision; 6.6\%), and \texttt{cs.LG} (Machine Learning; 3.3\%). This is expected, since these categories were used as our arXiv search criteria. Notably, we also observe papers with categories like \texttt{cs.CY} (cybersecurity), \texttt{cs.SE} (software engineering), and \texttt{cs.HC} (human-computer interaction), which appear as a result of multi-categorization.

\begin{figure}
    \centering
    \caption{Limitation topic trends across four main arXiv categories (cs.CL, cs.LG, cs.AI, cs.CV). To maintain visual clarity, only the five most frequent topics (based on their overall percentage distribution across categories, as shown in Figure~\ref{fig:arxiv-top-categories-lloom} in the supplementary material) are shown; others are grouped as \enquote{Other.} Grey shading in cs.AI and cs.CV marks periods before 2023 with insufficient data for these categories.} 
    \label{fig:arxiv-categories-trends}
    \includegraphics[width=0.9\textwidth]{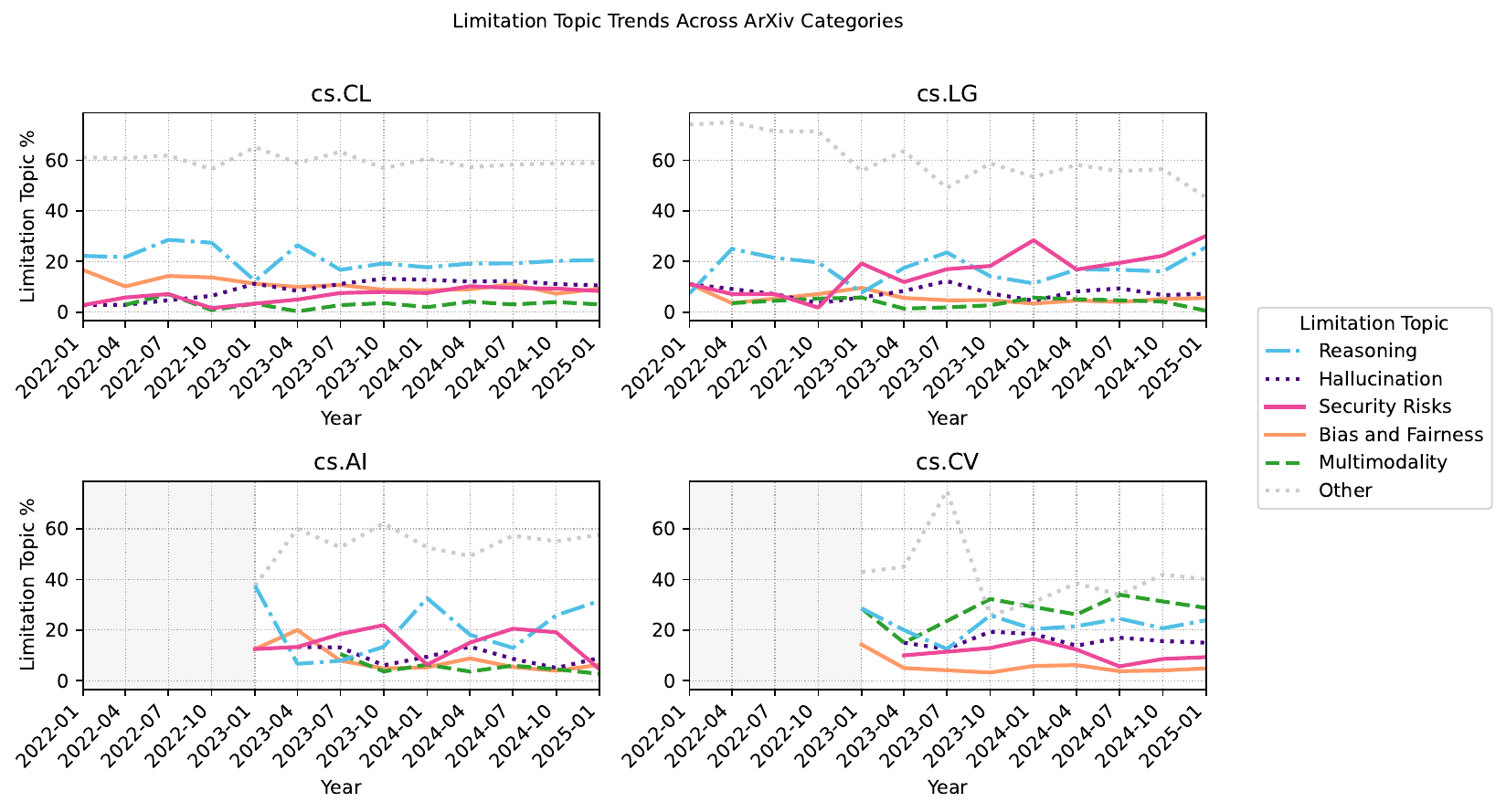}
\end{figure}

\begin{figure}
    \centering
    \includegraphics[width=0.9\textwidth]{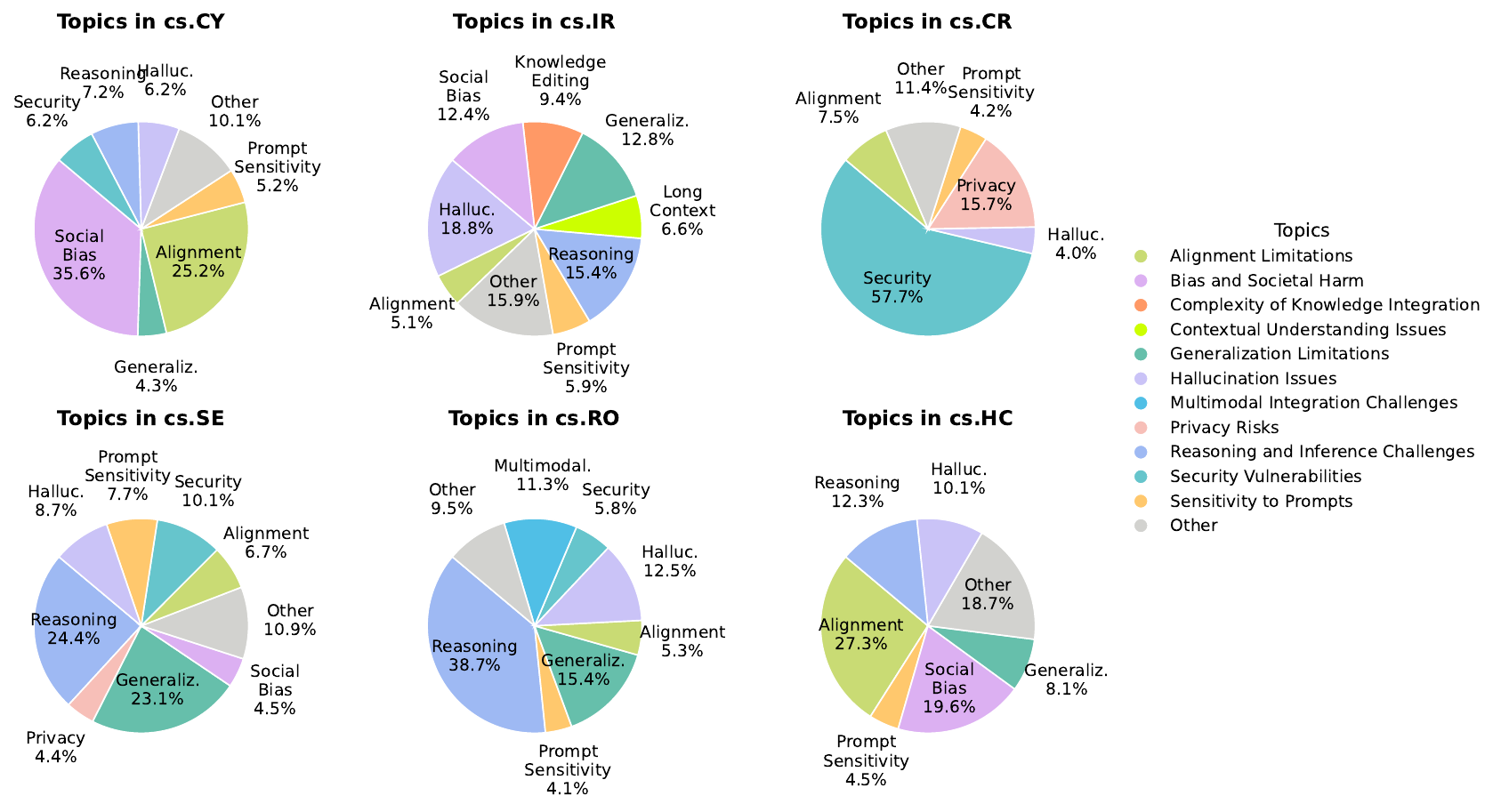}
    \caption{Distribution of limitations-related topics in six arXiv categories with lower paper counts in our dataset, as shown in Figure~\ref{fig:arxiv-top-categories-lloom}. Each chart includes only papers where the category is assigned as the primary arXiv category. Topics that make up less than 3\% are grouped under \textit{Other}.}
    \label{fig:arxiv-small-categories}
\end{figure}

Although many categories share topics such as \textit{Trustworthiness}, \textit{Reasoning}, \textit{Generalization}, and \textit{Alignment Limitations}, both their overall topic composition and temporal dynamics vary by field. Figure~\ref{fig:arxiv-top-categories-lloom} (right) in the supplementary material, shows how topic shares differ across arXiv categories, while Figure~\ref{fig:arxiv-categories-trends} illustrates how topics in the four largest categories evolve over time. \textbf{Key trends in the largest categories are as follows:}

\begin{itemize}
    \item \texttt{cs.CL} (Computation and Language) covers nearly all \lllm, with \textit{Reasoning} dominating across the full time range. \textit{Bias and Fairness} rises mid-2023, but is overtaken by \textit{Hallucination} by the end of the year. Other topics like \textit{Security} and \textit{Multimodality} remain marginal.
    
    \item \texttt{cs.LG} (Machine Learning) and \texttt{cs.AI} (Artificial Intelligence) follow similar distributions to \texttt{cs.CL}, but put more weight on \textit{Security Risks} (10.8\% in cs.LG, 9.1\% in cs.AI). In cs.LG, this topic rises sharply in late 2023, reflecting growing concern with adversarial attacks. In contrast, cs.AI shows more fluctuation, alternating between a focus on \textit{Reasoning} and \textit{Security Risks}, indicating a split between safety and inference evaluation concerns.
    
    \item \texttt{cs.CV} (Computer Vision and Pattern Recognition) diverges from the others in its dominant focus on \textit{Multimodality} (21.7\%), which becomes the leading limitation category from early 2023 onward, driven by the rise of vision-language models. \textit{Hallucination} and \textit{Reasoning} remain present but secondary.
\end{itemize}

Beyond the four primary categories included in our analysis, several \textbf{smaller arXiv categories} show up, often exhibiting a clear focus on domain-specific concerns. Figure~\ref{fig:arxiv-small-categories} highlights six such cases. \texttt{cs.CY} (Computers and Society) and \texttt{cs.HC} (Human-Computer Interaction) emphasize value alignment and societal impact, with high shares of \textit{Social Bias} and \textit{Alignment Limitations}, reflecting ethical and user-centered concerns. \texttt{cs.CR} (Cryptography and Security) is dominated by \textit{Security Risks} (57.7\%), consistent with its focus on adversarial threats and privacy vulnerabilities. \texttt{cs.IR} (Information Retrieval) distributes attention across \textit{Hallucination}, \textit{Reasoning}, and \textit{Knowledge Editing}, likely due to challenges in document-grounded generation and factual consistency. \texttt{cs.SE} (Software Engineering) frequently discusses \textit{Reasoning} and \textit{Generalization}, which aligns with LLMs used in code generation and developer tooling. Finally, \texttt{cs.RO} (Robotics) highlights \textit{Reasoning} and \textit{Multimodality}, reflecting perception and control challenges in embodied settings. Though smaller in volume, these categories reflect more targeted concerns tied to specific application domains.

Together, these disciplinary patterns illustrate how research on \lllm\ is not only growing, but also diversifying in focus based on domain needs.

%% file: structure/4-3-results-comparison.tex
\subsection{\ak{Clustering Methods Comparison}}\subsectionlabel{sec:hdbscan-vs-lloom-comparison}

The aforementioned topic distributions differ depending on the clustering method. In this section, we compare HDBSCAN+BERTopic (Section~\ref{sec:hdbscan-trend-analysis}) and LlooM (\ref{sec:lloom-trend-analysis}) to identify which findings are stable and which may be method-specific. We first assess trend agreement, then explore methodological differences to explain any result divergences.

\subsubsection{\ak{Cross-Method Comparison Between HDBSCAN and LlooM}}

\begin{table}[h]
    \centering
    \renewcommand{\thesubtable}{\roman{subtable}}
    \caption{Comparison of limitation trends identified by HDBSCAN and LlooM for ACL and arXiv datasets. 
    Trend direction (\textit{\textcolor{green}{\textuparrow} increasing, \textcolor{gray}{\textrightarrow} flat, \textcolor{red}{\textdownarrow} decreasing}) is based on Kendall's Tau from the Mann-Kendall test. Significance is indicated with an asterisk (*) and reported only for increasing or decreasing trends. Match symbols: \textit{\checkmark $=$\ full agreement (direction and significance), \textcolor{lightgray}{\checkmark} $=$\ partial agreement (direction only), \texttimes $=$\ disagreement.}
    \newline For Spearman $\rho$ values, asterisk (*) indicates $p > 0.05$ (significant correlation).}
    \label{tab:trend_comparison_hdbscan_lloom_combined}
    \renewcommand{\arraystretch}{1.2}
    \small

    \begin{subtable}[t]{\textwidth}
        \centering
        \caption{ACL Dataset}
        \label{tab:trend_comparison_hdbscan_lloom_acl}
        \begin{tabular}{l l c c c c}
            \toprule
            \textbf{HDBSCAN Topic} & \textbf{LlooM Topic} & \textbf{HDBSCAN} & \textbf{LlooM} & \textbf{Match} & \textbf{Spearman $\rho$} \\
            \midrule
            Security & Security Risks & \textcolor{gray}{\textrightarrow} & \textcolor{green}{\textuparrow} & \texttimes & 0.661* \\
            Generalization & Generalization & \textcolor{gray}{\textrightarrow} & \textcolor{gray}{\textrightarrow} & \checkmark & 0.552 \\
            Social Bias & Bias and Fairness & \textcolor{gray}{\textrightarrow} & \textcolor{gray}{\textrightarrow} & \checkmark & 0.539 \\
            Hallucination & Hallucination & \textcolor{green}{\textuparrow} & \textcolor{gray}{\textrightarrow} & \texttimes & 0.539 \\
            Reasoning & Reasoning & \textcolor{red}{\textdownarrow} & \textcolor{gray}{\textrightarrow} & \texttimes & 0.430 \\
            Long Context & Long Context & \textcolor{gray}{\textrightarrow} & \textcolor{green}{\textuparrow}* & \texttimes & 0.006 \\
            \bottomrule
        \end{tabular}
    \end{subtable}

    \vspace{1em}

    \begin{subtable}[t]{\textwidth}
        \centering
        \caption{arXiv Dataset}
        \label{tab:trend_comparison_hdbscan_lloom_arxiv}
        \begin{tabular}{l l c c c c}
            \toprule
            \textbf{HDBSCAN Topic} & \textbf{LlooM Topic} & \textbf{HDBSCAN} & \textbf{LlooM} & \textbf{Match} & \textbf{Spearman $\rho$} \\
            \midrule
            Multimodality & Multimodality & \textcolor{green}{\textuparrow}* & \textcolor{green}{\textuparrow}* & \checkmark & 0.855* \\
            Hallucination & Hallucination & \textcolor{green}{\textuparrow}* & \textcolor{green}{\textuparrow} & \textcolor{lightgray}{\checkmark} & 0.552 \\
            Context \& Memory Lim. & Long Context & \textcolor{red}{\textdownarrow}* & \textcolor{gray}{\textrightarrow} & \texttimes & -0.224 \\
            Knowledge Editing$\dag$ & Knowledge Editing$\dag$ & \textcolor{green}{\textuparrow} & \textcolor{green}{\textuparrow}* & \textcolor{lightgray}{\checkmark} & 0.632* \\
            Security Risks & Security Risks & \textcolor{green}{\textuparrow} & \textcolor{green}{\textuparrow}* & \textcolor{lightgray}{\checkmark} & 0.782* \\
            Multilinguality & Language \& Cultural Lim. & \textcolor{gray}{\textrightarrow} & \textcolor{gray}{\textrightarrow} & \checkmark & 0.927* \\
            Social Bias & Bias \& Fairness & \textcolor{gray}{\textrightarrow} & \textcolor{red}{\textdownarrow}* & \texttimes & 0.758* \\
            Reasoning & Reasoning & \textcolor{gray}{\textrightarrow} & \textcolor{gray}{\textrightarrow} & \checkmark & 0.624 \\
            \bottomrule
        \end{tabular}
    \end{subtable}

    \vspace{0.5em}
    \raggedright
    \small
    \textit{$\dag$ Topics are matched by Jaccard overlap, except for \textit{Knowledge Editing}, which was manually aligned based on topic names due to weak overlap.}
\end{table}

To compare the trends, we report (i) Kendall's Tau for each individual topic (as determined in the trend analyses in Section~\ref{sec:hdbscan-trend-analysis} in the supplementary material and \ref{sec:lloom-trend-analysis}) to assess alignment in trend direction and significance, and (ii) the Spearman correlation of the time series between matched topics from HDBSCAN and LlooM to assess the similarity of overall trend shapes.
To identify matching topics between HDBSCAN and LlooM, \ak{we first identify candidate matches} for each cluster based on identical or semantically similar names \ak{and validate these} by computing the Jaccard overlap between their associated paper sets\footnote{\ak{All matches except \textit{Knowledge Editing} were confirmed by highest Jaccard overlap; \textit{Knowledge Editing} was aligned manually.}} Specifically, for each cluster produced by HDBSCAN, we compute its Jaccard similarity with all LlooM topics, and vice versa. Given two paper sets $X$ from HDBSCAN and $Y$ from LlooM, the Jaccard similarity is defined as $\text{Jaccard}(X, Y) = \frac{|X \cap Y|}{|X \cup Y|}$. We select the top-1 Jaccard score for each cluster, representing its highest similarity with any topic from the other method. Best-matching topic pairs are shown in Table~\ref{tab:cluster_alignment_hdbscan_to_lloom} and Table~\ref{tab:cluster_alignment_lloom_to_hdbscan} in the supplementary material, confirming that major topics identified by name also show substantial paper overlap.

\paragraph{Trend Alignment Between HDBSCAN and LlooM} Table~\ref{tab:trend_comparison_hdbscan_lloom_combined} summarizes trend agreement between the two clustering approaches across matched topics in the ACL and arXiv datasets.

\textbf{In the ACL dataset}, 4 out of 6 matched topics (67\%) share the same trend direction between HDBSCAN and LlooM based on Kendall's Tau. However, only half of these (33\% overall) are also aligned in trend significance. Spearman $\rho$ values for trend shape similarity are generally moderate to low and mostly not significant, with the exception of \textit{Security}/\textit{Security Risks}.

\textbf{The arXiv dataset} shows strong overall agreement between the HDBSCAN and LlooM clustering approaches. Most topics demonstrate matching trend directions according to Kendall's Tau (6 out of 8 topics, 75\%), and the quarter-to-quarter fluctuations also correlate strongly, as reflected by high Spearman $\rho$ values. For instance, \textit{Multimodality} ($\rho=0.86$) and \textit{Multilinguality} ($\rho=0.92$) achieve strong and statistically significant trend similarity across methods. Nonetheless, slight divergences remain: although trend directions often align, significance levels or trend shapes occasionally differ. For example, \textit{Hallucination} trends upward in both methods but is statistically significant only in HDBSCAN, with a moderate but insignificant trend similarity ($\rho=0.55$). One notable case of strong disagreement is \textit{Long Context}, which displays both opposing trend directions and poor trend shape similarity ($\rho=-0.22$). These stronger results may reflect the greater reliability of the arXiv dataset due to its larger size, in contrast to the smaller ACL sample.

\paragraph{Sources of Divergence Between HDBSCAN and LlooM} Although trend agreement between HDBSCAN and LlooM is stronger in the arXiv dataset compared to ACL, it is still not fully consistent across all topics. These differences likely reflect methodological differences between the clustering pipelines. To better understand this, we compare HDBSCAN+BERTopic and LlooM in Table~\ref{tab:clustering_comparison_table} across both datasets. We report overall clustering characteristics (e.g., number of topics) and alignment metrics: (i) average top-1 Jaccard scores across clusters for topic-level similarity, and (ii) Adjusted Mutual Information (AMI) for structural agreement.\footnote{AMI compares how often data points are grouped similarly, while adjusting for the similarity that would be expected by random chance. To account for LlooM's multi-topic assignments, we compute a shuffle-based baseline: for each paper, we randomly select one of its LlooM topics and compare it to the HDBSCAN label. This process is repeated over 10 runs, and we report the mean and standard deviation.}


\begin{table}
    \centering
    \caption{Comparison of HDBSCAN and LlooM clustering methods across key metrics.}
    \label{tab:clustering_comparison_table}
    \renewcommand{\arraystretch}{1.1}
    \small
    \begin{tabular}{l l l}
        \toprule
        \textbf{Metric} & \textbf{HDBSCAN+BERTopic} & \textbf{LlooM} \\
        \midrule
        \# of Topics & 7 (ACL), 15 (arXiv) & 13 (ACL), 15 (arXiv) \\
        \% of Papers Assigned & 86.9\% (ACL), 85.9\% (arXiv) & 92.5\% (ACL), 93.5\% (arXiv) \\
        Avg. Topics per Paper & 1 & 1.5 (ACL), 1.8 (arXiv) \\
        Avg. Jaccard Overlap (top-1) & 0.313 (ACL), 0.201 (arXiv) & 0.239 (ACL), 0.244 (arXiv) \\
        AMI Shuffled & \multicolumn{2}{c}{0.2285 $\pm$ 0.0085 (ACL), 0.2206 $\pm$ 0.0028 (arXiv)} \\
        \bottomrule
    \end{tabular}
\end{table}


Compared to HDBSCAN, LlooM achieves slightly higher coverage of papers across both datasets due to multi-topic assignment. Moreover, the topic-level Jaccard overlaps between methods are only moderate (0.313 for ACL, 0.244 for arXiv), and overall structural alignment, as measured by AMI, remains relatively low (0.229 for ACL, 0.221 for arXiv). 

These results show that LlooM and HDBSCAN identify similar broad limitation areas but organize papers differently at a finer-grained level. This is supported by Tables~\ref{tab:cluster_alignment_hdbscan_to_lloom} and \ref{tab:cluster_alignment_lloom_to_hdbscan} in the supplementary material, which report the closest topic alignments between methods. Large areas such as \textit{Reasoning}, \textit{Hallucination}, \textit{Security Risks}, and \textit{Bias and Fairness} appear relatively stable and align well across methods and datasets ($J > 0.4$), whereas smaller topics, including \textit{Overconfidence} and \textit{Prompt Sensitivity}, are often absorbed into broader categories such as \textit{Hallucination}. This reflects differences in clustering granularity: LlooM splits topics into overlapping subcategories, while HDBSCAN merges related issues into broader clusters. For example, HDBSCAN merges LlooM’s \textit{Security Risks}, \textit{Privacy Risks}, and \textit{Trustworthiness} into a single \textit{Security} cluster.

While broad limitation areas and general trend directions are reliably identified across HDBSCAN and LlooM, finer-grained topic structures and trend significance vary depending on the clustering method, which highlights that clustering choice impacts the interpretation of limitation trends. We return to these methodological considerations in Section~\ref{sec:limitations-future-work}.

\subsubsection{Human Evaluation}

\ak{While a full gold-standard clustering is not scalable, we assess cluster validity through stratified human re-annotation of cluster assignments and measure both inter-annotator and human–method agreement for both HDBSCAN and LlooM. 

For both ACL and arXiv, we construct a stratified human evaluation dataset (50 papers for ACL and 60 for arXiv) by sampling papers with approximately uniform per-topic sample sizes, including outliers, stratified with respect to HDBSCAN clusters. Each sampled paper is annotated with both its HDBSCAN and LlooM topic assignments. Because LlooM allows multi-label assignments, LlooM topic frequencies in the evaluation dataset are not explicitly controlled during sampling; however, all LlooM topics identified in the corpus are represented.\footnote{Except for the broad \textit{Trustworthiness} cluster in arXiv, which was excluded from human annotation due to substantial overlap with other topics.}

The dataset was annotated by a professor (machine learning) and a PhD student (NLP). Annotators were provided with each paper's title, extracted evidence, and keyphrases, corresponding to the input used by the clustering methods (as described in Section~\ref{sec:clustering}). For HDBSCAN, annotators selected a single topic per paper. For LlooM, annotators could assign multiple topics per paper, reflecting its multi-label clustering setup. Annotators were shown representative BERTopic keyword lists and LlooM assignment prompts to convey cluster semantics.

We measure both inter-annotator agreement and human–method agreement, using Cohen's $\kappa$ for HDBSCAN, which produces single-label assignments, and boot-F1 \citep{marchal2022establishing} for LlooM to account for chance agreement in the multi-label setting.}

\begin{table}[h]
\centering
\caption{\ak{Inter-annotator and human--method agreement for HDBSCAN and LlooM on ACL and arXiv datasets. Cohen's $\kappa$ is used for HDBSCAN (single-label), and boot-F1 is used for LlooM (multi-label).}}
\label{tab:human-eval-agreement}
\renewcommand{\arraystretch}{1.1}
\small
\begin{tabular}{l l c c}
\toprule
\ak{\textbf{Dataset}} & \ak{\textbf{Method}} & \ak{\textbf{Inter-Annotator}} & \ak{\textbf{Human--Method}} \\
\midrule
\ak{ACL} & \ak{HDBSCAN} & \ak{$\kappa = 0.79$} & \ak{$\kappa = 0.74$} \\
\ak{ACL} & \ak{LlooM} & \ak{boot-F1 $= 0.70$} & \ak{boot-F1 $= 0.61$} \\
\midrule
\ak{arXiv} & \ak{HDBSCAN} & \ak{$\kappa = 0.71$} & \ak{$\kappa = 0.55$--$0.59$} \\
\ak{arXiv} & \ak{LlooM} & \ak{boot-F1 $= 0.53$} & \ak{boot-F1 $= 0.49$--$0.51$} \\
\bottomrule
\end{tabular}
\end{table}

\ak{Table~\ref{tab:human-eval-agreement} shows substantial inter-annotator agreement for both datasets ($\kappa=0.71$–$0.79$ for HDBSCAN and boot-F1 $=0.53$–$0.70$ for LlooM), indicating consistent human judgments across both single-label and multi-label settings. Human–method agreement is also substantial for ACL ($\kappa=0.74$, boot-F1 $=0.61$) and moderate for arXiv ($\kappa=0.55$–$0.59$, boot-F1 $=0.49$–$0.51$), indicating that human annotations broadly correspond to the topics produced by the clustering methods.}

%% file: structure/5-discussion.tex
\section{Discussion and Conclusion}\label{sec:discussion}

Based on the detailed results in \ref{sec:results}, we conclude four major findings.

\paragraph{1. LLM limitation research grew rapidly in 2022-2025, outpacing even the overall growth of LLM research.} 
LLM research now dominates NLP and increasingly influences neighboring fields: by the end of 2024, over 75\% of ACL papers and more than 30\% of arXiv submissions across cs.CL, cs.AI, cs.LG, and cs.CV focus on LLMs, with growth continuing into 2025. Within arXiv, LLM engagement in cs.CL closely mirrors ACL trends (reaching ~80\%), while areas like cs.CV and cs.LG remain below 20\% but show steady growth. While only about 10\% of LLM-related papers in early 2022 focused on limitations, the fraction increased to about one third by 2025. This growth in \lllm\ research may indicate a maturation of LLM research: the very early enthusiasm for LLMs and their capabilities, driven by the public deployment of systems like ChatGPT, is now increasingly accompanied with a more critical perspective towards limitations \citep{guo2023evaluating}. Meta-analyses confirm this trend, showing a sharp rise in evaluation-focused papers from 2020 to 2023~\citep{chang2024survey}.

\paragraph{2. Within \lllm\ research, reasoning is the most frequent topic, but research is diverse}

\textit{Reasoning} is the most frequent limitation topic in ACL across both clustering approaches and remains among the top concerns in arXiv, ranking third in HDBSCAN and second (after \textit{Trustworthiness}) in LlooM. Other prominent topics include \textit{Generalization}, \textit{Hallucination}, \textit{Bias}, and \textit{Security}.

Beyond these, LLM limitation research is notably diverse. Our clustering analyses (HDBSCAN and LlooM) reveal a broad spectrum of concerns, ranging from \textit{code generation} and \textit{benchmark contamination} to \textit{prompt sensitivity} and \textit{long context}. This breadth reflects the current state of LLM limitation research: a fast-growing, methodologically diverse field still defining its major challenges. Additionally, as shown in Section~\ref{sec:topic-co-occurrence} in the supplementary materials, many papers address multiple limitations simultaneously, reflecting the complexity of emerging concerns.

\paragraph{3. The distribution of limitations appears relatively stable in the ACL dataset, whereas the arXiv dataset shows a rise in concern for topics related to safety and controllability}

This contrast is nuanced and is reflected in two key trends, discussed below.

\textit{3.1 Emerging trends in LLLMs research.} Our trend analysis reveals mixed dynamics within \lllm\ research over the studied time period. Safety and controllability concerns (e.g., \textit{Security Risks}, \textit{Alignment Limitations}, \textit{Knowledge Editing}, \textit{Hallucination}), model capacity advances (e.g., \textit{Long Context} in ACL), and \textit{Multimodality} generally rise over time. In contrast, topics like \textit{Bias and Fairness} decline, while others remain flat. Notably, we observe a shift around 2023-Q2, following the release of models like ChatGPT. After this point, early fluctuations diminish, topics that had been growing continue at a steadier pace, and the decline in certain areas becomes more pronounced.

These trends align with shifting priorities in the LLM community. The growing attention to \textit{alignment} and \textit{security} reflects their increasingly central role in both training and evaluation of LLMs. Though still relatively new and unsettled \citep{shen2023large}, these concerns became prominent in 2022 with the rise of Reinforcement Learning from Human Feedback (RLHF), which is now foundational in the training pipelines of major models \citep{wang2024comprehensive}. Yet ensuring safety without compromising performance remains an open challenge \citep{qi2023fine, tan2025equilibrate}, making this an active and fast-moving research area.

\textit{3.2. Limitations persist as applications expand.} As LLMs are deployed in high-stakes domains, interest in \textit{hallucination} and \textit{knowledge editing} is growing due to the increasing demand for factual accuracy and controllability. These remain deeply challenging: hallucination is increasingly seen as an inherent property of LLMs, rooted in model architecture itself \citep{banerjee2024llms, xu2024hallucination}. 


Finally, these challenges grow as models move beyond text. The rise of \textit{multimodality}-related limitations suggests that LLMs not only inherit existing issues but also encounter new ones with inputs like images and audio \citep{zhang2024mm}. This trend likely reflects growing interest caused by the release of GPT-4V \citep{achiam2023gpt}, LLaVA \citep{liu2023visual}, and other vision-language models in mid-2023. These and other trends discussed above coincide with broader shifts \ak{already noted in previous studies of the arXiv corpus: in particular, previous work finds that} from early 2023 to late 2024, top-cited LLM papers increasingly came from cs.CV and cs.LG, with cs.CL seeing a relative decline \citep{leiter2024nllg}. Similarly, authorship diversified, with many newcomers from computer vision, security, and software engineering \citep{movva2023topics}.

\paragraph{4. Despite methodological differences, HDBSCAN and LlooM identify overlapping high-frequency topics (e.g., \textit{Reasoning}, \textit{Hallucination}, \textit{Security Risks}) and similar trend patterns, supporting the stability of the main findings.}

We validate our results by comparing HDBSCAN+BERTopic (single-topic, density-based) with LlooM (LLM-based, multi-topic). Despite their differences, both methods recover the same dominant topics, especially \textit{Reasoning}, \textit{Hallucination}, and \textit{Security Risks}, with strong agreement in topic composition and trend trajectories. And although smaller topics (e.g., \textit{Prompt Sensitivity}) and trend significance can vary, the main trends 
appear stable across methods.

\subsection*{Limitations and Future Work}\label{sec:limitations-future-work}

While our analysis involves multiple datasets and clustering approaches, several methodological and temporal constraints should be kept in mind when interpreting the results:

\begin{itemize}
    \item Although Llama-3.1-70B performs near human level in annotating limitation relevance, it still slightly lags, particularly in extracting supporting evidence, possibly leading to missed or incorrect information. However, as noted in Section~\ref{sec:model_eval_results}, human annotators also overlooked some cases, suggesting that some level of imprecision is inherent to the task.
    \item Both of our clustering approaches are prone to some instability. LlooM can be variable due to its reliance on LLM outputs, a limitation noted by its authors as well \citep{lam2024concept}. HDBSCAN+BERTopic can also vary across runs because UMAP is stochastic and sensitive to embedding changes. While high-level patterns are generally stable, topic composition and temporal trends may shift slightly. We mitigate these issues by validating results across both methods.
    \item While we adopt a broad definition of LLMs in both automated and human annotation (including transformer-based, foundational, multimodal models), this scope may still introduce bias. \ak{In particular, our reliance on abstract- and keyword-level coverage and on the extraction of explicitly stated limitation evidence may undercount limitations discussed using non-standard terminology,} or newly emerging terms (e.g., in fields such as CV).
    \item Our trend analysis for the arXiv dataset includes data up to early 2025. Therefore, apparent declines or plateaus in the latest quarter should be interpreted with caution. We also exclude data prior to 2022, even though interest in limitations of smaller-scale LMs had already been rising since the introduction of models like BERT in 2018 \citep{rogers2021primer, zhao2023survey}.
    \item \ak{Given our automated survey design, trend analyses reflect the prevalence of limitation-related discussions in research rather than the severity of the discussed limitations. Although sustained research interest can reflect practical relevance, changes in topic prevalence should primarily be interpreted as shifts in research attention.}
    \item \ak{This survey provides a large-scale overview of \lllm\ research rather than a fine-grained taxonomy; while the identified categories do not capture detailed limitation subtypes, the topics are high-level to reflect dominant and stable themes across methods and datasets.}
\end{itemize}

Future work can extend these findings in several ways. First, limitation topics can be decomposed into finer subcategories, such as specific reasoning types or bias forms, using hierarchical or agglomerative clustering. Second, expanding the analysis to earlier years, especially after BERT's introduction in 2018, could show how concerns about smaller PLMs evolved with model scaling. \ak{Third, the automatic pipeline can be adapted to incorporate newly published papers, enabling regular updates of the dataset over time.}

%% file: structure/appendix.tex
\clearpage
\appendix

\section*{Supplementary Material}\label{sec:appendix}

\section{Keyword List}
\label{sec:keyword_list}

The final keyword list used to filter LLM-focused papers is provided in Table~\ref{tab:keyword_list}.

\begin{table}[h]
    \centering
    \small
    \caption{Final list of keywords sorted by N-grams}
    \label{tab:keyword_list}
    \renewcommand{\arraystretch}{1.1}
    \begin{tabular}{p{4cm} p{4cm} p{4cm}}  
        \toprule
        \emph{cot} & \emph{gpt} & \emph{api} \\
        \emph{rag} & \emph{judge} & \emph{chat} \\
        \emph{llms} & \emph{dpo} & \emph{mllms} \\
        \emph{llm} & \emph{lora} & \emph{hallucination} \\
        \emph{jailbreak} & \emph{speculative} & \emph{self consistency} \\
        \emph{agent} & \emph{cot prompting} & \emph{llm agents} \\
        \emph{model editing} & \emph{self correction} & \emph{prompting} \\
        \emph{self reflection} & \emph{function calling} & \emph{language agents} \\
        \emph{hallucination detection} & \emph{preference learning} & \emph{long context} \\
        \emph{language models} & \emph{data contamination} & \emph{injection attacks} \\
        \emph{instruction tuned} & \emph{prompt engineering} & \emph{jailbreak attack} \\
        \emph{preference alignment} & \emph{knowledge editing} & \emph{text watermarking} \\
        \emph{prompt optimization} & \emph{self evaluation} & \emph{instruction tuning} \\
        \emph{tree of thoughts} & \emph{evaluating llms} & \emph{multi agent framework} \\
        \emph{benchmarking llms} & \emph{retrieval augmented generation} & \emph{direct preference optimization} \\
        \emph{multimodal llms} & \emph{commonsense reasoning} & \emph{chain of thought reasoning} \\
        \emph{chain of thought prompting} & \emph{multi agent collaboration} & \emph{augmented llms} \\
        \emph{generated text detection} & \emph{jailbreaking attacks} & \emph{jailbreak attacks} \\
        \emph{prompting techniques} & & \\
        \bottomrule
    \end{tabular}
\end{table}

\begin{table*}[h]
    \centering
    \caption{Representative examples from the human rating annotation task, with brief explanation for abstracts' ratings on the 0–5 scale (defined in Table~\ref{tab:annotation_labels}). In higher-rated papers (3–5), evidence of \lllm\ is highlighted in bold.}
    \label{tab:annotated_examples}
    \renewcommand{\arraystretch}{1.1}
    \begin{tabularx}{\textwidth}{c >{\raggedright}p{9cm} X}
        \toprule
        \textbf{Rating} & \textbf{Abstract} & \textbf{Explanation} \\
        \midrule
        0 & \enquote{\small Multi-query attention (MQA), which only uses a single key-value head, drastically speeds up decoder inference. However, MQA can lead to quality degradation. [...]} \citep{ainslie2023gqa}, EMNLP 2023 & \small No mention of LLMs or their limitations. \\ 
        \midrule
        1 & \enquote{\small We introduce the framework of \enquote{social learning} in the context of LLMs, whereby models share knowledge with each other in a privacy-aware manner using natural language. [...]} \citep{mohtashami2023social}, arXiv, 2023 & \small Mentions LLMs but does not discuss any limitations. \\ 
        \midrule
        2 & \enquote{\small [...] \textbf{LLMs... often display a considerable level of overconfidence even when the question does not have a definitive answer.} [...] we propose a novel and scalable self-alignment method to utilize the LLM itself to enhance its response-ability to different types of unknown questions. [...]} \citep{deng2024gotcha}, EMNLP 2024 & \small Briefly identifies overconfidence as a limitation; used as motivation for the method. \\ 
        \midrule
        3 & \enquote{\small [...] \textbf{it remains difficult to distinguish effects of statistical correlation from more systematic logical reasoning grounded on the understanding of real world [in PLMs]}. [...] We find that models are consistently able to override real-world knowledge in counterfactual scenarios... however, we also find that \textbf{for most models this effect appears largely to be driven by simple lexical cues}. [...]} \citep{li2023counterfactual}, ACL 2023 & \small Discusses reasoning limitations in moderate detail, but not as the main focus. \\ 
        \midrule
        4 & \enquote{\small [...] we aim to assess the performance of OpenAI's newest model, GPT-4V(ision), specifically in the realm of multimodal medical diagnosis. [...] \textbf{While GPT-4V demonstrates proficiency in distinguishing between medical image modalities and anatomy, it faces significant challenges in disease diagnosis and generating comprehensive reports.} [...] while large multimodal models have made significant advancements in computer vision and NLP, \textbf{it remains far from being used to effectively support real-world medical applications and clinical decision-making.} [...]} \citep{wu2023can}, arXiv, 2023 & \small Focuses extensively on GPT-4V's limitations in medical applications, but also explores strengths of the model. \\ 
        \midrule
        5 & \enquote{\small We analyze the capabilities of Transformer LMs on learning discrete algorithms. [...] \textbf{We observe that the compositional capabilities of state-of-the-art Transformer language models are very limited and sample-wise scale worse than relearning all sub-tasks for a new algorithmic composition. We also present a theorem in complexity theory, showing that gradient descent on memorizing feedforward models can be exponentially data inefficient.}} \citep{thomm2024limits}, arXiv, 2024 & \small Entirely focused on identifying and analyzing LLM limitations in algorithmic learning. \\ 
        \bottomrule
    \end{tabularx}
\end{table*}

\begin{figure}
    \centering
    \includegraphics[width=0.86\textwidth]{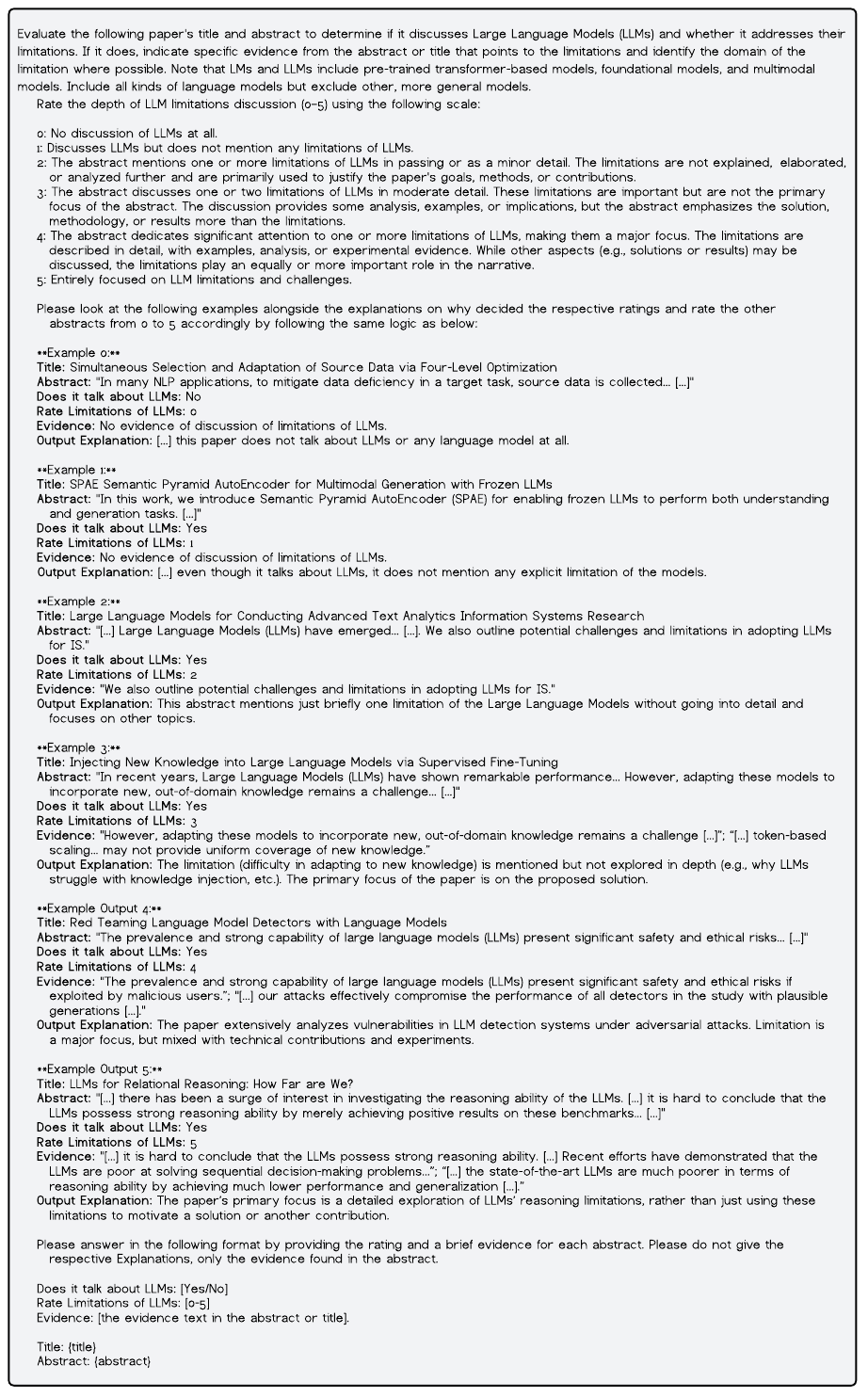}
    \caption{Prompt 3 used for LLM-based classification and evidence extraction (as discussed in \ref{sec:arxiv_categories_analysis}). Abstract texts and evidence are shortened for brevity.}
    \label{fig:evaluation_prompt}
\end{figure}

\clearpage 
\section{Clustering Details}\label{sec:clustering_details}

For HDBSCAN+BERTopic, we perform a grid search over UMAP, HDBSCAN and BERTopic to produce meaningful clusters while minimizing the outliers. The final configurations are as follows: 

\begin{itemize}
    \item for ACL, UMAP is set to \texttt{n\_neighbors} = \texttt{25}, \texttt{n\_components} = \texttt{10}, \texttt{min\_dist} = \texttt{0.0}; HDBSCAN to \texttt{min\_cluster\_size} = \texttt{25}, \texttt{min\_samples} = \texttt{10}, and BERTopic to \texttt{min\_topic\_size} = \texttt{10}.
    \item For ArXiv, UMAP is adjusted to \texttt{n\_neighbors} = \texttt{15}, \texttt{n\_components} = \texttt{5}, \texttt{min\_dist} = \texttt{0.05}, while HDBSCAN and BERTopic are set to \texttt{min\_cluster\_size} = \texttt{40} and \texttt{min\_topic\_size} = \texttt{25}.
\end{itemize}

Our initial runs revealed that many outliers were not true noise but rather positioned between clusters, causing HDBSCAN (used within BERTopic) to misclassify them. To address this, we implement a distance-based outlier reassignment strategy, where outliers identified by HDBSCAN and BERTopic are reassigned to their most probable topic based on the topic probability distribution, but only if they fall within a certain distance threshold to ensure proximity to an existing cluster.

\begin{figure}[h] 
    \centering
    \includegraphics[width=0.8\textwidth]{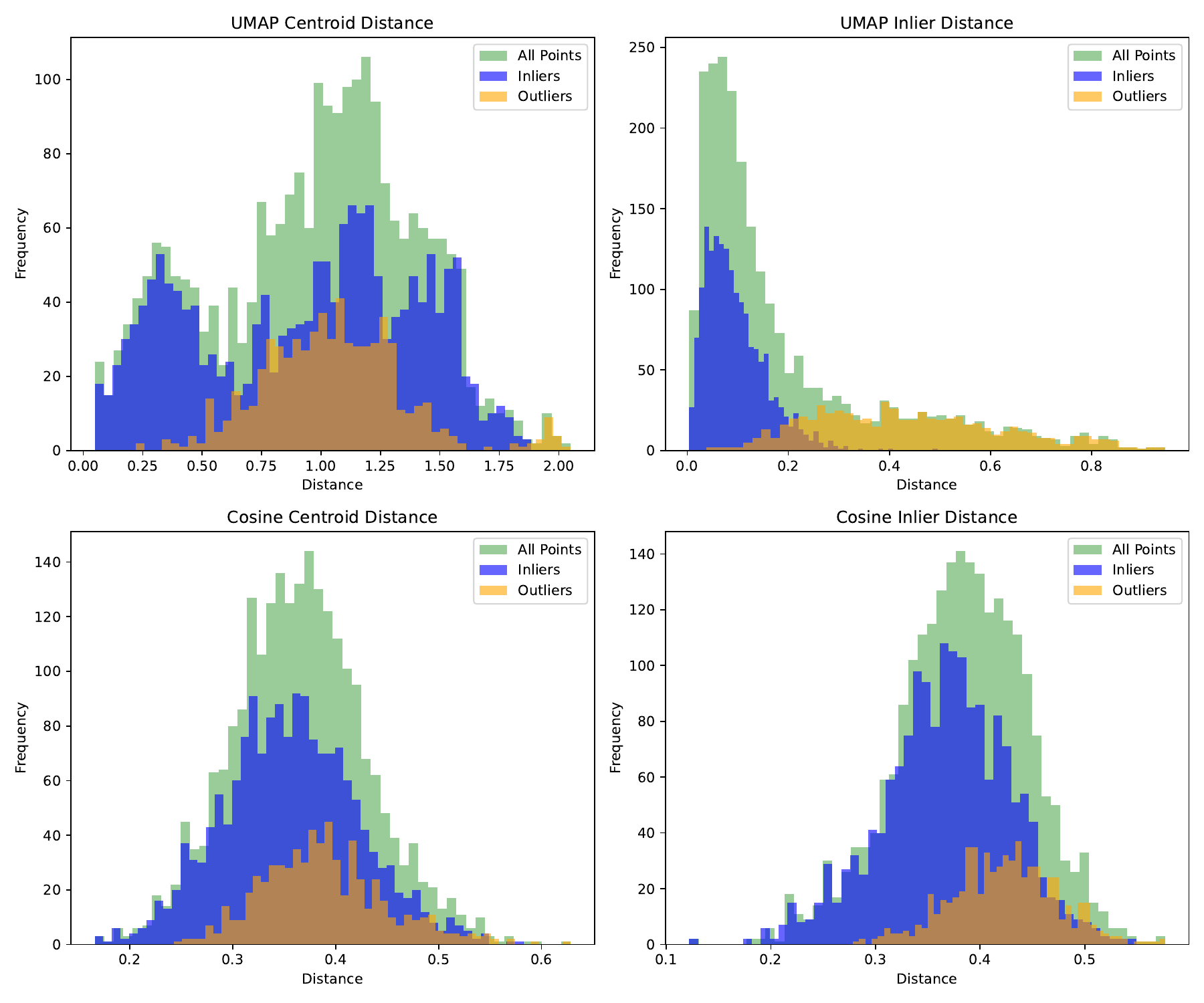}
    \caption{Histograms of distance metrics used for outlier detection in ACL dataset clustering. Each subplot visualizes distances between a paper’s embedding (based on evidence and keyphrases) and either a cluster centroid or nearest inlier, in both UMAP-reduced and original embedding spaces. These metrics help determine the threshold for reassigning outliers to the closest cluster.} 
    \label{fig:distances_histogram_acl}
\end{figure}

To determine this threshold, we test several distance metrics, namely 1) the Euclidean distance to the closest cluster centroid in UMAP space, 2) the Euclidean distance to the closest inlier in UMAP space, 3) the cosine distance to the closest cluster centroid in the original embedding space, and 4) the cosine distance to the closest inlier in the original embedding space. As shown in Figure~\ref{fig:distances_histogram_acl}, the UMAP Inlier Distance provides the clearest separation between inliers and outliers, such that we selected it for subsequent analysis. We set the outlier threshold at the dip between the two peaks, using a value of 0.3. 

\begin{table}[h]
\centering
\caption{Prompt descriptions for each cluster in the ACL dataset identified by LlooM.}
\label{tab:acl-cluster-prompts}
\small
\begin{tabularx}{\textwidth}{@{}lX@{}}
\toprule
\textbf{Cluster} & \textbf{Prompt} \\
\midrule
Reasoning & Does the text example highlight limitations of large language models in performing reasoning or inference tasks, such as multi-hop reasoning or deductive logic? \\
Generalization & Does the text example highlight limitations in a model's ability to understand context or generalize effectively across different domains and scenarios? \\
Knowledge Editing & Does the text discuss limitations in storing, encoding, or updating knowledge in large language models, including factual inaccuracies or difficulties in modifying learned information? \\
Hallucination & Does the text describe instances where language models generate factually incorrect information, describe non-existent content, or produce contextually unfaithful data? \\
Language and Cultural Limitations & Does the text example highlight limitations of large language models in handling multilingual tasks, language-specific or cross-cultural issues? \\
Bias and Fairness & Does the text address issues related to social biases or stereotyping within language models, including challenges in presence, amplification, and mitigation thereof? \\
Security Risks & Does this text example address the susceptibility of language models to adversarial attacks, including backdoor or jailbreak methods? \\
Multimodality & Does the text example address the struggles of large language models or multimodal models in integrating and reasoning across different modalities? \\
Long Context & Does the text example discuss challenges faced by large language models in handling long contexts or maintaining coherence over extended inputs? \\
Privacy Risks & Does the text example discuss privacy risks associated with language models, such as data leakage or memorization of sensitive information? \\
Computational Cost & Does the text example highlight the high computational cost or resource demands associated with using large language models? \\
Catastrophic Forgetting & Does the example discuss problems caused by models forgetting previously learned information during fine-tuning or continual learning? \\
Data Contamination & Does the text example discuss issues of training data containing test data, causing overestimation of model performance through memorization? \\
\bottomrule
\end{tabularx}
\end{table}

\begin{table}[h]
\centering
\caption{Prompt descriptions for each cluster in the arXiv dataset identified by LlooM.}
\label{tab:arxiv-cluster-prompts}
\small
\begin{tabularx}{\textwidth}{@{}lX@{}}
\toprule
\textbf{Cluster} & \textbf{Prompt} \\
\midrule
Trustworthiness & Does the text refer to concerns about the reliability or trustworthiness of outputs generated by language models? \\
Reasoning & Does the text example highlight limitations of large language models in performing reasoning or inference tasks, such as multi-hop reasoning or deductive logic? \\
Generalization & Does the text discuss the inability of language models to generalize across different tasks or inputs? \\
Long Context & Does this paper explore challenges faced by large language models (LLMs) in handling long context lengths, such as limitations in memory, performance, or understanding over extended inputs? \\
Bias and Fairness & Does the text describe how LLMs propagate biases, stereotypes, or misinformation, leading to potential societal harms? \\
Hallucination & Does the text mention inaccuracies or fabricated content in the outputs of large language models (also multimodal)? \\
Alignment Limitations & Does the text highlight any limitations or challenges in aligning large language models with human values or safety protocols? \\
Security Risks & Does the text address the security risks associated with adversarial attacks or exploits that can manipulate large language models? \\
Prompt Sensitivity & Does the text highlight how slight variations in prompts affect the performance of language models? \\
Multimodality & Does the text discuss difficulties faced by LLMs in effectively integrating or coordinating information across multiple modalities, such as text and images? \\
Language and Cultural Limitations & Does this text highlight struggles of Language Models (LLMs) in effectively handling low-resource languages? \\
Privacy Risks & Does the text discuss vulnerabilities related to the potential leakage of private or sensitive information by language models? \\
Knowledge Editing & Does the text discuss challenges in knowledge editing, such as knowledge distortion, struggles with updating specific knowledge types, or difficulty integrating new knowledge while maintaining coherence? \\
Overconfidence & Does the text describe challenges related to calibration or overconfidence in language models? \\
Catastrophic Forgetting & Does this text describe Large Language Models (LLMs) losing previously acquired knowledge when learning new information? \\
\bottomrule
\end{tabularx}
\end{table}

\begin{table}[h]
    \centering
    \caption{Relative percentage growth of LLM-normalized shares for limitation topics across ACL and arXiv datasets (LlooM clustering). Values are based on relative change between consecutive years.}
    \label{tab:relative_growth_acl_arxiv}
    \renewcommand{\arraystretch}{1.2}
    \small

    \begin{subtable}[t]{\textwidth}
        \centering
        \caption{ACL Dataset}
        \label{tab:relative_growth_acl}
        \begin{tabular}{l c c c}
            \toprule
            \textbf{Topic} & \textbf{Topic Shares (2022 → 2024)} & \textbf{→2023 (\%)} & \textbf{→2024 (\%)} \\
            \midrule
            Reasoning & 5.5 → 7.5 → 8.9 & 36.36\% & 18.67\% \\
            Generalization & 3.9 → 7.2 → 8.6 & 84.62\% & 19.44\% \\
            Knowledge Editing & 4.6 → 5.8 → 7.3 & 26.09\% & 25.86\% \\
            Hallucination & 3.2 → 5.2 → 7.2 & 62.50\% & 38.46\% \\
            Language \& Cultural Lim. & 2.8 → 3.8 → 4.1 & 35.71\% & 7.89\% \\
            Bias and Fairness & 2.3 → 3.9 → 3.7 & 69.57\% & -5.13\% \\
            Security Risks & 1.2 → 2.1 → 3.5 & 75.00\% & 66.67\% \\
            Multimodality & 1.2 → 1.5 → 3.5 & 25.00\% & 133.33\% \\
            Long Context & 1.1 → 1.3 → 2.7 & 18.18\% & 107.69\% \\
            Computational Cost & 0.6 → 1.1 → 1.4 & 83.33\% & 27.27\% \\
            Privacy Risks & 0.7 → 1.2 → 1.3 & 71.43\% & 8.33\% \\
            Catastrophic Forgetting & 0.6 → 0.5 → 1.2 & -16.67\% & 140.00\% \\
            Data Contamination & 0.4 → 0.4 → 0.9 & 0.00\% & 125.00\% \\
            \bottomrule
        \end{tabular}
    \end{subtable}

    \vspace{1em}

    \begin{subtable}[t]{\textwidth}
        \centering
        \caption{arXiv Dataset}
        \label{tab:relative_growth_arxiv}
        \begin{tabular}{l c c c c}
            \toprule
            \textbf{Topic} & \textbf{Topic Shares (2022 → 2025)} & \textbf{→2023 (\%)} & \textbf{→2024 (\%)} & \textbf{→2025 (\%)} \\
            \midrule
            Trustworthiness & 6.12 → 10.85 → 11.99 → 12.18 & 77.29\% & 10.51\% & 1.58\% \\
            Reasoning & 5.15 → 6.43 → 7.09 → 9.09 & 24.85\% & 10.26\% & 28.21\% \\
            Generalization & 5.88 → 5.99 → 5.83 → 5.95 & 1.87\% & -2.67\% & 2.06\% \\
            Hallucination & 1.28 → 4.14 → 4.64 → 4.40 & 223.44\% & 12.08\% & -5.17\% \\
            Alignment Limitations & 1.87 → 3.78 → 4.51 → 5.23 & 102.14\% & 19.31\% & 15.96\% \\
            Bias and Fairness & 3.73 → 4.14 → 4.25 → 4.12 & 10.99\% & 2.66\% & -3.06\% \\
            Security Risks & 1.04 → 2.74 → 3.78 → 3.68 & 163.46\% & 37.96\% & -2.65\% \\
            Multimodality & 0.86 → 1.19 → 2.29 → 2.70 & 38.37\% & 92.44\% & 17.90\% \\
            Prompt Sensitivity & 1.07 → 2.03 → 1.84 → 1.55 & 89.72\% & -9.36\% & -15.76\% \\
            Language \& Cultural Lim. & 0.93 → 1.27 → 1.46 → 1.68 & 36.56\% & 14.96\% & 15.07\% \\
            Long Context & 0.69 → 1.02 → 1.26 → 1.63 & 47.83\% & 23.53\% & 29.37\% \\
            Privacy Risks & 0.83 → 1.03 → 1.11 → 1.23 & 24.10\% & 7.77\% & 10.81\% \\
            Knowledge Editing & 0.45 → 0.86 → 1.06 → 1.12 & 91.11\% & 23.26\% & 5.66\% \\
            Overconfidence & 0.45 → 0.83 → 0.96 → 0.96 & 84.44\% & 15.66\% & 0.00\% \\
            Catastrophic Forgetting & 0.62 → 0.54 → 0.76 → 0.59 & -12.90\% & 40.74\% & -22.37\% \\
            \bottomrule
        \end{tabular}
    \end{subtable}
\end{table}

\begin{figure}[h]
    \centering

    \begin{subfigure}[t]{0.9\textwidth}
        \includegraphics[width=\textwidth]{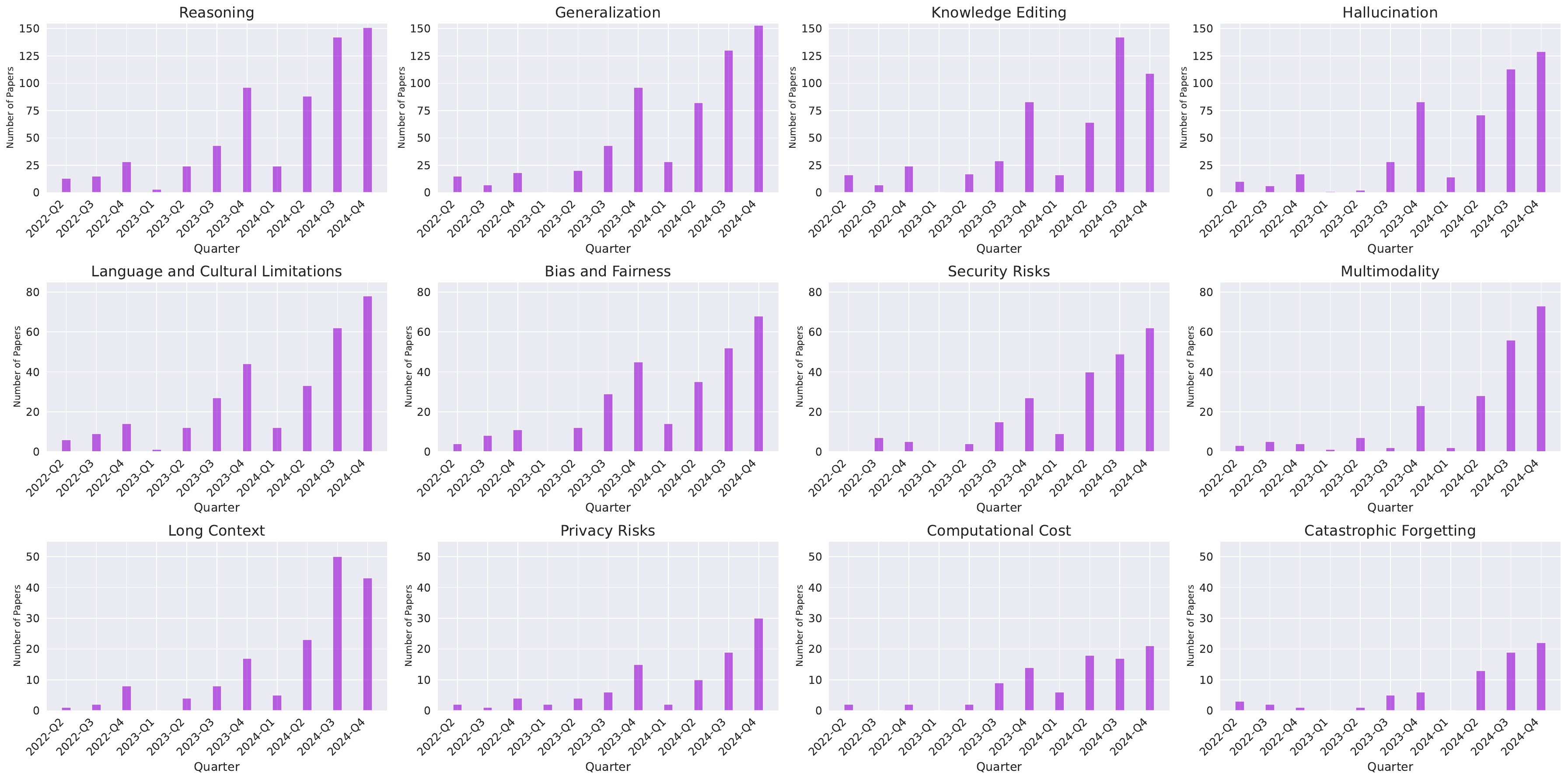}
        \caption{Absolute counts of papers discussing limitation topics in ACL.}
        \label{fig:lloom-acl-absolute-counts}
    \end{subfigure}
    \hfill
    \begin{subfigure}[t]{0.9\textwidth}
        \includegraphics[width=\textwidth]{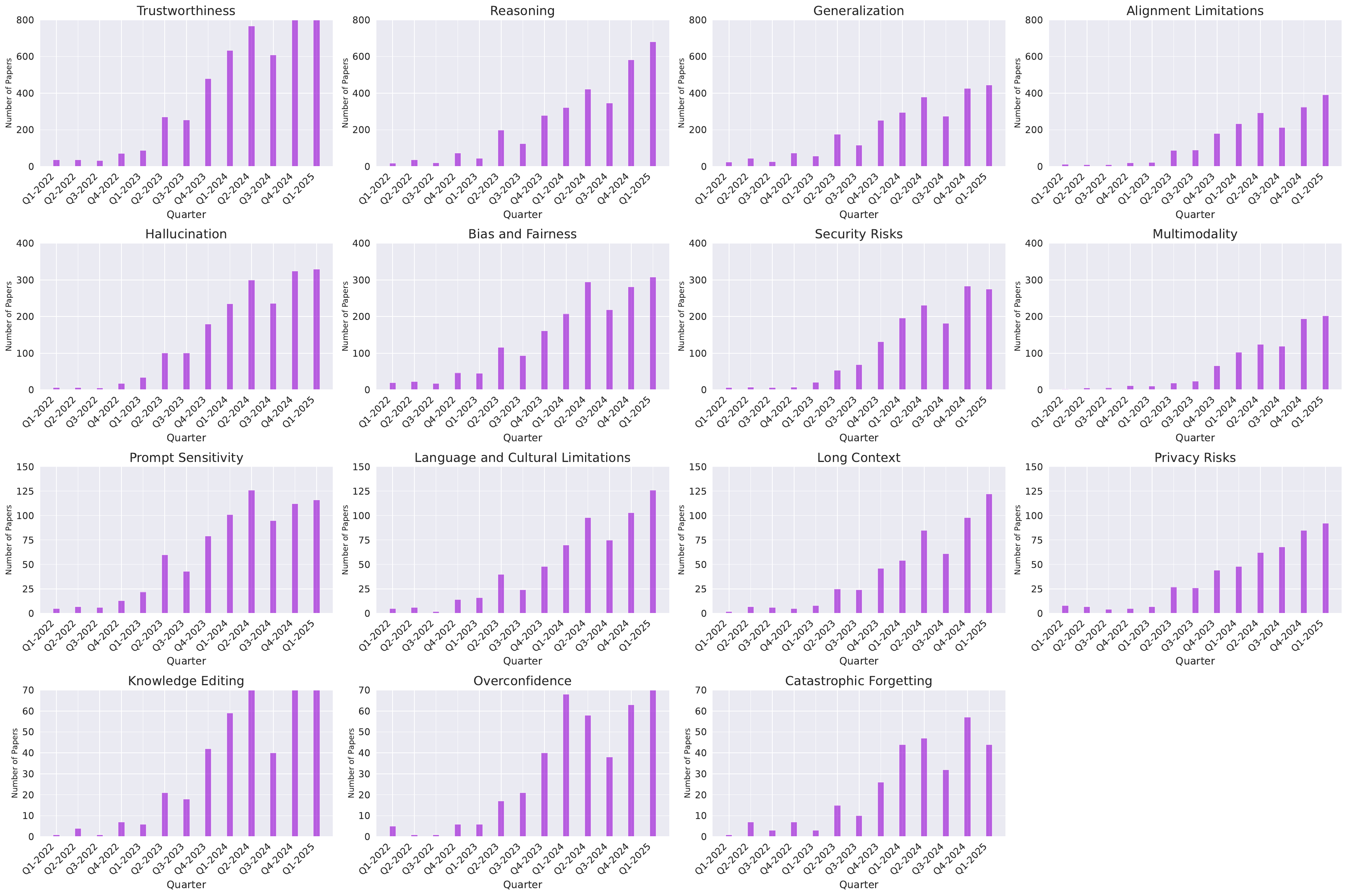}
        \caption{Absolute counts of papers discussing limitation topics in arXiv.}
        \label{fig:lloom-arxiv-absolute-counts}
    \end{subfigure}

    \caption{Absolute counts of papers discussing limitation topics in ACL and arXiv, as identified by LlooM clustering approach.}
    \label{fig:acl-arxiv-absolute-counts}
\end{figure}

\begin{figure}[h] 
    \centering
    \includegraphics[width=\textwidth]{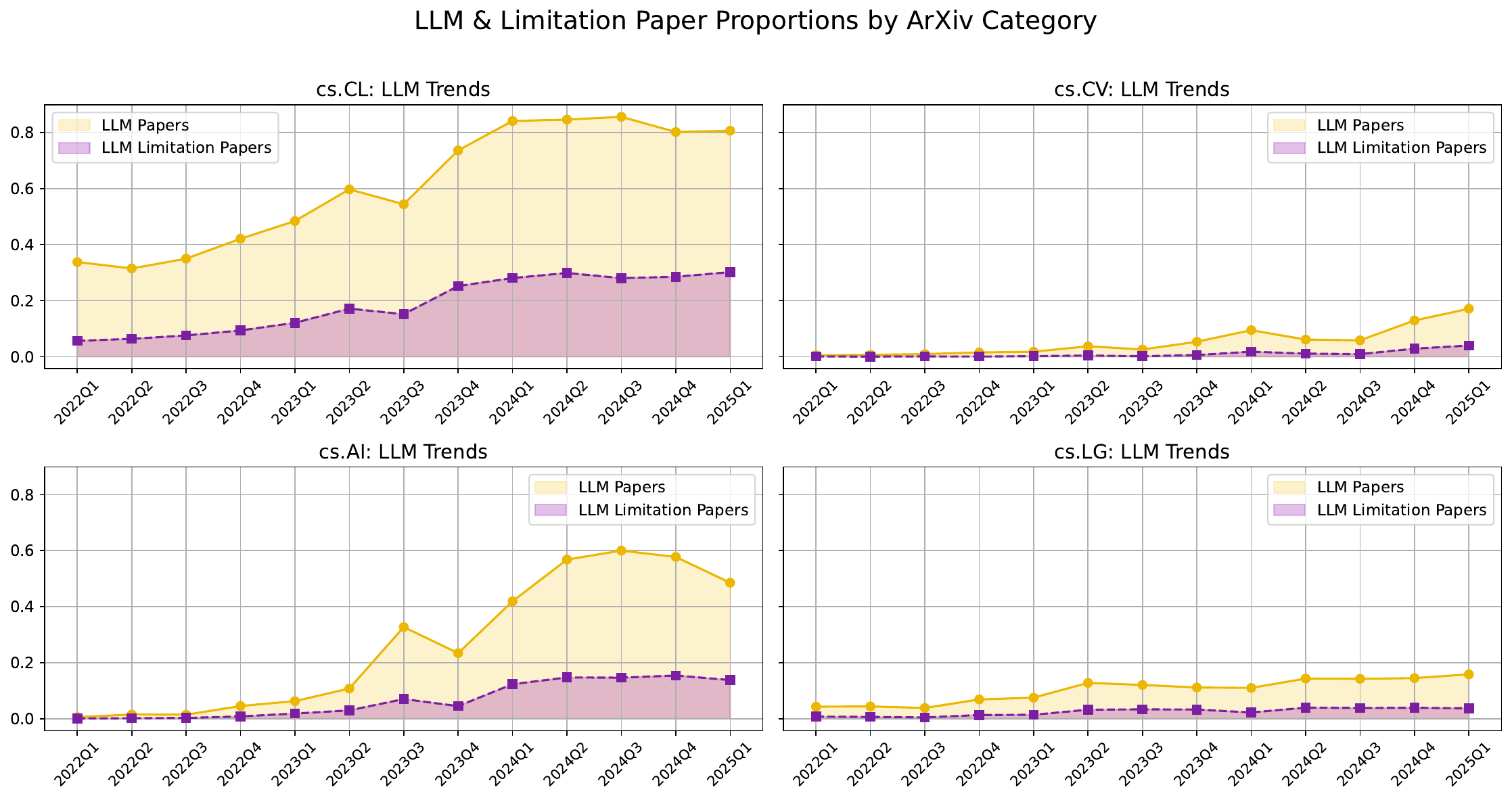}
    \caption{Trends in LLM and LLM limitation research in arXiv over time, relative to all crawled papers, based on the main category only and broken down by category.} 
    \label{fig:llm-lllms-trends-arxiv-categories}
\end{figure}

\begin{figure}[h] 
    \centering
    \includegraphics[width=0.9\textwidth]{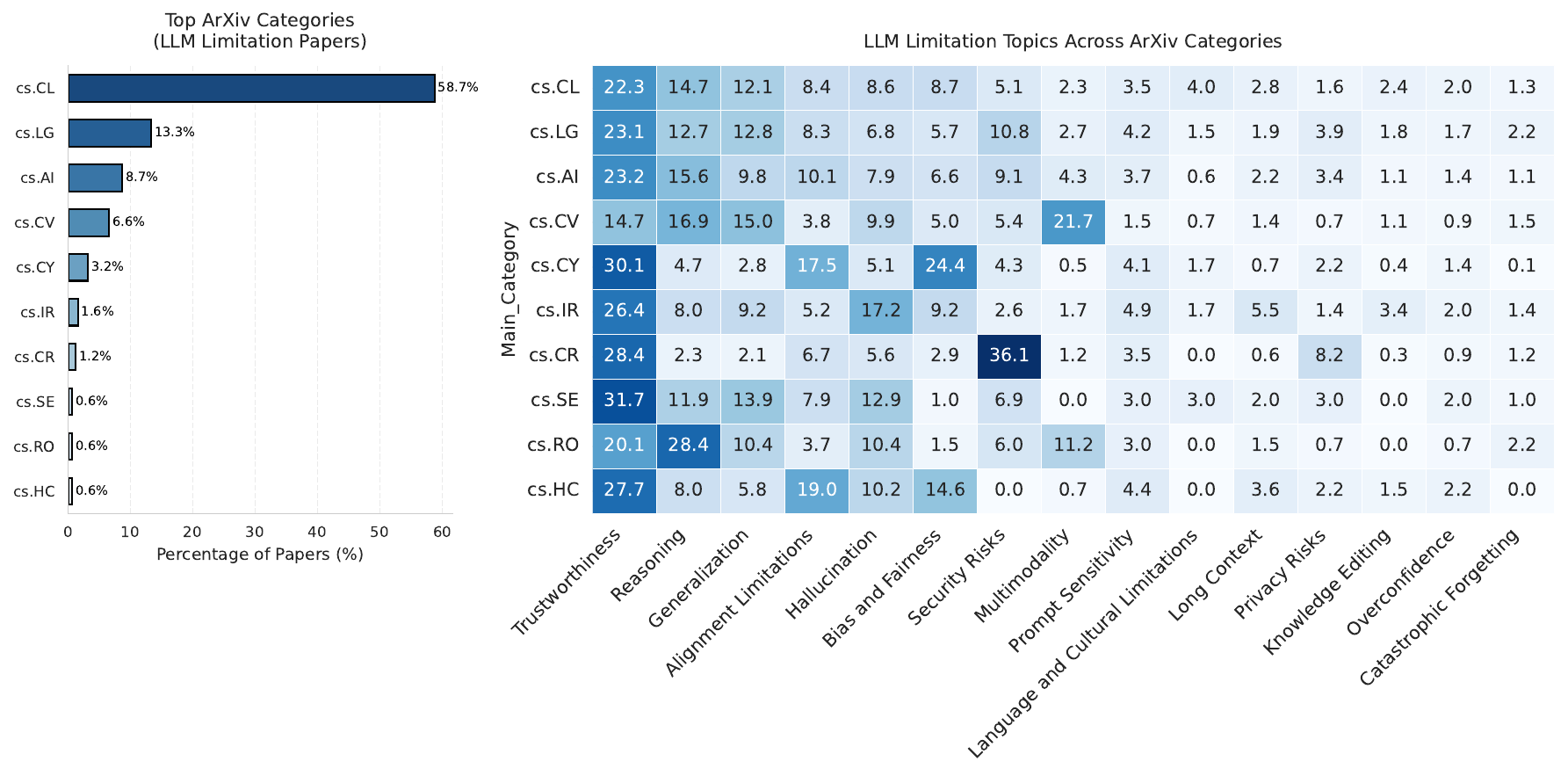}
    \caption{Top nine arXiv (main) categories discussing \lllm\ (left), and distribution of limitation topics within each category (right), shown as the percentage of papers in each category that mention a given limitation.} 
    \label{fig:arxiv-top-categories-lloom}
\end{figure}

\begin{table*}[t]
\centering
\footnotesize
\setlength{\tabcolsep}{6pt}
\renewcommand{\arraystretch}{1.0}

\caption{\ak{Top 20 keywords per LLM limitation topic, identified using HDBSCAN with BERTopic in the ACL Anthology dataset; keywords are extracted via TF-IDF and ranked by marginal document coverage (\%).}}
\label{tab:cluster_keywords_acl_long}

\begin{tabularx}{\textwidth}{@{}l X@{}}
\toprule
\textbf{\ak{Cluster}} & \textbf{\ak{Top 20 keywords (marginal document coverage})} \\
\midrule

\textbf{\ak{Reasoning}} &
\textit{\ak{reasoning (27\%);}} 
\textit{\ak{understand (16\%);}} 
\textit{\ak{benchmark (11\%);}}
\textit{\ak{logical (6\%);}} 
\textit{\ak{multilingual (5\%);}} 
\textit{\ak{multimodal (4\%);}}
\textit{\ak{visual (4\%);}} 
\textit{\ak{comprehension (4\%);}} 
\textit{\ak{semantic (4\%);}}
\textit{\ak{prompt (4\%);}} 
\textit{\ak{concept (4\%);}} 
\textit{\ak{linguistic (4\%);}}
\textit{\ak{image (3\%);}} 
\textit{\ak{contamination (3\%);}} 
\textit{\ak{commonsense (3\%);}}
\textit{\ak{inference (3\%);}} 
\textit{\ak{in-context (3\%);}} 
\textit{\ak{translation (2\%);}}
\textit{\ak{temporal (2\%);}} 
\textit{\ak{cross-lingual (2\%);}} \\
\midrule

\textbf{\ak{Hallucination}} &
\textit{\ak{hallucination (73\%);}}
\textit{\ak{accuracy (37\%);}}
\textit{\ak{factual (30\%);}}
\textit{\ak{inaccuracy (17\%);}}
\textit{\ak{reliability (17\%);}}
\textit{\ak{incorrect (13\%);}}
\textit{\ak{content (11\%);}}
\textit{\ak{inaccurate (11\%);}}
\textit{\ak{summarization (7\%);}}
\textit{\ak{object (7\%);}}
\textit{\ak{trustworthiness (7\%);}}
\textit{\ak{domain (6\%);}}
\textit{\ak{summary (6\%);}}
\textit{\ak{inconsistency (6\%);}}
\textit{\ak{reliance (6\%);}}
\textit{\ak{LVLM (5\%);}}
\textit{\ak{visual (4\%);}}
\textit{\ak{unfaithful (4\%);}}
\textit{\ak{misinformation (4\%);}}
\textit{\ak{detection (4\%);}} \\
\midrule

\textbf{\ak{Security Risks}} &
\textit{\ak{vulnerability (43\%);}}
\textit{\ak{attack (35\%);}}
\textit{\ak{risk (30\%);}}
\textit{\ak{security (25\%);}}
\textit{\ak{safety (22\%);}}
\textit{\ak{adversarial (18\%);}}
\textit{\ak{privacy (16\%);}}
\textit{\ak{prompt (13\%);}}
\textit{\ak{harmful (13\%);}}
\textit{\ak{robustness (12\%);}}
\textit{\ak{jailbreak (11\%);}}
\textit{\ak{detection (11\%);}}
\textit{\ak{malicious (9\%);}}
\textit{\ak{leakage (8\%);}}
\textit{\ak{backdoor (7\%);}}
\textit{\ak{defense (6\%);}}
\textit{\ak{threat (5\%);}}
\textit{\ak{toxicity (4\%);}}
\textit{\ak{copyright (3\%);}} \\
\midrule

\textbf{\ak{Generalization}} &
\textit{\ak{finetune (31\%);}}
\textit{\ak{training (22\%);}}
\textit{\ak{generalization (18\%);}}
\textit{\ak{domain (16\%);}}
\textit{\ak{distribution (16\%);}}
\textit{\ak{knowledge (15\%);}}
\textit{\ak{forget (12\%);}}
\textit{\ak{zero-shot (10\%);}}
\textit{\ak{few-shot (10\%);}}
\textit{\ak{catastrophic (8\%);}}
\textit{\ak{degradation (8\%);}}
\textit{\ak{scale (8\%);}}
\textit{\ak{instruction (8\%);}}
\textit{\ak{out-of-distribution (7\%);}}
\textit{\ak{adaptation (7\%);}}
\textit{\ak{prompt (7\%);}}
\textit{\ak{generation (6\%);}}
\textit{\ak{downstream (6\%);}}
\textit{\ak{constraint (6\%);}}
\textit{\ak{shift (5\%);}}
\textit{\ak{robustness (5\%);}} \\
\midrule

\textbf{\ak{Social Bias}} &
\textit{\ak{bias (93\%);}}
\textit{\ak{gender (18\%);}}
\textit{\ak{social (16\%);}}
\textit{\ak{stereotype (12\%);}}
\textit{\ak{representation (11\%);}}
\textit{\ak{fairness (10\%);}}
\textit{\ak{demographic (9\%);}}
\textit{\ak{harm (9\%);}}
\textit{\ak{cultural (8\%);}}
\textit{\ak{risk (8\%);}}
\textit{\ak{impact (7\%);}}
\textit{\ak{English (7\%);}}
\textit{\ak{societal (7\%);}}
\textit{\ak{ethical (6\%);}}
\textit{\ak{male (6\%);}}
\textit{\ak{moral (5\%);}}
\textit{\ak{western (4\%);}}
\textit{\ak{female (4\%);}}
\textit{\ak{encode (4\%);}}
\textit{\ak{mitigation (4\%);}} \\
\midrule

\textbf{\ak{Long Context}} &
\textit{\ak{text (48\%);}}
\textit{\ak{token (30\%);}}
\textit{\ak{context (27\%);}}
\textit{\ak{long (25\%);}}
\textit{\ak{use (18\%);}}
\textit{\ak{high (17\%);}}
\textit{\ak{length (16\%);}}
\textit{\ak{constraint (16\%);}}
\textit{\ak{generation (15\%);}}
\textit{\ak{training (15\%);}}
\textit{\ak{pre-trained (13\%);}}
\textit{\ak{word (12\%);}}
\textit{\ak{sequence (11\%);}}
\textit{\ak{processing (11\%);}}
\textit{\ak{size (11\%);}}
\textit{\ak{document (9\%);}}
\textit{\ak{embed (9\%);}}
\textit{\ak{representation (9\%);}}
\textit{\ak{tokenization (8\%);}}
\textit{\ak{cost (8\%);}} \\
\midrule

\textbf{\ak{Uncertainty}} &
\textit{\ak{prompt (25\%);}}
\textit{\ak{response (20\%);}}
\textit{\ak{high (14\%);}}
\textit{\ak{sensitivity (14\%);}}
\textit{\ak{correct (13\%);}}
\textit{\ak{reliability (12\%);}}
\textit{\ak{variability (12\%);}}
\textit{\ak{inconsistency (11\%);}}
\textit{\ak{accuracy (10\%);}}
\textit{\ak{alignment (10\%);}}
\textit{\ak{trust (10\%);}}
\textit{\ak{feedback (8\%);}}
\textit{\ak{confidence (8\%);}}
\textit{\ak{knowledge (8\%);}}
\textit{\ak{bias (8\%);}}
\textit{\ak{uncertainty (7\%);}}
\textit{\ak{calibration (7\%);}}
\textit{\ak{trustworthiness (7\%);}}
\textit{\ak{robustness (7\%);}}
\textit{\ak{preference (6\%);}}
\textit{\ak{variation (6\%);}} \\

\bottomrule
\end{tabularx}

\end{table*}

\begin{table*}[t]
\centering
\footnotesize
\setlength{\tabcolsep}{6pt}
\renewcommand{\arraystretch}{1.0}

\caption{\ak{Top 20 keywords per LLM limitation topic, identified using HDBSCAN with BERTopic in the arXiv dataset; keywords are extracted via TF-IDF and ranked by marginal document coverage (\%).}}
\label{tab:cluster_keywords_arxiv_long}

\begin{tabularx}{\textwidth}{@{}l X@{}}
\toprule
\textbf{\ak{Cluster}} & \textbf{\ak{Top 20 keywords (marginal document coverage)}} \\
\midrule

\textbf{\ak{Security Risks}} &
\textit{\ak{vulnerability (52\%);}}
\textit{\ak{risk (48\%);}}
\textit{\ak{attack (37\%);}}
\textit{\ak{security (28\%);}}
\textit{\ak{safety (24\%);}}
\textit{\ak{rate (24\%);}}
\textit{\ak{adversarial (19\%);}}
\textit{\ak{jailbreak (16\%);}}
\textit{\ak{prompt (15\%);}}
\textit{\ak{harmful (15\%);}}
\textit{\ak{content (13\%);}}
\textit{\ak{privacy (12\%);}}
\textit{\ak{alignment (11\%);}}
\textit{\ak{manipulation (11\%);}}
\textit{\ak{robustness (10\%);}}
\textit{\ak{malicious (10\%);}}
\textit{\ak{detection (9\%);}}
\textit{\ak{susceptibility (8\%);}}
\textit{\ak{ethical (8\%);}}
\textit{\ak{exploitation (8\%);}} \\
\midrule

\textbf{\ak{Social Bias}} &
\textit{\ak{bias (41\%);}}
\textit{\ak{prompt (15\%);}}
\textit{\ak{variability (11\%);}}
\textit{\ak{response (11\%);}}
\textit{\ak{reliability (11\%);}}
\textit{\ak{accuracy (10\%);}}
\textit{\ak{alignment (10\%);}}
\textit{\ak{sensitivity (10\%);}}
\textit{\ak{impact (8\%);}}
\textit{\ak{representation (7\%);}}
\textit{\ak{gender (7\%);}}
\textit{\ak{risk (7\%);}}
\textit{\ak{inconsistency (6\%);}}
\textit{\ak{quality (6\%);}}
\textit{\ak{preference (6\%);}}
\textit{\ak{social (6\%);}}
\textit{\ak{uncertainty (5\%);}}
\textit{\ak{content (5\%);}}
\textit{\ak{group (4\%);}}
\textit{\ak{stereotype (4\%);}} \\
\midrule

\textbf{\ak{Hallucination}} &
\textit{\ak{hallucination (95\%);}}
\textit{\ak{accuracy (29\%);}}
\textit{\ak{generation (28\%);}}
\textit{\ak{reliability (24\%);}}
\textit{\ak{text (20\%);}}
\textit{\ak{factual (17\%);}}
\textit{\ak{knowledge (15\%);}}
\textit{\ak{response (15\%);}}
\textit{\ak{content (14\%);}}
\textit{\ak{incorrect (12\%);}}
\textit{\ak{risk (11\%);}}
\textit{\ak{reasoning (9\%);}}
\textit{\ak{LVLM (9\%);}}
\textit{\ak{visual (8\%);}}
\textit{\ak{trustworthiness (8\%);}}
\textit{\ak{domain (7\%);}}
\textit{\ak{image (6\%);}}
\textit{\ak{object (6\%);}}
\textit{\ak{multimodal (6\%);}} \\
\midrule

\parbox[t]{2.6cm}{\raggedright
\textbf{\ak{Context}}\\
\textbf{\ak{\& Memory Lim.}}} &
\textit{\ak{limit (76\%);}}
\textit{\ak{training (22\%);}}
\textit{\ak{context (18\%);}}
\textit{\ak{token (18\%);}}
\textit{\ak{long (16\%);}}
\textit{\ak{finetune (15\%);}}
\textit{\ak{constraint (14\%);}}
\textit{\ak{high (13\%);}}
\textit{\ak{pre-trained (12\%);}}
\textit{\ak{knowledge (12\%);}}
\textit{\ak{length (12\%);}}
\textit{\ak{degradation (11\%);}}
\textit{\ak{generalization (11\%);}}
\textit{\ak{transformer (10\%);}}
\textit{\ak{domain (10\%);}}
\textit{\ak{distribution (9\%);}}
\textit{\ak{forget (9\%);}}
\textit{\ak{efficiency (9\%);}}
\textit{\ak{time (8\%);}}
\textit{\ak{computational (8\%);}} \\

\bottomrule
\end{tabularx}
\end{table*}

\begin{table*}[t]
\ContinuedFloat
\centering
\footnotesize
\setlength{\tabcolsep}{6pt}
\renewcommand{\arraystretch}{1.0}

\caption{\ak{Top 20 keywords per LLM limitation topic, identified using HDBSCAN with BERTopic in the arXiv dataset; keywords are extracted via TF-IDF and ranked by marginal document coverage (\%), continued.}}
\label{tab:cluster_keywords_arxiv_long}

\begin{tabularx}{\textwidth}{@{}l X@{}}
\toprule
\textbf{\ak{Cluster}} & \textbf{\ak{Top 20 keywords (continued)}} \\
\midrule

\parbox[t]{2.6cm}{\raggedright
\textbf{\ak{Code}}\\
\textbf{\ak{Generation}}} &
\textit{\ak{code (49\%);}}
\textit{\ak{generation (31\%);}}
\textit{\ak{benchmark (26\%);}}
\textit{\ak{program (15\%);}}
\textit{\ak{understand (12\%);}}
\textit{\ak{accuracy (11\%);}}
\textit{\ak{programming (10\%);}}
\textit{\ak{application (6\%);}}
\textit{\ak{software (6\%);}}
\textit{\ak{bug (5\%);}}
\textit{\ak{complexity (5\%);}}
\textit{\ak{variability (5\%);}}
\textit{\ak{instruction (5\%);}}
\textit{\ak{correctness (5\%);}}
\textit{\ak{real-world (5\%);}}
\textit{\ak{development (5\%);}}
\textit{\ak{open-source (4\%);}}
\textit{\ak{engineering (4\%);}}
\textit{\ak{structure (4\%);}}
\textit{\ak{execution (3\%);}} \\
\midrule

\textbf{\ak{Multimodality}} &
\textit{\ak{visual (28\%);}}
\textit{\ak{multimodal (25\%);}}
\textit{\ak{understand (24\%);}}
\textit{\ak{VLM (22\%);}}
\textit{\ak{image (21\%);}}
\textit{\ak{reasoning (21\%);}}
\textit{\ak{MLLM (20\%);}}
\textit{\ak{vision (17\%);}}
\textit{\ak{object (10\%);}}
\textit{\ak{spatial (9\%);}}
\textit{\ak{video (9\%);}}
\textit{\ak{comprehension (8\%);}}
\textit{\ak{perception (8\%);}}
\textit{\ak{vision–language (7\%);}}
\textit{\ak{recognition (7\%);}}
\textit{\ak{open-source (6\%);}}
\textit{\ak{representation (6\%);}}
\textit{\ak{ground (5\%);}}
\textit{\ak{LVLM (5\%);}}
\textit{\ak{alignment (5\%);}} \\
\midrule

\textbf{\ak{Reasoning}} &
\textit{\ak{reasoning (58\%);}}
\textit{\ak{understand (14\%);}}
\textit{\ak{logic (11\%);}}
\textit{\ak{math (9\%);}}
\textit{\ak{benchmark (9\%);}}
\textit{\ak{logical (9\%);}}
\textit{\ak{knowledge (9\%);}}
\textit{\ak{plan (9\%);}}
\textit{\ak{step (8\%);}}
\textit{\ak{solve (8\%);}}
\textit{\ak{mathematical (7\%);}}
\textit{\ak{agent (7\%);}}
\textit{\ak{cognitive (7\%);}}
\textit{\ak{chain (6\%);}}
\textit{\ak{CoT (6\%);}}
\textit{\ak{prompt (5\%);}}
\textit{\ak{multistep (4\%);}}
\textit{\ak{temporal (4\%);}}
\textit{\ak{inference (4\%);}}
\textit{\ak{structure (4\%);}} \\
\midrule

\textbf{\ak{Multilinguality}} &
\textit{\ak{multilingual (34\%);}}
\textit{\ak{bias (32\%);}}
\textit{\ak{cultural (30\%);}}
\textit{\ak{English (28\%);}}
\textit{\ak{representation (26\%);}}
\textit{\ak{resource (24\%);}}
\textit{\ak{low-resource (19\%);}}
\textit{\ak{translation (15\%);}}
\textit{\ak{disparity (15\%);}}
\textit{\ak{diversity (14\%);}}
\textit{\ak{linguistic (13\%);}}
\textit{\ak{knowledge (12\%);}}
\textit{\ak{understanding (9\%);}}
\textit{\ak{culture (9\%);}}
\textit{\ak{cross-lingual (8\%);}}
\textit{\ak{language-specific (8\%);}}
\textit{\ak{value (8\%);}}
\textit{\ak{English-centric (7\%);}}
\textit{\ak{western (7\%);}} \\
\midrule

\parbox[t]{2.6cm}{\raggedright
\textbf{\ak{Conversational}}\\
\textbf{\ak{Limitations}}} &
\textit{\ak{text (32\%);}}
\textit{\ak{understand (26\%);}}
\textit{\ak{context (25\%);}}
\textit{\ak{knowledge (15\%);}}
\textit{\ak{response (12\%);}}
\textit{\ak{generation (12\%);}}
\textit{\ak{retrieval (11\%);}}
\textit{\ak{RAG (11\%);}}
\textit{\ak{instruction (11\%);}}
\textit{\ak{reasoning (10\%);}}
\textit{\ak{comprehension (9\%);}}
\textit{\ak{conversation (9\%);}}
\textit{\ak{semantic (9\%);}}
\textit{\ak{interaction (8\%);}}
\textit{\ak{query (7\%);}}
\textit{\ak{structure (7\%);}}
\textit{\ak{follow (7\%);}}
\textit{\ak{processing (7\%);}}
\textit{\ak{long (7\%);}}
\textit{\ak{dialogue (7\%);}} \\
\midrule

\parbox[t]{2.6cm}{\raggedright
\textbf{\ak{Healthcare}}\\
\textbf{\ak{Application}}} &
\textit{\ak{medical (37\%);}}
\textit{\ak{application (28\%);}}
\textit{\ak{clinical (24\%);}}
\textit{\ak{interpretability (21\%);}}
\textit{\ak{domain (21\%);}}
\textit{\ak{understand (18\%);}}
\textit{\ak{health (17\%);}}
\textit{\ak{accuracy (15\%);}}
\textit{\ak{transparency (12\%);}}
\textit{\ak{risk (10\%);}}
\textit{\ak{reasoning (10\%);}}
\textit{\ak{complexity (10\%);}}
\textit{\ak{mental (9\%);}}
\textit{\ak{healthcare (8\%);}}
\textit{\ak{domain-specific (8\%);}}
\textit{\ak{reliability (8\%);}}
\textit{\ak{trust (8\%);}}
\textit{\ak{black-box (7\%);}}
\textit{\ak{variability (7\%);}}
\textit{\ak{diagnostic (7\%);}} \\
\midrule

\parbox[t]{2.6cm}{\raggedright
\textbf{\ak{Computational}}\\
\textbf{\ak{Cost}}} &
\textit{\ak{computation (49\%);}}
\textit{\ak{resource (47\%);}}
\textit{\ak{cost (38\%);}}
\textit{\ak{deploy (38\%);}}
\textit{\ak{high (32\%);}}
\textit{\ak{constraint (31\%);}}
\textit{\ak{memory (27\%);}}
\textit{\ak{inference (24\%);}}
\textit{\ak{requirement (23\%);}}
\textit{\ak{demand (22\%);}}
\textit{\ak{efficiency (22\%);}}
\textit{\ak{parameter (19\%);}}
\textit{\ak{training (18\%);}}
\textit{\ak{size (17\%);}}
\textit{\ak{significant (17\%);}}
\textit{\ak{device (16\%);}}
\textit{\ak{time (15\%);}}
\textit{\ak{consumption (14\%);}}
\textit{\ak{application (13\%);}}
\textit{\ak{resource-intensive (13\%);}}
\textit{\ak{energy (12\%);}} \\
\midrule

\parbox[t]{2.6cm}{\raggedright
\textbf{\ak{Benchmark}}\\
\textbf{\ak{Contamination}}} &
\textit{\ak{contamination (82\%);}}
\textit{\ak{train (69\%);}}
\textit{\ak{benchmark (52\%);}}
\textit{\ak{pre-trained (31\%);}}
\textit{\ak{leakage (28\%);}}
\textit{\ak{test (27\%);}}
\textit{\ak{reliability (19\%);}}
\textit{\ak{inflate (16\%);}}
\textit{\ak{assessment (15\%);}}
\textit{\ak{validity (13\%);}}
\textit{\ak{risk (13\%);}}
\textit{\ak{overlap (10\%);}}
\textit{\ak{memorization (9\%);}}
\textit{\ak{vulnerability (9\%);}}
\textit{\ak{integrity (9\%);}}
\textit{\ak{metric (9\%);}}
\textit{\ak{code (9\%);}}
\textit{\ak{score (7\%);}}
\textit{\ak{downstream (7\%);}}
\textit{\ak{exposure (7\%);}} \\
\midrule

\parbox[t]{2.6cm}{\raggedright
\textbf{\ak{Knowledge}}\\
\textbf{\ak{Editing}}} &
\textit{\ak{edit (100\%);}}
\textit{\ak{knowledge (49\%);}}
\textit{\ak{fact (30\%);}}
\textit{\ak{effect (28\%);}}
\textit{\ak{general (21\%);}}
\textit{\ak{degradation (17\%);}}
\textit{\ak{subject (13\%);}}
\textit{\ak{change (13\%);}}
\textit{\ak{reasoning (11\%);}}
\textit{\ak{relation (11\%);}}
\textit{\ak{factual (11\%);}}
\textit{\ak{lifelong (9\%);}}
\textit{\ak{consistency (9\%);}}
\textit{\ak{sequential (9\%);}}
\textit{\ak{update (9\%);}}
\textit{\ak{generalization (9\%);}}
\textit{\ak{localization (6\%);}}
\textit{\ak{deterioration (6\%);}}
\textit{\ak{distortion (6\%);}}
\textit{\ak{scalability (6\%);}} \\
\midrule

\textbf{\ak{Quantization}} &
\textit{\ak{quantization (85\%);}}
\textit{\ak{degradation (45\%);}}
\textit{\ak{bit (38\%);}}
\textit{\ak{int (34\%);}}
\textit{\ak{accuracy (32\%);}}
\textit{\ak{low (26\%);}}
\textit{\ak{trade-off (23\%);}}
\textit{\ak{activation (17\%);}}
\textit{\ak{inference (17\%);}}
\textit{\ak{memory (17\%);}}
\textit{\ak{impact (17\%);}}
\textit{\ak{outlier (15\%);}}
\textit{\ak{constraint (15\%);}}
\textit{\ak{computation (15\%);}}
\textit{\ak{low-bit (13\%);}}
\textit{\ak{quality (13\%);}}
\textit{\ak{size (13\%);}}
\textit{\ak{weight (13\%);}}
\textit{\ak{PTQ (13\%);}}
\textit{\ak{parameter (11\%);}} \\

\bottomrule
\end{tabularx}
\end{table*}

\begin{table*}
\centering
\footnotesize
\caption{\ak{Representative limitation-focused papers grouped by HDBSCAN- and LlooM-specific clusters across both ACL and arXiv.} The numbers in parentheses in the \enquote{Topic} column indicate the example ID referenced in the main text.}
\label{tab:hdbscan-lloom-specifc-examples}
\renewcommand{\arraystretch}{1.1}
\begin{tabularx}{\textwidth}{>{\centering\arraybackslash}p{1.9cm} >{\raggedright\arraybackslash}X}
\toprule
\textbf{Topic} & \textbf{Evidence} \\
\midrule
\multicolumn{2}{c}{\footnotesize\textit{\ak{HDBSCAN-specific clusters across both arXiv and ACL}}} \\
\midrule
\footnotesize (10) \newline Uncertainty & [...] \footnotesize \enquote{We find that methods aimed at improving usability, such as fine-tuning and chain-of-thought (CoT) prompting, can lead to miscalibration and unreliable natural language explanations.} (NAACL 2024~\citep{zhang2023study}) \\
\midrule
\footnotesize (11) Code \newline Generation & \footnotesize \enquote{LLMs have demonstrated impressive code generation capabilities but struggle with real-world software engineering tasks, such as revising source code to address code reviews, hindering their practical use.} (arXiv, March 2025 \citep{lin2025codereviewqa}) \\ 
\midrule
\footnotesize (12) Conversational \newline Limitations & \footnotesize \enquote{[...] Answers from LLMs can be improved with additional context. [...] Our results show multi-turn interactions are usually required for datasets which have a high proportion of incompleteness or ambiguous questions} (arXiv, March 2025 \citep{naik2025empirical}) \\ 
\midrule
\footnotesize (13) Healthcare \newline Application & \footnotesize \enquote{However, GPT3.5 performance falls behind BERT and a radiologist. [...] By analyzing the explanations of GPT3.5 for misclassifications, we reveal systematic errors that need to be resolved to enhance its safety and suitability for clinical use.} (arXiv, June 2023 \citep{talebi2023beyond}) \\ 
\midrule
\footnotesize (14) Benchmark \newline Contamination & \footnotesize \enquote{However, as LLMs are typically trained on vast amounts of data, a significant concern in their evaluation is data contamination, where overlap between training data and evaluation datasets inflates performance assessments.} (arXiv, October 2023 \citep{fu2024does}) \\ 
\midrule
\footnotesize (15) \newline Quantization & \footnotesize \enquote{This study examines 4-bit quantization methods like GPTQ in large language models (LLMs), highlighting GPTQ's overfitting and limited enhancement in Zero-Shot tasks.} (arXiv, December 2023 \citep{wu2023zeroquant}) \\ 
\midrule
\multicolumn{2}{c}{\footnotesize\textit{\ak{LlooM-specific clusters across both arXiv and ACL}}} \\
\midrule
\footnotesize (16) Privacy \newline Risks & \footnotesize \enquote{Parts of this memorized content have been shown to be extractable by simply querying the model, which poses a privacy risk.} (ACL 2023 \citep{ozdayi2023controlling}) \\ 
\midrule
\footnotesize (17) Catastrophic Forgetting & \footnotesize \enquote{Fine-tuning with only few-shot samples, the LMs can easily forget pretrained knowledge, overfit spurious biases, and suffer from compositionally out-of-distribution generalization errors.} (NAACL 2022 \citep{yang2022seqzero}) \\ 
\midrule
\footnotesize (18) Data \newline Contamination & \footnotesize \enquote{Data contamination in model evaluation has become increasingly prevalent with the growing popularity of LLMs. It allows models to cheat via memorisation instead of displaying true capabilities.} (EMNLP 2024 \citep{li2023open}) \\ 
\midrule
\footnotesize (19) Trustworthiness & \footnotesize \enquote{However, ChatGPT is a closed-source product which has major drawbacks with regards to transparency, reprodicibility, cost, and data protection.} (arXiv, November 2023 \citep{kristensen2023chatbots}) \\ 
\midrule
\footnotesize (20) Alignment \newline \& Trustworthiness & \footnotesize \enquote{LLMs can generate outputs that are untruthful, toxic, or simply not helpful to the user. In other words, these models are not aligned with their users. [...]} (arXiv, March 2022 \citep{ouyang2022training}) \\ 
\midrule
\footnotesize (21) Prompt \newline Sensitivity & \footnotesize \enquote{[...] prompting [...] has been shown to be susceptible to miscalibration and brittleness to slight prompt variations, caused by its discriminative prompting approach, i.e., predicting the label given the input.} (arXiv, November 2023 \citep{kumar2023gen}) \\ 
\midrule
\footnotesize (22) Overconfidence & \footnotesize \enquote{Despite their broad utility, LLMs tend to generate information that conflicts with real-world facts, and their persuasive style can make these inaccuracies appear confident and convincing.} (arXiv, September 2024 \citep{chaudhry2024finetuning}) \\ 
\bottomrule
\end{tabularx}
\end{table*}

\begin{table}[h]
    \centering
    \caption{Best-aligned LlooM topics for each HDBSCAN cluster based on Jaccard overlap of paper sets.}
    \label{tab:cluster_alignment_hdbscan_to_lloom}
    \renewcommand{\arraystretch}{1.1}

    \begin{subtable}[t]{\textwidth}
        \centering
        \caption{ACL Dataset}
        \small
        \begin{tabular}{l l c}
            \toprule
            \textbf{HDBSCAN Topic} & \textbf{Best-Matching LlooM Topic} & \textbf{Jaccard Overlap} \\
            \midrule
            Social Bias & Bias and Fairness & 0.498 \\
            Security & Security Risks & 0.495 \\
            Hallucination & Hallucination & 0.429 \\
            Reasoning & Reasoning & 0.359 \\
            Long Context & Long Context & 0.172 \\
            Generalization & Generalization & 0.155 \\
            Uncertainty & Knowledge Editing & 0.085 \\
            \midrule
            \multicolumn{2}{r}{\textbf{Average Top Jaccard Overlap}} & \textbf{0.313} \\
            \bottomrule
        \end{tabular}
    \end{subtable}

    \vspace{1em}

    \begin{subtable}[t]{\textwidth}
        \centering
        \caption{arXiv Dataset}
        \small
        \begin{tabular}{l l c}
            \toprule
            \textbf{HDBSCAN Topic} & \textbf{Best-Matching LlooM Topic} & \textbf{Jaccard Overlap} \\
            \midrule
            Security Risks & Security Risks & 0.565 \\
            Social Bias & Bias and Fairness & 0.288 \\
            Context \& Memory Limitations & Long Context & 0.150 \\
            Benchmark Contamination & Privacy Risks & 0.011 \\
            Hallucination & Hallucination & 0.477 \\
            Code Generation & Generalization & 0.034 \\
            Reasoning & Reasoning & 0.302 \\
            Conversational Limitations & Reasoning & 0.070 \\
            Multimodality & Multimodality & 0.379 \\
            Multilinguality & Language and Cultural Limitations & 0.393 \\
            Healthcare Application & Alignment Limitations & 0.027 \\
            Computational Cost & Privacy Risks & 0.016 \\
            Knowledge Editing & Overconfidence & 0.005 \\
            Quantization & Knowledge Editing & 0.090 \\
            \midrule
            \multicolumn{2}{r}{\textbf{Average Top Jaccard Overlap}} & \textbf{0.201} \\
            \bottomrule
        \end{tabular}
    \end{subtable}
\end{table}

\begin{table}[h]
    \centering
    \caption{Best-aligned HDBSCAN topics for each LlooM topic based on Jaccard overlap of paper sets.}
    \label{tab:cluster_alignment_lloom_to_hdbscan}
    \renewcommand{\arraystretch}{1.1}

    \begin{subtable}[t]{\textwidth}
        \centering
        \caption{ACL Dataset}
        \small
        \begin{tabular}{l l c}
            \toprule
            \textbf{LlooM Topic} & \textbf{Best-Matching HDBSCAN Topic} & \textbf{Jaccard Overlap} \\
            \midrule
            Generalization & Reasoning & 0.194 \\
            Long Context & Long Context & 0.172 \\
            Hallucination & Hallucination & 0.429 \\
            Language and Cultural Limitations & Social Bias & 0.204 \\
            Reasoning & Reasoning & 0.359 \\
            Bias and Fairness & Social Bias & 0.498 \\
            Security Risks & Security & 0.495 \\
            Knowledge Editing & Reasoning & 0.120 \\
            Catastrophic Forgetting & Generalization & 0.155 \\
            Computational Cost & Generalization & 0.114 \\
            Privacy Risks & Security & 0.191 \\
            Multimodality & Reasoning & 0.127 \\
            Data Contamination & Reasoning & 0.042 \\
            \midrule
            \multicolumn{2}{r}{\textbf{Average Top Jaccard Overlap}} & \textbf{0.239} \\
            \bottomrule
        \end{tabular}
    \end{subtable}

    \vspace{1em}
    
    \begin{subtable}[t]{\textwidth}
        \centering
        \caption{arXiv Dataset}
        \small
        \begin{tabular}{l l c}
            \toprule
            \textbf{LlooM Topic} & \textbf{Best-Matching HDBSCAN Topic} & \textbf{Jaccard Overlap} \\
            \midrule
            Multimodality & Multimodality & 0.379 \\
            Alignment Limitations & Hallucination & 0.160 \\
            Hallucination & Hallucination & 0.477 \\
            Trustworthiness & Security Risks & 0.263 \\
            Generalization & Reasoning & 0.116 \\
            Language and Cultural Limitations & Multilinguality & 0.393 \\
            Bias and Fairness & Social Bias & 0.288 \\
            Privacy Risks & Security Risks & 0.152 \\
            Prompt Sensitivity & Hallucination & 0.111 \\
            Reasoning & Reasoning & 0.302 \\
            Knowledge Editing & Knowledge Editing & 0.090 \\
            Overconfidence & Hallucination & 0.081 \\
            Security Risks & Security Risks & 0.565 \\
            Long Context & Context \& Memory Limitations & 0.150 \\
            Catastrophic Forgetting & Context \& Memory Limitations & 0.125 \\
            \midrule
            \multicolumn{2}{r}{\textbf{Average Top Jaccard Overlap}} & \textbf{0.244} \\
            \bottomrule
        \end{tabular}
    \end{subtable}
\end{table}

\clearpage

\section{HDBSCAN + BERTopic Trend Analysis}
\label{sec:hdbscan-trend-analysis}

In this section, we discuss \lllm\ topic dynamics over time as identified by HDBSCAN + BERTopic. We apply the following perspectives:

\begin{enumerate}[label=(\roman*)]
    \item \textbf{LLM-wide share}, measured annually as \( \frac{N^{\text{lim}}_{t,y}}{N^{\text{LLM}}_y} \), to reflect how often limitation topic \( t \) appears in LLM research in year \( y \), relative to the total LLM papers. This shows whether a topic is gaining attention beyond \lllm\ research and becoming part of the general LLM research agenda.
    
    \item \textbf{Limitations share}, measured quarterly as \( \frac{N^{\text{lim}}_{t,q}}{N^{\text{lim}}_q} \), to reflect the share of limitation-focused papers in quarter \( q \) that address topic \( t \). This shows the topic's visibility within \lllm\ subfield.

    
\end{enumerate}

\noindent
Here, \( N^{\text{lim}}_{t,y} \) is the number of limitation papers on topic \( t \) in year \( y \); \( N^{\text{LLM}}_y \) is the total number of LLM papers in that year; and \( N^{\text{lim}}_{t,q} \), \( N^{\text{lim}}_q \) are the number of limitation papers on topic \( t \) and the total number of limitation papers, respectively, in quarter \( q \).

\paragraph{(i) How are limitation topics represented in the broader growth of LLM research?}

\begin{center}
\begin{tcolorbox}[title=Key Insights, fontupper=\small, width=0.9\textwidth]
\begin{itemize}
    \item Limitation topics show growing visibility within LLM research across both ACL and arXiv datasets. Some concerns, such as \textit{Security} and \textit{Long Context} in ACL, surge specifically in 2024, while others like \textit{Reasoning} and \textit{Hallucination} rise more steadily.
    \item On arXiv, the dramatic relative increases seen in 2023 (e.g., \textit{Hallucination}, \textit{Code Generation}, \textit{Healthcare Application}) reflect the emergence of new concerns from a low baseline, but many of these remain small in absolute share.
\end{itemize}
\end{tcolorbox}
\end{center}

\begin{figure}[h]
    \centering
    \resizebox{0.8\textwidth}{!}{
    \begin{tabular}{cc}
        \begin{subfigure}[b]{0.49\textwidth}
            \centering
            \includegraphics[width=\textwidth]{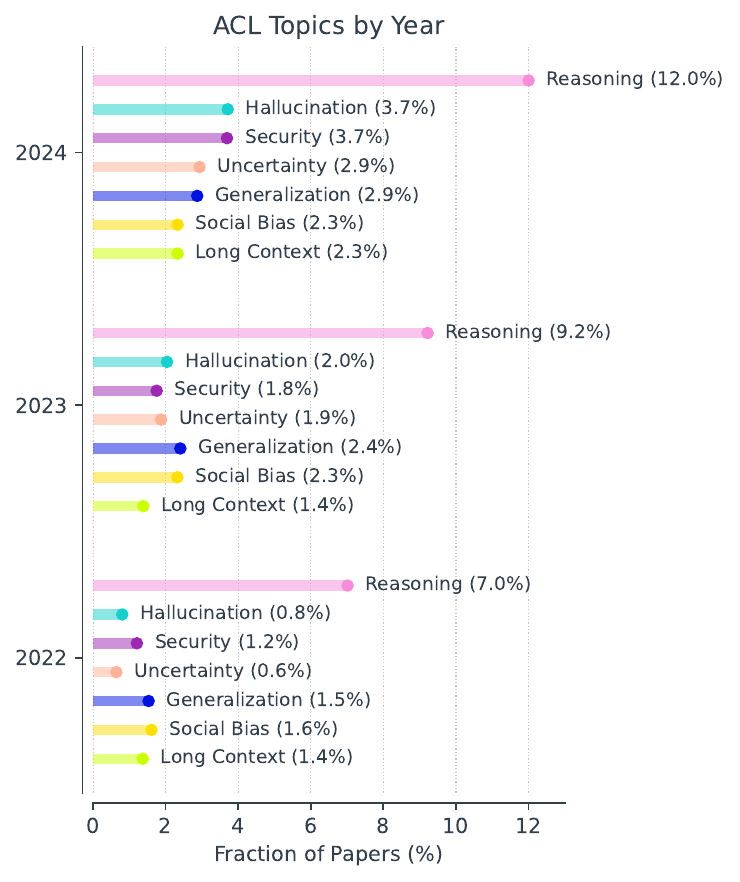}
            \caption{\footnotesize ACL Topic Distribution Per Year}
            \label{fig:acl-topic-per-year-hdbscan}
        \end{subfigure} &
        
        \begin{subfigure}[b]{0.49\textwidth}
            \centering
            \includegraphics[width=\textwidth]{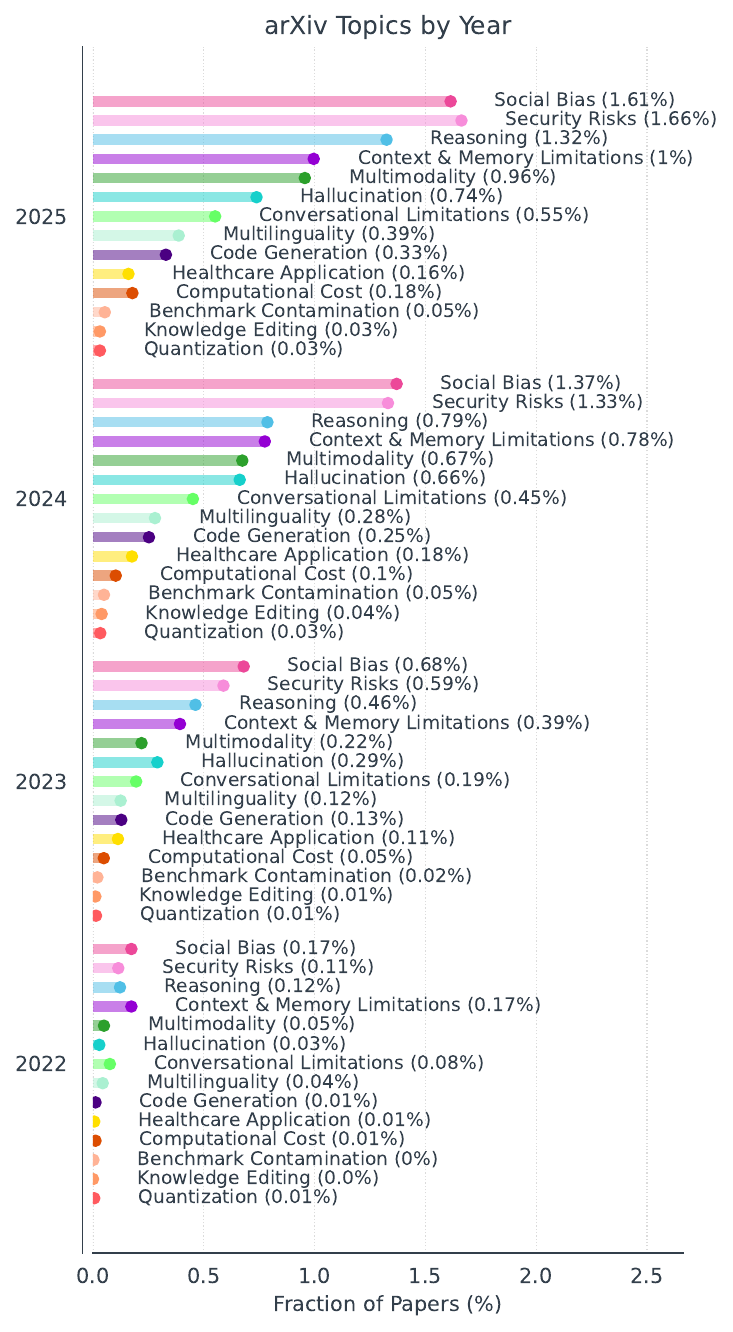}
            \caption{\footnotesize arXiv Topic Distribution Per Year}
            \label{fig:arxiv-topic-per-year-hdbscan}
        \end{subfigure} 
    \end{tabular}
    }
    \caption{Distribution of LLM limitation topics over years for ACL and arXiv, based on clustering results with HDBSCAN + BERT. Percentages reflect each topic’s proportion out of the total LLM-focused papers (8,635 in ACL and 41,991 in arXiv).}
    \label{fig:hdbscan-per-year}
\end{figure}

Figure~\ref{fig:hdbscan-per-year} shows the annual distribution of LLM limitation topics as a proportion of all LLM-focused papers in ACL and arXiv. 
Across both ACL and arXiv datasets, most limitation topics show grow in visibility across LLM research from 2022 to 2024 in ACL and 2022 to early 2025 in arXiv. However, as shown in Table~\ref{tab:relative_growth_acl_arxiv_hdbscan} in the supplementary material, they do so at different pace. On ACL, some concerns surged sharply in 2024, such as \textit{Security} (+106\%) and \textit{Long Context} (+64\%), while others like \textit{Reasoning} and \textit{Hallucination} grew more steadily across years. In contrast, topics such as \textit{Generalization} and \textit{Social Bias} showed growth in 2023 but flattened or declined by 2024.

On arXiv, most topics were marginal in 2022, leading to dramatic relative growth in 2023 as new concerns entered the field: for example, \textit{Hallucination} (+867\%), \textit{Code Generation} (+1200\%), and \textit{Healthcare Application} (+1000\%). Despite steep increases, many remained low in absolute share (e.g., \textit{Code Generation} at 0.13\% in 2023). After this emergence, growth remained strong though less extreme, especially for \textit{Reasoning}, \textit{Security Risks}, \textit{Multimodality}, and \textit{Computational Cost}. A few topics declined by 2025, including \textit{Benchmark Contamination}, \textit{Quantization}, and \textit{Knowledge Editing}. However, these topics are the least represented, and since 2025 data covers only one quarter, recent drops should be interpreted with caution.

These comparisons reflect annual, macro-level trends in topical focus and relative prominence. To examine shorter-term dynamics and account for changes in overall LLM publication volume, measure the percentage of papers on each topic, relative to all LLM limitation papers, for each quarter. Additionally, we apply the Mann-Kendall trend test \citep{mann1945nonparametric} to identify whether each topic's share exhibits a consistent upward or downward trend across quarters. This non-parametric test is well-suited for detecting monotonic trends in time series without assuming linearity or normality. However, non-monotonic changes (peaking behavior or decreases followed by increases) are not detected by this test, such that some observations below will be associated with high $p$-values.

\paragraph{(ii) What are the internal trends within LLM limitation research?}

\begin{center}
\begin{tcolorbox}[title=Key Insights, fontupper=\small, width=0.9\textwidth]
\begin{itemize}
    \item On ACL, \textit{Uncertainty} is the only topic with a statistically significant upward trend, while \textit{Social Bias} and \textit{Reasoning} show non-significant declines after early peaks. Most other topics remain relatively stable over time.
    
    \item On arXiv, \textit{Multimodality} and \textit{Hallucination} show significant growth, with the former likely linked to the rise of vision-language models. \textit{Knowledge Editing} also increases, though not significantly.
\end{itemize}
\end{tcolorbox}
\end{center}
\begin{figure}[h]
    \centering
    \begin{subfigure}[t]{0.9\textwidth}
        \includegraphics[width=\textwidth]{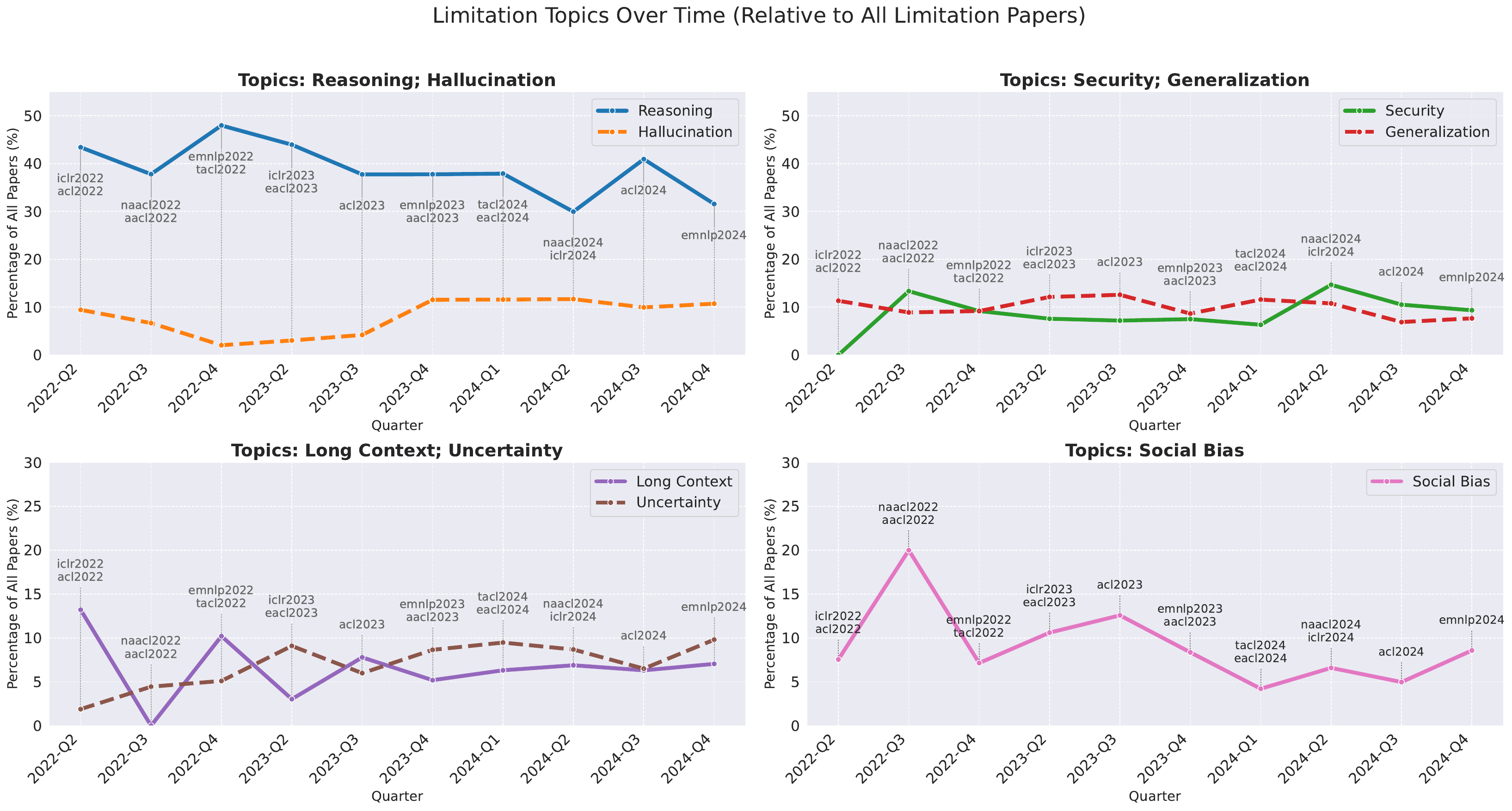}
        \caption{ACL \lllm\ trends normalized by \lllm\ papers.}
        \label{fig:acl-HDBSCAN-subfig}
    \end{subfigure}
    \hfill
    \begin{subfigure}[t]{0.9\textwidth}
        \includegraphics[width=\textwidth]{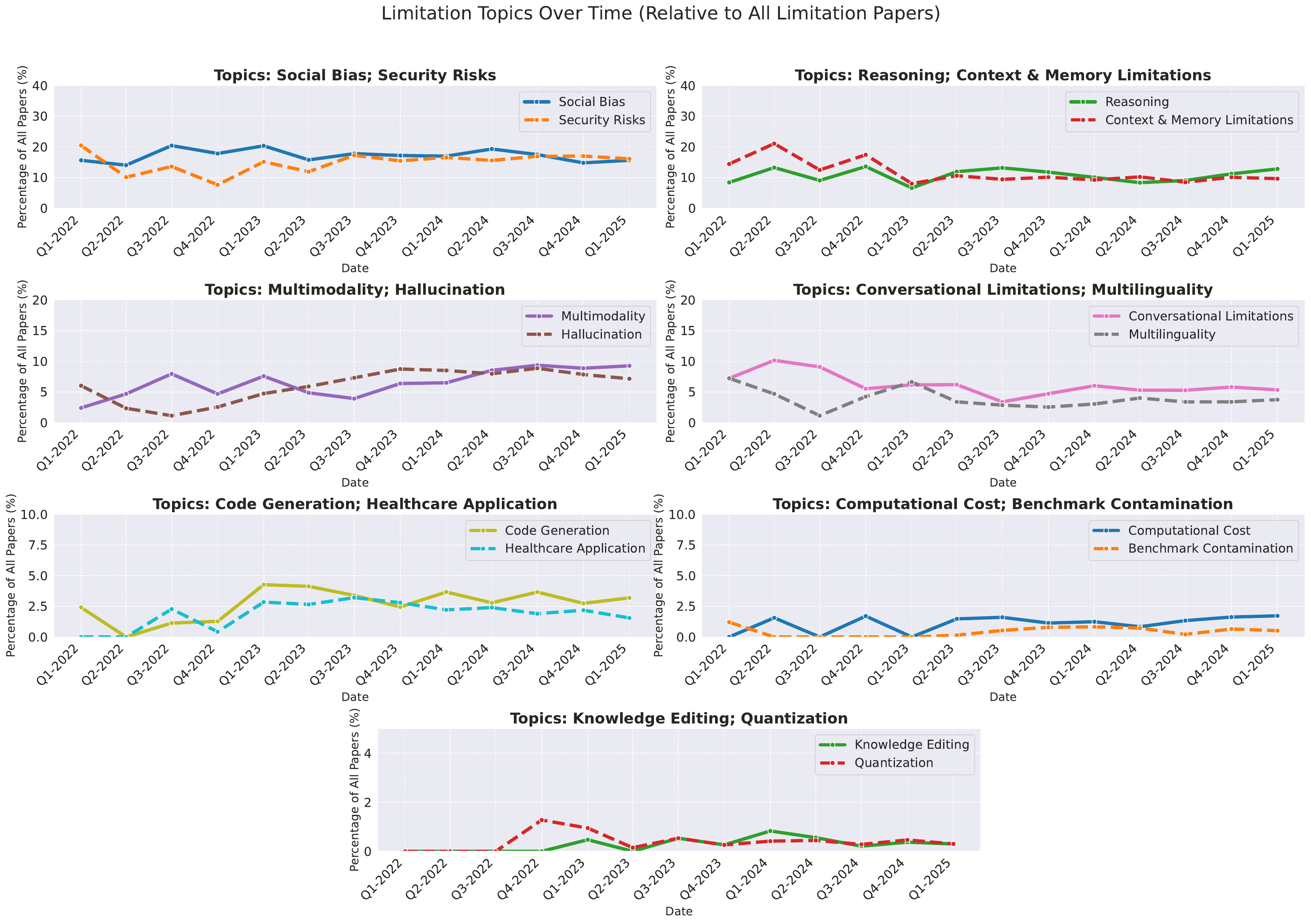}
        \caption{arXiv \lllm\ trends normalized by \lllm\ papers.}
        \label{fig:arxiv-HDBSCAN-subfig}
    \end{subfigure}

    \caption{\lllm\ topics trends for ACL and arXiv datasets based on HDBSCAN + BERTopic clustering approach (normalized by \lllm\ papers). Note that y-axis limits vary across subplots to reflect differences in topic prevalence and improve visualization.}
    \label{fig:acl-arxiv-HDBSCAN-trends}
\end{figure}


Among ACL limitation topics (Figure~\ref{fig:acl-HDBSCAN-subfig}), we observe the following temporal patterns:

\begin{itemize} 
\item $\uparrow$ \textbf{Increasing:}
\textit{Uncertainty} shows the clearest upward trend, rising from around 5–6\% in mid-2022 to over 10\% by late 2024. This trend is statistically significant according to the Mann-Kendall test ($\tau$ = 0.64, $p$ = 0.0123). \textit{Hallucination} also trends upward but does not reach statistical significance ($\tau$ = 0.42, $p$ = 0.107).

\item $\downarrow$ \textbf{Decreasing:} In contrast, \textit{Social Bias} peaks sharply in 2022-Q3 (26\%), then declines and stabilizes below 15\% in subsequent quarters. Due to the inconsistency of this change, it is not significant in the Mann-Kendal test ($p$ = 0.283). \textit{Reasoning} remains the most prevalent topic, consistently ranging between 30–40\% per quarter. While the test indicates a downward tendency ($\tau$ = –0.38), the trend is not statistically significant ($p$ = 0.152).

Other topics remain relatively stable over time:

\item $\rightarrow$ \textbf{Plateau / Stable:}
\textit{Hallucination}, \textit{Security}, and \textit{Generalization} each remain within a 5–15\% range with no significant directional movement.
\textit{Long Context} also shows minor variation but no sustained change. These patterns are consistent with the Mann-Kendall test, which detects no significant trends for any of these topics. 
\end{itemize}

Some topics also show sharp, isolated changes. \textit{Social Bias} spikes in 2022-Q3 (26\%), then drops to around 10\% by early 2023. This may reflect heightened concern about bias in the early stages of LLM deployment, which is also visible in the annual distribution (Figure~\ref{fig:hdbscan-per-year}), where \textit{Social Bias} is the second most prominent topic in 2022 but (relative to the overall expansion of \lllm\ research) declines in later years. \textit{Reasoning} peaks in 2022-Q4 and again in 2023-Q3 (~35–40\%), likely reflecting increased interest following the launch of ChatGPT and GPT-4 \citep{achiam2023gpt} in late 2022 and early 2023. 
\textit{Security} sees a modest increase in 2024-Q2 and Q3, which might be related to growing concerns about jailbreaks and adversarial attacks.

Among arXiv limitation topics (Figure~\ref{fig:arxiv-HDBSCAN-subfig}), we observe the following temporal patterns:

\begin{itemize} 
\item $\uparrow$ \textbf{Increasing:}
\textit{Multimodality} shows the most consistent upward trend, increasing from around 5\% in early 2022 to 15\% by early 2025. This pattern is statistically significant according to the Mann-Kendall trend test ($\tau$ = 0.59, $p$ = 0.006), and likely reflects heightened interest following the release of GPT-4V \citep{achiam2023gpt}, LLaVA \citep{liu2023visual}, and other vision-language models in mid-2023. 

\textit{Hallucination} also grows slightly over time, from ~10\% to ~13\%, and this increase is statistically significant as well ($\tau$ = 0.51, $p$ = 0.017). 
\textit{Knowledge Editing} also exhibits an upward trajectory, though this trend does not reach statistical significance ($\tau$ = 0.41, $p$ = 0.051).

\item $\downarrow$ \textbf{Decreasing:}
\textit{Context \& Memory Limitations} exhibits a clear and statistically significant decline, dropping from 25\% in 2022-Q1 to around 12–13\% by 2025 ($\tau$ = –0.46, $p$ = 0.033). This shift may reflect improved handling of long inputs through retrieval-augmented methods and the emergence of models with extended context capabilities \citep{li2024retrieval}. 

\textit{Conversational Limitations} follows a similar downward trajectory ($\tau$ = –0.44, $p$ = 0.044), though its decline appears to plateau at the beginning of 2024.

\item $\rightarrow$ \textbf{Stable / Plateau:}
Other topics remain relatively stable: \textit{Security Risks} appears to grow from 12\% to 18\%, and \textit{Social Bias} gradually declines from ~23\% to ~17\%, but neither trend is statistically significant ($p$ = 0.127 and $p$ = 0.502, respectively). This may be because concerns regarding social bias increased immediately following the release of ChatGPT but then returned to the pre-ChatGPT level. 
\textit{Reasoning}, \textit{Multilinguality}, \textit{Code Generation}, and \textit{Healthcare Application} all fluctuate without a clear directional trend, while \textit{Computational Cost}, \textit{Benchmark Contamination}, \textit{Knowledge Editing}, and \textit{Quantization} remain low throughout. 
\end{itemize}

\clearpage

\begin{table}[h]
    \centering
    \caption{Relative percentage growth of LLM-normalized shares for limitation topics across ACL and arXiv datasets (HDBSCAN+BERTopic). Values are based on relative change between consecutive years. Percentage change is not reported (—) where the earlier year's value is zero.}
    \label{tab:relative_growth_acl_arxiv_hdbscan}
    \renewcommand{\arraystretch}{1.2}
    \small

    \begin{subtable}[t]{\textwidth}
        \centering
        \caption{ACL Dataset}
        \label{tab:relative_growth_acl_alt}
        \begin{tabular}{l c c c}
            \toprule
            \textbf{Topic} & \textbf{Topic Shares (2022 → 2024)} & \textbf{→2023 (\%)} & \textbf{→2024 (\%)} \\
            \midrule
            Reasoning & 7.0 → 9.2 → 12.0 & 31.43\% & 30.43\% \\
            Hallucination & 0.8 → 2.0 → 3.7 & 150.00\% & 85.00\% \\
            Security & 1.2 → 1.8 → 3.7 & 50.00\% & 105.56\% \\
            Uncertainty & 0.6 → 1.9 → 2.9 & 216.67\% & 52.63\% \\
            Generalization & 1.5 → 2.4 → 2.9 & 60.00\% & 20.83\% \\
            Social Bias & 1.6 → 2.3 → 2.3 & 43.75\% & 0.00\% \\
            Long Context & 1.4 → 1.4 → 2.3 & 0.00\% & 64.29\% \\
            \bottomrule
        \end{tabular}
    \end{subtable}

    \vspace{1em}

    \begin{subtable}[t]{\textwidth}
        \centering
        \caption{arXiv Dataset}
        \label{tab:relative_growth_arxiv_alt}
        \begin{tabular}{l c c c c}
            \toprule
            \textbf{Topic} & \textbf{Topic Shares (2022 → 2025)} & \textbf{→2023 (\%)} & \textbf{→2024 (\%)} & \textbf{→2025 (\%)} \\
            \midrule
            Social Bias & 0.17 → 0.68 → 1.37 → 1.61 & 300.00\% & 101.47\% & 17.52\% \\
            Security Risks & 0.11 → 0.59 → 1.33 → 1.66 & 436.36\% & 125.42\% & 24.81\% \\
            Reasoning & 0.12 → 0.46 → 0.79 → 1.32 & 283.33\% & 71.74\% & 67.09\% \\
            Context \& Memory Limitations & 0.17 → 0.39 → 0.78 → 1.0 & 129.41\% & 100.00\% & 28.21\% \\
            Multimodality & 0.05 → 0.22 → 0.67 → 0.96 & 340.00\% & 204.55\% & 43.28\% \\
            Hallucination & 0.03 → 0.29 → 0.66 → 0.74 & 866.67\% & 127.59\% & 12.12\% \\
            Conversational Limitations & 0.08 → 0.19 → 0.45 → 0.55 & 137.50\% & 136.84\% & 22.22\% \\
            Multilinguality & 0.04 → 0.12 → 0.28 → 0.39 & 200.00\% & 133.33\% & 39.29\% \\
            Code Generation & 0.01 → 0.13 → 0.25 → 0.33 & 1200.00\% & 92.31\% & 32.00\% \\
            Healthcare Application & 0.01 → 0.11 → 0.18 → 0.16 & 1000.00\% & 63.64\% & -11.11\% \\
            Computational Cost & 0.01 → 0.05 → 0.1 → 0.18 & 400.00\% & 100.00\% & 80.00\% \\
            Benchmark Contamination & 0.0 → 0.02 → 0.05 → 0.05 & — & 150.00\% & 0.00\% \\
            Knowledge Editing & 0.0 → 0.01 → 0.04 → 0.03 & — & 300.00\% & -25.00\% \\
            Quantization & 0.01 → 0.01 → 0.03 → 0.03 & 0.00\% & 200.00\% & 0.00\% \\
            \bottomrule
        \end{tabular}
    \end{subtable}
\end{table}

\clearpage

\section{LlooM Topic Co-Occurrence}
\label{sec:topic-co-occurrence}


\lllm\ are often studied in combination, as reflected in the high number of multi-topic papers: 43\% in ACL and over 60\% in arXiv, with some spanning up to eight topics (see Table~\ref{tab:topic-count-distribution}). These overlaps often reflect shared task setups (e.g., multimodal hallucination) or related concerns (e.g., alignment and bias). We analyze these links via topic co-occurrence. 

\begin{table}[h] 
\centering 
\caption{Distribution of topic counts per paper in ACL and arXiv datasets} 
\label{tab:topic-count-distribution} 
\begin{tabular}{@{}lcc@{}} 
\toprule 
\textbf{Number of Topics} & \textbf{ACL Papers} & \textbf{arXiv Papers} \\
\midrule 
1 topic & 1,093 & 3,317 \\
2 topics & 726 & 4,182 \\
3 topics & 253 & 2,307 \\
4 topics & 77 & 667 \\
5 topics & 9 & 159 \\
6 topics & 2 & 29 \\
7 topics & --- & 5 \\
8 topics & --- & 1 \\
\bottomrule 
\end{tabular} 
\end{table}

\begin{figure}[h]
    \centering
    \renewcommand{\thesubfigure}{\roman{subfigure}}
    \resizebox{1.0\textwidth}{!}{
    \begin{tabular}{cc}
        \begin{subfigure}[b]{0.49\textwidth}
            \centering
            \includegraphics[width=\textwidth]{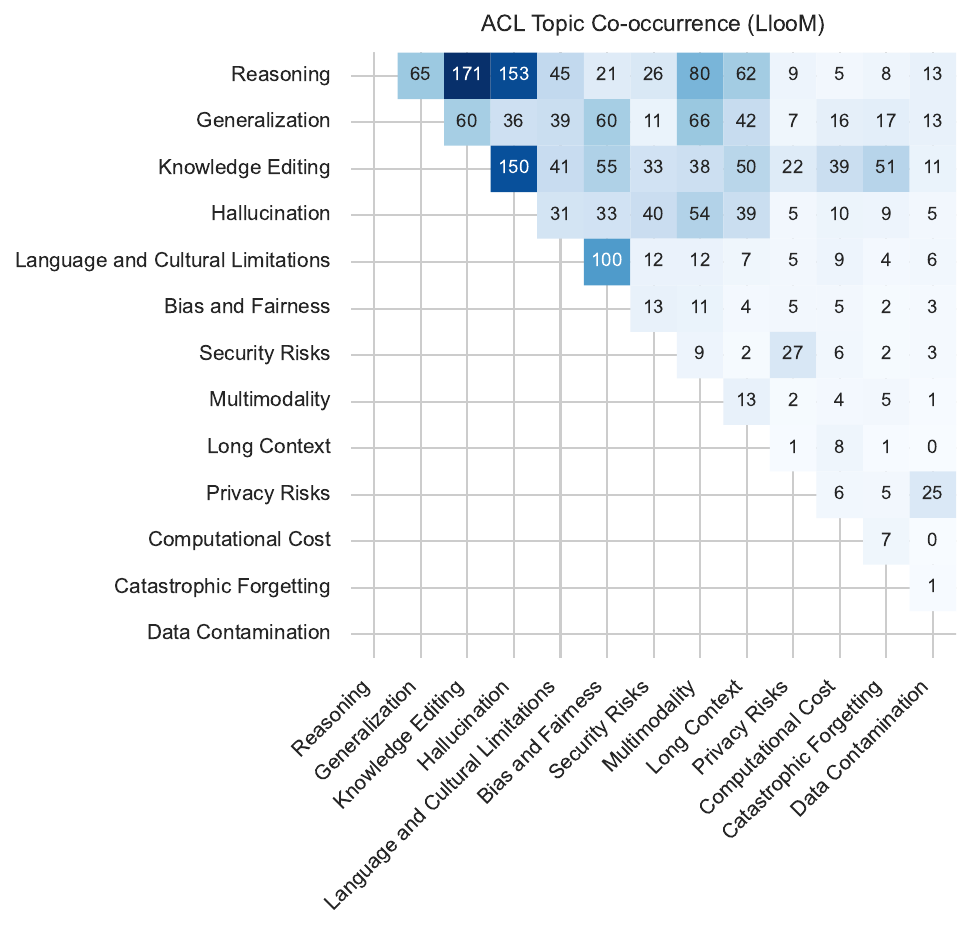}
            \caption{\footnotesize ACL Topic Co-occurrence Matrix}
            \label{fig:acl-cooccurrence-matrix}
        \end{subfigure} &
        
        \begin{subfigure}[b]{0.49\textwidth}
            \centering
            \includegraphics[width=\textwidth]{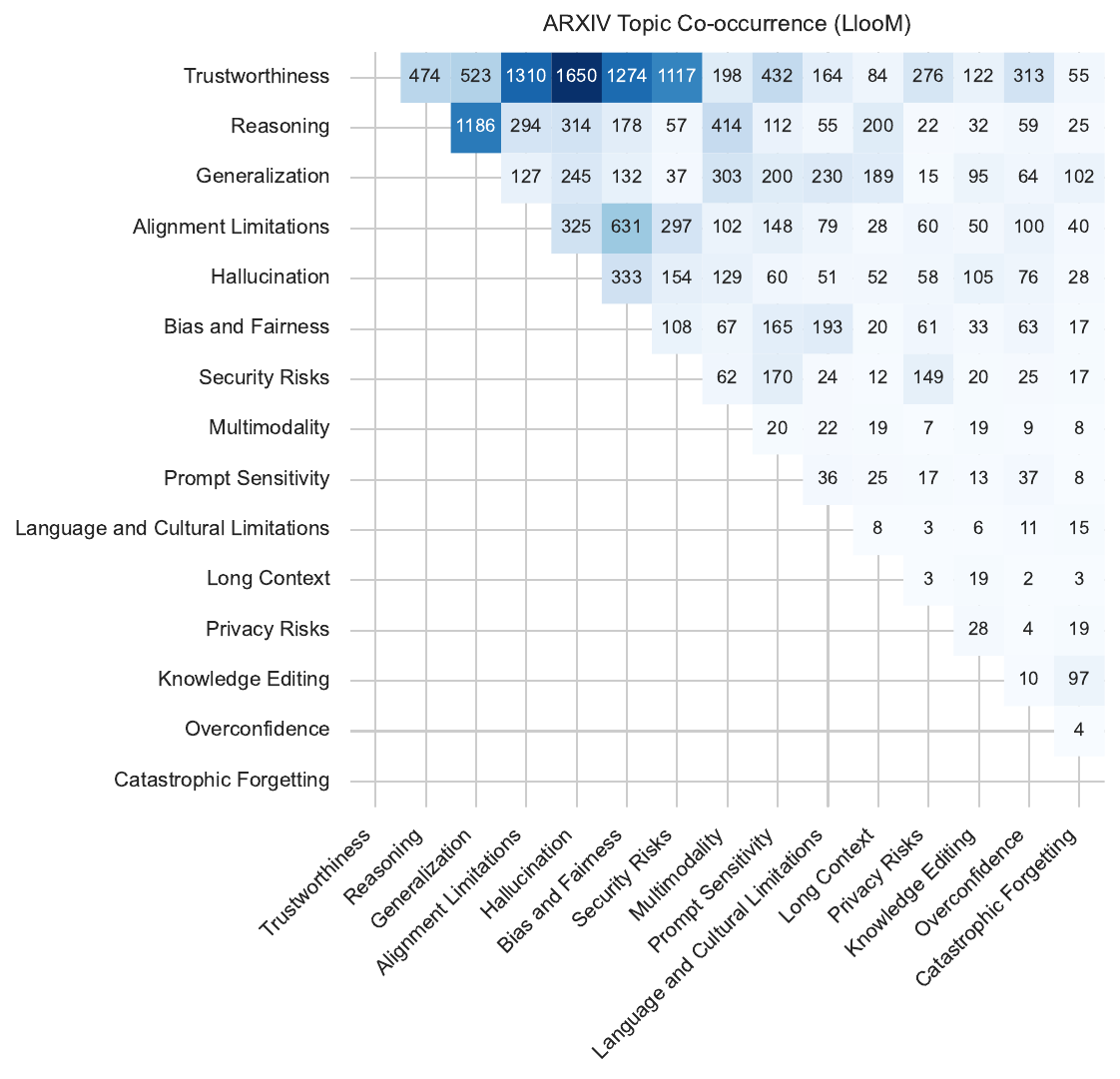}
            \caption{\footnotesize arXiv Topic Co-occurrence Matrix}
            \label{fig:arxiv-cooccurrence-matrix}
        \end{subfigure} 
    \end{tabular}
    }

    \caption{Topic co-occurrence matrices for ACL and arXiv LLM limitation papers, clustered by LlooM approach.}
    \label{fig:topic-cooccurrence}
\end{figure}

In both ACL and arXiv papers, topic co-occurrences tend to concentrate around the largest clusters (see Figure~\ref{fig:topic-cooccurrence}). The main patterns are as follows:

\begin{itemize}

    \item \textit{Reasoning} shows the highest co-occurrence with other limitations, especially in ACL. It frequently overlaps with \textit{Knowledge Editing} (171 ACL), \textit{Hallucination} (153 ACL; 314 arXiv), \textit{Multimodality} (80 ACL; 414 arXiv), and \textit{Generalization} (65 ACL; 1186 arXiv), as well as \textit{Long Context} and \textit{Alignment Limitations}. These links reflect how flawed reasoning can lead to other issues, such as brittle updates in knowledge editing \citep{zhong2023mquake}, incoherent multimodal responses \citep{lu2023mathvista}, and hallucinations \citep{meadows2024exploring}.

    \item \textit{Trustworthiness} covers not only failures across \textit{Alignment Limitations} (1310), \textit{Security Risks} (1117), \textit{Privacy Risks} (276), but also general output reliability, co-occurring with \textit{Reasoning} (1650), and \textit{Hallucination} (117). Trust breaks down in different ways: flawed reasoning yields faulty responses \citep{haji2024improving}, misalignment leads to persuasive but incorrect outputs \citep{wen2409language}, and adversarial prompts can trigger data leaks \citep{rosati2024defending}.

    \item \textit{Generalization} co-occurs with \textit{Reasoning}, \textit{Multimodality}, \textit{Knowledge Editing}, \textit{Prompt Sensitivity}, \textit{Security Risks}, and \textit{Bias and Fairness}, reflecting how it is stress-tested across prompts, tasks, and domains. For instance, models often fail under prompt perturbations \citep{ball2024can}, OOD reasoning \citep{stechly2024chain}, or clinical QA \citep{ness2024medfuzz}.

    \item \textit{Alignment Limitations and Security Risks} cluster with issues focused on output control: \textit{Bias and Fairness} (631), \textit{Prompt Sensitivity} (170), \textit{Privacy Risks} (149), and \textit{Hallucination} (325 with alignment; 154 with security). These links reflect how attempts to control model behavior can introduce new failures: e.g. fine-tuning for alignment can increase hallucination rates \citep{lin2024flame}. In high-stakes domains like medicine, these concerns combine: hallucination, misalignment, and privacy risks all compromise safe deployment of LLMs in clinical settings \citep{wang2024safety}. 
    
    \item \textit{Hallucination} appears with a wide range of issues: \textit{Knowledge Editing}, \textit{Language and Cultural Limitations}, and \textit{Long Context} in ACL; \textit{Bias and Fairness} (333), \textit{Multimodality} (129), and \textit{Generalization} in arXiv, highlighting its role across both technical and sociocultural settings.

    \item \textit{Bias and Fairness} with sociocultural and safety concerns: \textit{Language and Cultural Limitations} (100 ACL; 193 arXiv), \textit{Prompt Sensitivity} (165), \textit{Security Risks} (108), and \textit{Multimodality} (67), forming a distinct subcluster focused on language representation and cultural alignment \citep{watson2023social, felkner2023winoqueer}.
\end{itemize}

These co-occurrence patterns show that many failures are often studied in combination. As a result, even if individual topics are not always the primary focus, they continue to appear in multi-topic work. This overlap likely contributes to the trend stability observed earlier: few topics disappear entirely because they remain relevant through their connections to others.

%% file: bibliography.bib
@article{srivastava2022beyond,
  title={Beyond the imitation game: Quantifying and extrapolating the capabilities of language models},
  author={Srivastava, Aarohi and Rastogi, Abhinav and Rao, Abhishek and Shoeb, Abu Awal Md and Abid, Abubakar and Fisch, Adam and Brown, Adam R and Santoro, Adam and Gupta, Aditya and Garriga-Alonso, Adri{\`a}, et al.},
  journal={arXiv preprint arXiv:2206.04615},
  year={2022}
}

@article{chen2024survey,
  title={A survey on large language models for critical societal domains: Finance, healthcare, and law},
  author={Chen, Zhiyu Zoey and Ma, Jing and Zhang, Xinlu and Hao, Nan and Yan, An and Nourbakhsh, Armineh and Yang, Xianjun and McAuley, Julian and Petzold, Linda and Wang, William Yang},
  journal={arXiv preprint arXiv:2405.01769},
  year={2024}
}

@misc{wei2022emergent,
      title={Emergent Abilities of Large Language Models}, 
      author={Jason Wei and Yi Tay and Rishi Bommasani and Colin Raffel and Barret Zoph and Sebastian Borgeaud and Dani Yogatama and Maarten Bosma and Denny Zhou and Donald Metzler and Ed H. Chi and Tatsunori Hashimoto and Oriol Vinyals and Percy Liang and Jeff Dean and William Fedus},
      journal={arXiv preprint arXiv:2206.07682},
      year={2022},
}

@article{eger2025transforming,
  title={Transforming Science with Large Language Models: A Survey on AI-assisted Scientific Discovery, Experimentation, Content Generation, and Evaluation},
  author={Steffen Eger and Yong Cao and Jennifer D'Souza and Andreas Geiger and Christian Greisinger and Stephanie Gross and Yufang Hou and Brigitte Krenn and Anne Lauscher and Yizhi Li and Chenghua Lin and Nafise Sadat Moosavi and Wei Zhao and Tristan Miller},
  journal={arXiv preprint arXiv:2502.05151},
  year={2025}
}

@article{gupta2024generative,
  author={Gupta, Priyanka and Ding, Bosheng and Guan, Chong and Ding, Ding},
  doi = {https://doi.org/10.1016/j.dim.2024.100066},
  journal={Data and Information Management},
  number={2},
  pages={100066},
  title={Generative AI: A systematic review using topic modelling techniques},
  volume={8},
  year={2024},
}

@article{naveed2023comprehensive,
  title={A comprehensive overview of large language models},
  author={Naveed, Humza and Khan, Asad Ullah and Qiu, Shi and Saqib, Muhammad and Anwar, Saeed and Usman, Muhammad and Akhtar, Naveed and Barnes, Nick and Mian, Ajmal},
  journal={arXiv preprint arXiv:2307.06435},
  year={2023}
}

@article{hadi2023large,
  title={Large language models: a comprehensive survey of its applications, challenges, limitations, and future prospects},
  author={Hadi, Muhammad Usman and Tashi, Qasem Al and Qureshi, Rizwan and Shah, Abbas and Muneer, Amgad and Irfan, Muhammad and Zafar, Anas and Shaikh, Muhammad Bilal and Akhtar, Naveed and Hassan, Syed Zohaib and Shoman, Maged and and Wu, Jia and Mirjalili, Seyedali and Shah, Mubarak},
  doi = {10.36227/techrxiv.23589741.v8},
  journal={Authorea Preprints},
  volume={1},
  pages={1--26},
  year={2023},
}

@article{matarazzo2025survey,
  title={A Survey on Large Language Models with some Insights on their Capabilities and Limitations},
  author={Matarazzo, Andrea and Torlone, Riccardo},
  journal={arXiv preprint arXiv:2501.04040},
  year={2025}
}

@article{wu2024survey,
  title={A survey on large language models for recommendation},
  author={Likang Wu and Zhi Zheng and Zhaopeng Qiu and Hao Wang and Hongchao Gu and Tingjia Shen and Chuan Qin and Chen Zhu and Hengshu Zhu and Qi Liu and Hui Xiong and Enhong Chen},
  doi = {10.1007/s11280-024-01291-2},
  journal={World Wide Web},
  volume={27},
  number={5},
  pages={60},
  year={2024},
}

@article{zhu2023large,
  title={Large language models for information retrieval: A survey},
  author={Zhu, Yutao and Yuan, Huaying and Wang, Shuting and Liu, Jiongnan and Liu, Wenhan and Deng, Chenlong and Chen, Haonan and Liu, Zheng and Dou, Zhicheng and Wen, Ji-Rong},
  journal={arXiv preprint arXiv:2308.07107},
  year={2023}
}

@article{luo2025llm4sr,
  title={LLM4SR: A Survey on Large Language Models for Scientific Research},
  author={Luo, Ziming and Yang, Zonglin and Xu, Zexin and Yang, Wei and Du, Xinya},
  journal={arXiv preprint arXiv:2501.04306},
  year={2025}
}

@article{jin2024large,
  title={Large language models on graphs: A comprehensive survey},
  author={Jin, Bowen and Liu, Gang and Han, Chi and Jiang, Meng and Ji, Heng and Han, Jiawei},
  journal={IEEE Trans. Knowl. Data Eng.},
  year={2024},
  volume={36},
  number={12},
  pages={8622-8642},
  doi={10.1109/TKDE.2024.3469578}
}

@article{pan2024unifying,
  title={Unifying large language models and knowledge graphs: A roadmap},
  author={Pan, Shirui and Luo, Linhao and Wang, Yufei and Chen, Chen and Wang, Jiapu and Wu, Xindong},
  journal={IEEE Trans. Knowl. Data Eng.},
  doi={10.1109/TKDE.2024.3352100},
  volume={36},
  number={7},
  pages={3580--3599},
  year={2024}
}

@article{kim2025limitations,
  title={Limitations of Large Language Models in Clinical Problem-Solving Arising from Inflexible Reasoning},
  author={Kim, Jonathan and Podlasek, Anna and Shidara, Kie and Liu, Feng and Alaa, Ahmed and Bernardo, Danilo},
  journal={arXiv preprint arXiv:2502.04381},
  year={2025}
}

@article{van2024adapted,
  title={Adapted large language models can outperform medical experts in clinical text summarization},
  author={Van Veen, Dave and Van Uden, Cara and Blankemeier, Louis and Delbrouck, Jean-Benoit and Aali, Asad and Bluethgen, Christian and Pareek, Anuj and Polacin, Malgorzata and Reis, Eduardo Pontes and Seehofnerová, Anna and Rohatgi, Nidhi and Hosamani, Poonam and Collins, William and Ahuja, Neera and Langlotz, Curtis P. and Hom, Jason and Gatidis, Sergios and Pauly, John and Chaudhari, Akshay S.},
  journal={Nature medicine},
  doi = {10.1038/s41591-024-02855-5},
  volume={30},
  number={4},
  pages={1134--1142},
  year={2024}
}

@article{pham2024towards,
  title={Towards reliable medical question answering: Techniques and challenges in mitigating hallucinations in language models},
  author={Pham, Duy Khoa and Vo, Bao Quoc},
  journal={arXiv preprint arXiv:2408.13808},
  year={2024}
}

@article{hou2024large,
  title={Large language models for software engineering: A systematic literature review},
  author={Hou, Xinyi and Zhao, Yanjie and Liu, Yue and Yang, Zhou and Wang, Kailong and Li, Li and Luo, Xiapu and Lo, David and Grundy, John and Wang, Haoyu},
  journal={ACM Trans. Softw. Eng. Methodol.},
  doi = {10.1145/3695988},
  volume={33},
  number={8},
  pages={1--79},
  year={2024},
}

@article{shen2024understanding,
  title={Understanding the capabilities and limitations of large language models for cultural commonsense},
  author={Shen, Siqi and Logeswaran, Lajanugen and Lee, Moontae and Lee, Honglak and Poria, Soujanya and Mihalcea, Rada},
  journal={arXiv preprint arXiv:2405.04655},
  year={2024}
}

@article{wang2024survey,
  title={A survey on large language model based autonomous agents},
  author={Wang, Lei and Ma, Chen and Feng, Xueyang and Zhang, Zeyu and Yang, Hao and Zhang, Jingsen and Chen, Zhiyuan and Tang, Jiakai and Chen, Xu and Lin, Yankai and Zhao, Wayne Xin and Wei, Zhewei and Wen, Jirong},
  doi = {10.1007/s11704-024-40231-1},
  journal={Frontiers of Computer Science},
  volume={18},
  number={6},
  pages={186345},
  year={2024},
}

@article{huang2025survey,
  title={A survey on hallucination in large language models: Principles, taxonomy, challenges, and open questions},
  author={Huang, Lei and Yu, Weijiang and Ma, Weitao and Zhong, Weihong and Feng, Zhangyin and Wang, Haotian and Chen, Qianglong and Peng, Weihua and Feng, Xiaocheng and Qin, Bing and Liu, Ting},
  doi = {10.1145/3703155},
  journal={ACM Transactions on Information Systems},
  volume={43},
  number={2},
  pages={1--55},
  year={2025},
}

@article{ji2023survey,
  title={Survey of hallucination in natural language generation},
  author={Ji, Ziwei and Lee, Nayeon and Frieske, Rita and Yu, Tiezheng and Su, Dan and Xu, Yan and Ishii, Etsuko and Bang, Ye Jin and Madotto, Andrea and Fung, Pascale},
  doi = {10.1145/3571730},
  journal={ACM computing surveys},
  volume={55},
  number={12},
  pages={1--38},
  year={2023},
}

@article{tonmoy2024comprehensive,
  title={A comprehensive survey of hallucination mitigation techniques in large language models},
  author={Tonmoy, SM and Zaman, SM and Jain, Vinija and Rani, Anku and Rawte, Vipula and Chadha, Aman and Das, Amitava},
  journal={arXiv preprint arXiv:2401.01313},
  year={2024}
}

@article{liu2024survey,
  title={A survey on hallucination in large vision-language models},
  author={Liu, Hanchao and Xue, Wenyuan and Chen, Yifei and Chen, Dapeng and Zhao, Xiutian and Wang, Ke and Hou, Liping and Li, Rongjun and Peng, Wei},
  journal={arXiv preprint arXiv:2402.00253},
  year={2024}
}

@article{sahoo2024comprehensive,
  title={A comprehensive survey of hallucination in large language, image, video and audio foundation models},
  author={Sahoo, Pranab and Meharia, Prabhash and Ghosh, Akash and Saha, Sriparna and Jain, Vinija and Chadha, Aman},
  journal={arXiv preprint arXiv:2405.09589},
  year={2024}
}

@inproceedings{sahoo-etal-2024-comprehensive,
    title = {A Comprehensive Survey of Hallucination in Large Language, Image, Video and Audio Foundation Models},
    author = {Sahoo, Pranab  and
      Meharia, Prabhash  and
      Ghosh, Akash  and
      Saha, Sriparna  and
      Jain, Vinija  and
      Chadha, Aman},
    booktitle = {Findings of the EMNLP},
    doi = {10.18653/v1/2024.findings-emnlp.685},
    pages = {11709--11724},
}

@article{xu2025towards,
  title={Towards Large Reasoning Models: A Survey of Reinforced Reasoning with Large Language Models},
  author={Fengli Xu and Qianyue Hao and Zefang Zong and Jingwei Wang and Yunke Zhang and Jingyi Wang and Xiaochong Lan and Jiahui Gong and Tianjian Ouyang and Fanjin Meng and Chenyang Shao and Yuwei Yan and Qinglong Yang and Yiwen Song and Sijian Ren and Xinyuan Hu and Yu Li and Jie Feng and Chen Gao and Yong Li},
  journal={arXiv preprint arXiv:2501.09686},
  year={2025}
}

@article{plaat2024reasoning,
  title={Reasoning with large language models, a survey},
  author={Plaat, Aske and Wong, Annie and Verberne, Suzan and Broekens, Joost and van Stein, Niki and Back, Thomas},
  journal={arXiv preprint arXiv:2407.11511},
  year={2024}
}

@article{li2025system,
  title={From system 1 to system 2: A survey of reasoning large language models},
  author={Zhong-Zhi Li and Duzhen Zhang and Ming-Liang Zhang and Jiaxin Zhang and Zengyan Liu and Yuxuan Yao and Haotian Xu and Junhao Zheng and Pei-Jie Wang and Xiuyi Chen and Yingying Zhang and Fei Yin and Jiahua Dong and Zhiwei Li and Bao-Long Bi and Ling-Rui Mei and Junfeng Fang and Xiao Liang and Zhijiang Guo and Le Song and Cheng-Lin Liu},
  journal={arXiv preprint arXiv:2502.17419},
  year={2025}
}

@article{chen2025towards,
  title={Towards reasoning era: A survey of long chain-of-thought for reasoning large language models},
  author={Chen, Qiguang and Qin, Libo and Liu, Jinhao and Peng, Dengyun and Guan, Jiannan and Wang, Peng and Hu, Mengkang and Zhou, Yuhang and Gao, Te and Che, Wangxiang},
  journal={arXiv preprint arXiv:2503.09567},
  year={2025}
}

@article{huang2022towards,
  title={Towards reasoning in large language models: A survey},
  author={Huang, Jie and Chang, Kevin Chen-Chuan},
  journal={arXiv preprint arXiv:2212.10403},
  year={2022}
}

@article{ahn2024large,
  title={Large language models for mathematical reasoning: Progresses and challenges},
  author={Ahn, Janice and Verma, Rishu and Lou, Renze and Liu, Di and Zhang, Rui and Yin, Wenpeng},
  journal={arXiv preprint arXiv:2402.00157},
  year={2024}
}

@article{yao2024survey,
  title={A survey on large language model (llm) security and privacy: The good, the bad, and the ugly},
  author={Yao, Yifan and Duan, Jinhao and Xu, Kaidi and Cai, Yuanfang and Sun, Zhibo and Zhang, Yue},
  journal={High-Confidence Computing},
  doi = {https://doi.org/10.1016/j.hcc.2024.100211},
  pages={100211},
  year={2024},
}

@article{das2025security,
  title={Security and privacy challenges of large language models: A survey},
  author={Das, Badhan Chandra and Amini, M Hadi and Wu, Yanzhao},
  journal={ACM Computing Surveys},
  doi = {10.1145/3712001},
  volume={57},
  number={6},
  pages={1--39},
  year={2025},
}

@article{sun2024trustllm,
  title={Trustllm: Trustworthiness in large language models},
  author={Sun, Lichao and Huang, Yue and Wang, Haoran and Wu, Siyuan and Zhang, Qihui and Gao, Chujie and Huang, Yixin and Lyu, Wenhan and Zhang, Yixuan and others},
  journal={arXiv preprint arXiv:2401.05561},
  year={2024}
}

@article{jiao2024navigating,
  title={Navigating llm ethics: Advancements, challenges, and future directions},
  author={Jiao, Junfeng and Afroogh, Saleh and Xu, Yiming and Phillips, Connor},
  journal={arXiv preprint arXiv:2406.18841},
  year={2024}
}

@article{kumar2024ethics,
  title={The ethics of interaction: Mitigating security threats in llms},
  author={Kumar, Ashutosh and Murthy, Shiv Vignesh and Singh, Sagarika and Ragupathy, Swathy},
  journal={arXiv preprint arXiv:2401.12273},
  year={2024}
}

@article{gan2024navigating,
  title={Navigating the risks: A survey of security, privacy, and ethics threats in llm-based agents},
  author={Yuyou Gan and Yong Yang and Zhe Ma and Ping He and Rui Zeng and Yiming Wang and Qingming Li and Chunyi Zhou and Songze Li and Ting Wang and Yunjun Gao and Yingcai Wu and Shouling Ji},
  journal={arXiv preprint arXiv:2411.09523},
  year={2024}
}

@article{zhang2023clusterllm,
  title={Clusterllm: Large language models as a guide for text clustering},
  author={Zhang, Yuwei and Wang, Zihan and Shang, Jingbo},
  journal={arXiv preprint arXiv:2305.14871},
  year={2023}
}

@inproceedings{lam2024concept,
  title={Concept induction: Analyzing unstructured text with high-level concepts using lloom},
  author={Lam, Michelle S and Teoh, Janice and Landay, James A and Heer, Jeffrey and Bernstein, Michael S},
  booktitle={Proceedings of the 2024 CHI Conference on Human Factors in Computing Systems},
  doi = {10.1145/3613904.3642830},
  pages={1--28},
  year={2024}
}

@article{diaz2025k,
  title={k-LLMmeans: Summaries as Centroids for Interpretable and Scalable LLM-Based Text Clustering},
  author={Diaz-Rodriguez, Jairo},
  journal={arXiv preprint arXiv:2502.09667},
  year={2025}
}

@inproceedings{feng2024llmedgerefine,
  title={LLMEdgeRefine: Enhancing Text Clustering with LLM-Based Boundary Point Refinement},
  author={Feng, Zijin and Lin, Luyang and Wang, Lingzhi and Cheng, Hong and Wong, Kam-Fai},
  booktitle={Proceedings of the EMNLP},
  doi = {10.18653/v1/2024.emnlp-main.1025},
  pages={18455--18462},
  year={2024}
}

@inproceedings{pattnaik2024improving,
  title={Improving Hierarchical Text Clustering with LLM-guided Multi-view Cluster Representation},
  author={Pattnaik, Anup and George, Cijo and Tripathi, Rishabh and Vutla, Sasanka and Vepa, Jithendra},
  booktitle={Proceedings of the EMNLP: Industry Track},
  doi = {10.18653/v1/2024.emnlp-industry.54},
  pages={719--727},
  year={2024}
}

@article{gana2024leveraging,
  title={Leveraging llms for efficient topic reviews},
  author={Gana, Bady and Leiva-Araos, Andr{\'e}s and Allende-Cid, H{\'e}ctor and Garc{\'\i}a, Jos{\'e}},
  journal={Applied Sciences},
  doi = {10.3390/app14177675},
  volume={14},
  number={17},
  pages={7675},
  year={2024},
}

@article{ding2024unraveling,
author={Ding,Qinxu and Ding,Ding and Wang,Yue and Guan,Chong and Ding,Bosheng},
year={2024},
title={Unraveling the landscape of large language models: a systematic review and future perspectives},
journal={Journal of Electronic Business \& Digital Economics},
volume={3},
number={1},
pages={3-19},
}

@inproceedings{arsalan2025mapping,
  title={Mapping Data-Driven Research Impact Science: The Role of Machine Learning and Artificial Intelligence},
  author={Arsalan, Mudassar Hassan and Mubin, Omar and Al Mahmud, Abdullah and Khan, Imran Ahmed and Hassan, Ali Jan},
  booktitle={Metrics},
  volume={2},
  number={2},
  pages={5},
  year={2025},
  doi = {10.3390/metrics2020005}
}

@article{liang2025surveyx,
  title={Surveyx: Academic survey automation via large language models},
  author={Xun Liang and Jiawei Yang and Yezhaohui Wang and Chen Tang and Zifan Zheng and Shichao Song and Zehao Lin and Yebin Yang and Simin Niu and Hanyu Wang and Bo Tang and Feiyu Xiong and Keming Mao and Zhiyu li},
  journal={arXiv preprint arXiv:2502.14776},
  year={2025}
}

@article{wang2024autosurvey,
  title={Autosurvey: Large language models can automatically write surveys},
  author={Yidong Wang and Qi Guo and Wenjin Yao and Hongbo Zhang and Xin Zhang and Zhen Wu and Meishan Zhang and Xinyu Dai and Min Zhang and Qingsong Wen and Wei Ye and Shikun Zhang and Yue Zhang},
  journal={Advances in Neural Information Processing Systems},
  volume={37},
  pages={115119--115145},
  year={2024}
}

@article{he2025pasa,
  title={PaSa: An LLM Agent for Comprehensive Academic Paper Search},
  author={Yichen He and Guanhua Huang and Peiyuan Feng and Yuan Lin and Yuchen Zhang and Hang Li and Weinan E},
  journal={arXiv preprint arXiv:2501.10120},
  year={2025}
}

@article{agarwal2024litllm,
  title={Litllm: A toolkit for scientific literature review},
  author={Agarwal, Shubham and Sahu, Gaurav and Puri, Abhay and Laradji, Issam H and Dvijotham, Krishnamurthy DJ and Stanley, Jason and Charlin, Laurent and Pal, Christopher},
  journal={arXiv preprint arXiv:2402.01788},
  year={2024}
}

@article{zhao2023survey,
  title={A survey of large language models},
  author={Wayne Xin Zhao and Kun Zhou and Junyi Li and Tianyi Tang and Xiaolei Wang and Yupeng Hou and Yingqian Min and Beichen Zhang and Junjie Zhang and Zican Dong and Yifan Du and Chen Yang and Yushuo Chen and Zhipeng Chen and Jinhao Jiang and Ruiyang Ren and Yifan Li and Xinyu Tang and Zikang Liu and Peiyu Liu and Jian-Yun Nie and Ji-Rong Wen},
  journal={arXiv preprint arXiv:2303.18223},
  volume={1},
  number={2},
  year={2023}
}

@article{martinc2022tnt,
  title={TNT-KID: Transformer-based neural tagger for keyword identification},
  author={Martinc, Matej and {\v{S}}krlj, Bla{\v{z}} and Pollak, Senja},
  journal={Natural Language Engineering},
  volume={28},
  number={4},
  pages={409--448},
  year={2022},
  publisher={Cambridge University Press}
}

@article{giarelis2024deep,
  title={Deep learning and embeddings-based approaches for keyphrase extraction: a literature review},
  author={Giarelis, Nikolaos and Karacapilidis, Nikos},
  journal={Knowledge and Information Systems},
  volume={66},
  number={11},
  pages={6493--6526},
  year={2024},
  publisher={Springer}
}

@inproceedings{xu2025limitgen,
    title = "Can {LLM}s Identify Critical Limitations within Scientific Research? A Systematic Evaluation on {AI} Research Papers",
    author = "Xu, Zhijian  and
      Zhao, Yilun  and
      Patwardhan, Manasi  and
      Vig, Lovekesh  and
      Cohan, Arman",
    editor = "Che, Wanxiang  and
      Nabende, Joyce  and
      Shutova, Ekaterina  and
      Pilehvar, Mohammad Taher",
    booktitle = "Proceedings of the 63rd Annual Meeting of the Association for Computational Linguistics (Volume 1: Long Papers)",
    month = jul,
    year = "2025",
    address = "Vienna, Austria",
    publisher = "Association for Computational Linguistics",
    url = "https://aclanthology.org/2025.acl-long.1009/",
    doi = "10.18653/v1/2025.acl-long.1009",
    pages = "20652--20706",
    ISBN = "979-8-89176-251-0",
}

@inproceedings{azher2024limtopic,
  title={Limtopic: Llm-based topic modeling and text summarization for analyzing scientific articles limitations},
  author={Azher, Ibrahim Al and Seethi, Venkata Devesh Reddy and Akella, Akhil Pandey and Alhoori, Hamed},
  booktitle={Proceedings of the 24th ACM/IEEE Joint Conference on Digital Libraries},
  pages={1--12},
  year={2024}
}

@article{al2025bagels,
  title={BAGELS: Benchmarking the Automated Generation and Extraction of Limitations from Scholarly Text},
  author={Al Azher, Ibrahim and Mokarrama, Miftahul Jannat and Guo, Zhishuai and Choudhury, Sagnik Ray and Alhoori, Hamed},
  journal={arXiv preprint arXiv:2505.18207},
  year={2025}
}

@article{ainslie2023gqa,
  title={Gqa: Training generalized multi-query transformer models from multi-head checkpoints},
  author={Ainslie, Joshua and Lee-Thorp, James and De Jong, Michiel and Zemlyanskiy, Yury and Lebr{\'o}n, Federico and Sanghai, Sumit},
  journal={arXiv preprint arXiv:2305.13245},
  year={2023}
}

@article{mohtashami2023social,
  title={Social learning: Towards collaborative learning with large language models},
  author={Mohtashami, Amirkeivan and Hartmann, Florian and Gooding, Sian and Zilka, Lukas and Sharifi, Matt},
  journal={arXiv preprint arXiv:2312.11441},
  year={2023}
}

@article{deng2024gotcha,
  title={Gotcha! Don't trick me with unanswerable questions! Self-aligning Large Language Models for Responding to Unknown Questions},
  author={Deng, Yang and Zhao, Yong and Li, Moxin and Ng, See-Kiong and Chua, Tat-Seng},
  journal={arXiv e-prints},
  pages={arXiv--2402},
  year={2024}
}

@article{li2023counterfactual,
  title={Counterfactual reasoning: Testing language models' understanding of hypothetical scenarios},
  author={Li, Jiaxuan and Yu, Lang and Ettinger, Allyson},
  journal={arXiv preprint arXiv:2305.16572},
  year={2023}
}

@article{thomm2024limits,
  title={Limits of transformer language models on learning to compose algorithms},
  author={Thomm, Jonathan and Camposampiero, Giacomo and Terzic, Aleksandar and Hersche, Michael and Sch{\"o}lkopf, Bernhard and Rahimi, Abbas},
  journal={Advances in Neural Information Processing Systems},
  volume={37},
  pages={7631--7674},
  year={2024}
}

@article{deyoung2019eraser,
  title={ERASER: A benchmark to evaluate rationalized NLP models},
  author={DeYoung, Jay and Jain, Sarthak and Rajani, Nazneen Fatema and Lehman, Eric and Xiong, Caiming and Socher, Richard and Wallace, Byron C},
  journal={arXiv preprint arXiv:1911.03429},
  year={2019}
}

@article{reimers2019sentence,
  title={Sentence-bert: Sentence embeddings using siamese bert-networks},
  author={Reimers, Nils and Gurevych, Iryna},
  journal={arXiv preprint arXiv:1908.10084},
  year={2019}
}

@article{chawla2002smote,
  title={SMOTE: synthetic minority over-sampling technique},
  author={Chawla, Nitesh V and Bowyer, Kevin W and Hall, Lawrence O and Kegelmeyer, W Philip},
  journal={Journal of artificial intelligence research},
  volume={16},
  pages={321--357},
  year={2002}
}

@article{grattafiori2024llama,
  title={The llama 3 herd of models},
  author={Grattafiori, Aaron and Dubey, Abhimanyu and Jauhri, Abhinav and Pandey, Abhinav and Kadian, Abhishek and Al-Dahle, Ahmad and Letman, Aiesha and Mathur, Akhil and Schelten, Alan and Vaughan, Alex and others},
  journal={arXiv preprint arXiv:2407.21783},
  year={2024}
}

@misc{jiang2023mistral7b,
      title={Mistral 7B}, 
      author={Albert Q. Jiang and Alexandre Sablayrolles and Arthur Mensch and Chris Bamford and Devendra Singh Chaplot and Diego de las Casas and Florian Bressand and Gianna Lengyel and Guillaume Lample and Lucile Saulnier and Lélio Renard Lavaud and Marie-Anne Lachaux and Pierre Stock and Teven Le Scao and Thibaut Lavril and Thomas Wang and Timothée Lacroix and William El Sayed},
      year={2023},
      eprint={2310.06825},
      archivePrefix={arXiv},
      primaryClass={cs.CL},
      url={https://arxiv.org/abs/2310.06825}, 
}

@inproceedings{rother2020cmce,
  title={CMCE at SemEval-2020 task 1: Clustering on manifolds of contextualized embeddings to detect historical meaning shifts},
  author={Rother, David and Haider, Thomas and Eger, Steffen},
  booktitle={14th International Workshop on Semantic Evaluation},
  pages={187--193},
  year={2020}
}

@article{viswanathan2024large,
  title={Large language models enable few-shot clustering},
  author={Viswanathan, Vijay and Gashteovski, Kiril and Gashteovski, Kiril and Lawrence, Carolin and Wu, Tongshuang and Neubig, Graham},
  journal={TACL},
  volume={12},
  pages={321--333},
  year={2024},
  publisher={MIT Press One Broadway, 12th Floor, Cambridge, Massachusetts 02142, USA~…}
}

@article{shaikh2022second,
  title={On second thought, let's not think step by step! bias and toxicity in zero-shot reasoning},
  author={Shaikh, Omar and Zhang, Hongxin and Held, William and Bernstein, Michael and Yang, Diyi},
  journal={arXiv preprint arXiv:2212.08061},
  year={2022}
}

@article{mcinnes2018umap,
  title={Umap: Uniform manifold approximation and projection for dimension reduction},
  author={McInnes, Leland and Healy, John and Melville, James},
  journal={arXiv preprint arXiv:1802.03426},
  year={2018}
}

@article{mcinnes2017hdbscan,
  title={hdbscan: Hierarchical density based clustering.},
  author={McInnes, Leland and Healy, John and Astels, Steve},
  journal={J. Open Source Softw.},
  volume={2},
  number={11},
  pages={205},
  year={2017}
}

@article{grootendorst2022bertopic,
  title={BERTopic: Neural topic modeling with a class-based TF-IDF procedure},
  author={Grootendorst, Maarten},
  journal={arXiv preprint arXiv:2203.05794},
  year={2022}
}

@article{kendall1948rank,
  title={Rank correlation methods.},
  author={Kendall, Maurice George},
  year={1948},
  publisher={Griffin}
}

@article{jiang2024unlocking,
  title={Unlocking adversarial suffix optimization without affirmative phrases: Efficient black-box jailbreaking via llm as optimizer},
  author={Jiang, Weipeng and Wang, Zhenting and Zhai, Juan and Ma, Shiqing and Zhao, Zhengyu and Shen, Chao},
  journal={arXiv preprint arXiv:2408.11313},
  year={2024}
}

@article{wu2023can,
  title={Can gpt-4v (ision) serve medical applications? case studies on gpt-4v for multimodal medical diagnosis},
  author={Chaoyi Wu and Jiayu Lei and Qiaoyu Zheng and Weike Zhao and Weixiong Lin and Xiaoman Zhang and Xiao Zhou and Ziheng Zhao and Ya Zhang and Yanfeng Wang and Weidi Xie},
  journal={arXiv preprint arXiv:2310.09909},
  year={2023}
}

@article{levinstein2024still,
  title={Still no lie detector for language models: Probing empirical and conceptual roadblocks},
  author={Levinstein, Benjamin A and Herrmann, Daniel A},
  journal={Philosophical Studies},
  pages={1--27},
  year={2024},
  publisher={Springer}
}

@article{mann1945nonparametric,
  title={Nonparametric tests against trend},
  author={Mann, Henry B},
  journal={Econometrica: Journal of the econometric society},
  pages={245--259},
  year={1945},
  publisher={JSTOR}
}

@article{achiam2023gpt,
  title={Gpt-4 technical report},
  author={Achiam, Josh and Adler, Steven and Agarwal, Sandhini and Ahmad, Lama and Akkaya, Ilge and Aleman, Florencia Leoni and Almeida, Diogo and Altenschmidt, Janko and Altman, Sam and others},
  journal={arXiv preprint arXiv:2303.08774},
  year={2023}
}

@inproceedings{li2024retrieval,
  title={Retrieval augmented generation or long-context llms? a comprehensive study and hybrid approach},
  author={Li, Zhuowan and Li, Cheng and Zhang, Mingyang and Mei, Qiaozhu and Bendersky, Michael},
  booktitle={Proceedings of the EMNLP: Industry Track},
  pages={881--893},
  year={2024}
}

@article{liu2023visual,
  title={Visual instruction tuning},
  author={Liu, Haotian and Li, Chunyuan and Wu, Qingyang and Lee, Yong Jae},
  journal={Advances in neural information processing systems},
  volume={36},
  pages={34892--34916},
  year={2023}
}

@article{lv2024coarse,
  title={Coarse-to-fine highlighting: Reducing knowledge hallucination in large language models},
  author={Lv, Qitan and Wang, Jie and Chen, Hanzhu and Li, Bin and Zhang, Yongdong and Wu, Feng},
  journal={arXiv preprint arXiv:2410.15116},
  year={2024}
}

@article{shi2024red,
  title={Red teaming language model detectors with language models},
  author={Shi, Zhouxing and Wang, Yihan and Yin, Fan and Chen, Xiangning and Chang, Kai-Wei and Hsieh, Cho-Jui},
  journal={TACL},
  volume={12},
  pages={174--189},
  year={2024},
  publisher={MIT Press One Broadway, 12th Floor, Cambridge, Massachusetts 02142, USA~…}
}

@article{lin2022gendered,
  title={Gendered mental health stigma in masked language models},
  author={Lin, Inna Wanyin and Njoo, Lucille and Field, Anjalie and Sharma, Ashish and Reinecke, Katharina and Althoff, Tim and Tsvetkov, Yulia},
  journal={arXiv preprint arXiv:2210.15144},
  year={2022}
}

@article{li2023compressing,
  title={Compressing context to enhance inference efficiency of large language models},
  author={Li, Yucheng and Dong, Bo and Lin, Chenghua and Guerin, Frank},
  journal={arXiv preprint arXiv:2310.06201},
  year={2023}
}

@article{zhang2023study,
  title={A study on the calibration of in-context learning},
  author={Zhang, Hanlin and Zhang, Yi-Fan and Yu, Yaodong and Madeka, Dhruv and Foster, Dean and Xing, Eric and Lakkaraju, Himabindu and Kakade, Sham},
  journal={arXiv preprint arXiv:2312.04021},
  year={2023}
}

@article{ozdayi2023controlling,
  title={Controlling the extraction of memorized data from large language models via prompt-tuning},
  author={Ozdayi, Mustafa Safa and Peris, Charith and FitzGerald, Jack and Dupuy, Christophe and Majmudar, Jimit and Khan, Haidar and Parikh, Rahil and Gupta, Rahul},
  journal={arXiv preprint arXiv:2305.11759},
  year={2023}
}

@article{yang2022seqzero,
  title={SEQZERO: Few-shot compositional semantic parsing with sequential prompts and zero-shot models},
  author={Yang, Jingfeng and Jiang, Haoming and Yin, Qingyu and Zhang, Danqing and Yin, Bing and Yang, Diyi},
  journal={arXiv preprint arXiv:2205.07381},
  year={2022}
}

@article{li2023open,
  title={An open source data contamination report for large language models},
  author={Li, Yucheng and Guerin, Frank and Lin, Chenghua},
  journal={arXiv preprint arXiv:2310.17589},
  year={2023}
}

@article{kristensen2023chatbots,
  title={Chatbots are not reliable text annotators},
  author={Kristensen-McLachlan, Ross Deans and Canavan, Miceal and Kardos, M{\'a}rton and Jacobsen, Mia and Aar{\o}e, Lene},
  journal={arXiv preprint arXiv:2311.05769},
  year={2023}
}

@inproceedings{ouyang2022training,
  title={Training language models to follow instructions with human feedback},
  author={Long Ouyang and Jeff Wu and Xu Jiang and Diogo Almeida and Carroll L. Wainwright and Pamela Mishkin and Chong Zhang and Sandhini Agarwal and Katarina Slama and Alex Ray and John Schulman and Jacob Hilton and Fraser Kelton and Luke Miller and Maddie Simens and Amanda Askell and Peter Welinder and Paul Christiano and Jan Leike and Ryan Lowe},
  booktitle={Proceedings of the NeurIPS},
  pages={27730--27744},
  year={2022}
}

@article{kumar2023gen,
  title={Gen-Z: Generative Zero-Shot Text Classification with Contextualized Label Descriptions},
  author={Kumar, Sachin and Park, Chan Young and Tsvetkov, Yulia},
  journal={arXiv preprint arXiv:2311.07115},
  year={2023}
}

@article{lin2025codereviewqa,
  title={CodeReviewQA: The Code Review Comprehension Assessment for Large Language Models},
  author={Lin, Hong Yi and Liu, Chunhua and Gao, Haoyu and Thongtanunam, Patanamon and Treude, Christoph},
  journal={arXiv preprint arXiv:2503.16167},
  year={2025}
}

@article{naik2025empirical,
  title={An Empirical Study of the Role of Incompleteness and Ambiguity in Interactions with Large Language Models},
  author={Naik, Riya and Srinivasan, Ashwin and He, Estrid and Agarwal, Swati},
  journal={arXiv preprint arXiv:2503.17936},
  year={2025}
}

@article{nicholas2023lost,
  title={Lost in translation: Large language models in non-English content analysis},
  author={Nicholas, Gabriel and Bhatia, Aliya},
  journal={arXiv preprint arXiv:2306.07377},
  year={2023}
}

@inproceedings{long2023adapt,
  title={Adapt in contexts: Retrieval-augmented domain adaptation via in-context learning},
  author={Long, Quanyu and Wang, Wenya and Pan, Sinno},
  booktitle={Proceedings of the 2023 conference on empirical methods in natural language processing},
  pages={6525--6542},
  year={2023}
}

@article{talebi2023beyond,
  title={Beyond the hype: Assessing the performance, trustworthiness, and clinical suitability of gpt3. 5},
  author={Talebi, Salmonn and Tong, Elizabeth and Mofrad, Mohammad RK},
  journal={arXiv preprint arXiv:2306.15887},
  year={2023}
}

@article{zhang2023dissecting,
  title={Dissecting the runtime performance of the training, fine-tuning, and inference of large language models},
  author={Longteng Zhang and Xiang Liu and Zeyu Li and Xinglin Pan and Peijie Dong and Ruibo Fan and Rui Guo and Xin Wang and Qiong Luo and Shaohuai Shi and Xiaowen Chu},
  journal={arXiv preprint arXiv:2311.03687},
  year={2023}
}

@article{fu2024does,
  title={Does Data Contamination Detection Work (Well) for LLMs? A Survey and Evaluation on Detection Assumptions},
  author={Fu, Yujuan and Uzuner, Ozlem and Yetisgen, Meliha and Xia, Fei},
  journal={arXiv preprint arXiv:2410.18966},
  year={2024}
}

@article{wu2023zeroquant,
  title={Zeroquant (4+ 2): Redefining llms quantization with a new fp6-centric strategy for diverse generative tasks},
  author={Xiaoxia Wu and Haojun Xia and Stephen Youn and Zhen Zheng and Shiyang Chen and Arash Bakhtiari and Michael Wyatt and Reza Yazdani Aminabadi and Yuxiong He and Olatunji Ruwase and Leon Song and Zhewei Yao},
  journal={arXiv preprint arXiv:2312.08583},
  year={2023}
}

@article{li2024should,
  title={Should we really edit language models? on the evaluation of edited language models},
  author={Li, Qi and Liu, Xiang and Tang, Zhenheng and Dong, Peijie and Li, Zeyu and Pan, Xinglin and Chu, Xiaowen},
  journal={Advances in Neural Information Processing Systems},
  volume={37},
  pages={30850--30885},
  year={2024}
}

@article{yosef2023irfl,
  title={Irfl: Image recognition of figurative language},
  author={Yosef, Ron and Bitton, Yonatan and Shahaf, Dafna},
  journal={arXiv preprint arXiv:2303.15445},
  year={2023}
}

@article{zhong2023mquake,
  title={Mquake: Assessing knowledge editing in language models via multi-hop questions},
  author={Zhong, Zexuan and Wu, Zhengxuan and Manning, Christopher D and Potts, Christopher and Chen, Danqi},
  journal={arXiv preprint arXiv:2305.14795},
  year={2023}
}

@article{meadows2024exploring,
  title={Exploring the Limits of Fine-grained LLM-based Physics Inference via Premise Removal Interventions},
  author={Meadows, Jordan and James, Tamsin and Freitas, Andre},
  journal={arXiv preprint arXiv:2404.18384},
  year={2024}
}

@article{lu2023mathvista,
  title={Mathvista: Evaluating mathematical reasoning of foundation models in visual contexts},
  author={Lu, Pan and Bansal, Hritik and Xia, Tony and Liu, Jiacheng and Li, Chunyuan and Hajishirzi, Hannaneh and Cheng, Hao and Chang, Kai-Wei and Galley, Michel and Gao, Jianfeng},
  journal={arXiv preprint arXiv:2310.02255},
  year={2023}
}

@article{park2024picturing,
  title={Picturing Ambiguity: A Visual Twist on the Winograd Schema Challenge},
  author={Park, Brendan and Janecek, Madeline and Ezzati-Jivan, Naser and Li, Yifeng and Emami, Ali},
  journal={arXiv preprint arXiv:2405.16277},
  year={2024}
}

@article{haji2024improving,
  title={Improving LLM Reasoning with Multi-Agent Tree-of-Thought Validator Agent},
  author={Haji, Fatemeh and Bethany, Mazal and Tabar, Maryam and Chiang, Jason and Rios, Anthony and Najafirad, Peyman},
  journal={arXiv preprint arXiv:2409.11527},
  year={2024}
}

@article{wen2409language,
  title={Language Models Learn to Mislead Humans via RLHF},
  author={Wen, Jiaxin and Zhong, Ruiqi and Khan, Akbir and Perez, Ethan and Steinhardt, Jacob and Huang, Minlie and Boman, Samuel R and He, He and Feng, Shi},
  year={2024},
  journal={arXiv preprint arXiv:2409.12822}
}

@article{chaudhry2024finetuning,
  title={Finetuning language models to emit linguistic expressions of uncertainty},
  author={Chaudhry, Arslan and Thiagarajan, Sridhar and Gorur, Dilan},
  journal={arXiv preprint arXiv:2409.12180},
  year={2024}
}

@article{rosati2024defending,
  title={Defending against reverse preference attacks is difficult},
  author={Rosati, Domenic and Edkins, Giles and Raj, Harsh and Atanasov, David and Majumdar, Subhabrata and Rajendran, Janarthanan and Rudzicz, Frank and Sajjad, Hassan},
  journal={arXiv preprint arXiv:2409.12914},
  year={2024}
}

@article{ball2024can,
  title={Can we count on llms? the fixed-effect fallacy and claims of gpt-4 capabilities},
  author={Ball, Thomas and Chen, Shuo and Herley, Cormac},
  journal={arXiv preprint arXiv:2409.07638},
  year={2024}
}

@inproceedings{stechly2024chain,
  author = {Stechly, Kaya and Valmeekam, Karthik and Kambhampati, Subbarao},
  title = {Chain of Thoughtlessness? An Analysis of CoT in Planning},
  booktitle = {Proceedings of the NeurIPS},
  year = {2024},
}

@article{ness2024medfuzz,
  title={Medfuzz: Exploring the robustness of large language models in medical question answering},
  author={Ness, Robert Osazuwa and Matton, Katie and Helm, Hayden and Zhang, Sheng and Bajwa, Junaid and Priebe, Carey E and Horvitz, Eric},
  journal={arXiv preprint arXiv:2406.06573},
  year={2024}
}

@article{chowdhery2023palm,
  title={Palm: Scaling language modeling with pathways},
  author={Chowdhery, Aakanksha and Narang, Sharan and Devlin, Jacob and Bosma, Maarten and Mishra, Gaurav and Roberts, Adam and Barham, Paul and Chung, Hyung Won and Sutton, Charles and Gehrmann, Sebastian and others},
  journal={Journal of Machine Learning Research},
  doi = {10.5555/3648699.3648939},
  volume={24},
  number={240},
  pages={1--113},
  year={2023}
}

@article{touvron2023llama,
  title={Llama: Open and efficient foundation language models},
  author = {Touvron, Hugo and Lavril, Thibaut and Izacard, Gautier and Martinet, Xavier and Lachaux, Marie-Anne and Lacroix, Timoth{\'e}e and Rozi{\`e}re, Baptiste and others},
  journal={arXiv preprint arXiv:2302.13971},
  year={2023}
}

@article{shen2023large,
  title={Large language model alignment: A survey},
  author={Shen, Tianhao and Jin, Renren and Huang, Yufei and Liu, Chuang and Dong, Weilong and Guo, Zishan and Wu, Xinwei and Liu, Yan and Xiong, Deyi},
  journal={arXiv preprint arXiv:2309.15025},
  year={2023}
}

@inproceedings{lin2024flame,
  title={Flame: Factuality-aware alignment for large language models},
  author={Lin, Sheng-Chieh and Gao, Luyu and Oguz, Barlas and Xiong, Wenhan and Lin, Jimmy and Yih, Scott and Chen, Xilun},
  booktitle = {Proceedings of the NeurIPS},
  volume={37},
  pages={115588--115614},
  year={2024}
}

@article{wang2024safety,
  title={Safety challenges of AI in medicine},
  author={Xiaoye Wang and Nicole Xi Zhang and Hongyu He and Trang Nguyen and Kun-Hsing Yu and Hao Deng and Cynthia Brandt and Danielle S. Bitterman and Ling Pan and Ching-Yu Cheng and James Zou and Dianbo Liu},
  journal={arXiv preprint arXiv:2409.18968},
  year={2024}
}

@article{felkner2023winoqueer,
  title={Winoqueer: A community-in-the-loop benchmark for anti-lgbtq+ bias in large language models},
  author={Felkner, Virginia K and Chang, Ho-Chun Herbert and Jang, Eugene and May, Jonathan},
  journal={arXiv preprint arXiv:2306.15087},
  year={2023}
}

@inproceedings{watson2023social,
  title={What social attitudes about gender does BERT encode? Leveraging insights from psycholinguistics},
  author={Watson, Julia and Beekhuizen, Barend and Stevenson, Suzanne},
  booktitle={Proceedings of ACL (Volume 1: Long Papers)},
  doi={10.18653/v1/2023.acl-long.375},
  pages={6790--6809},
  year={2023}
}

@article{xu2024hallucination,
  title={Hallucination is inevitable: An innate limitation of large language models},
  author={Xu, Ziwei and Jain, Sanjay and Kankanhalli, Mohan},
  journal={arXiv preprint arXiv:2401.11817},
  year={2024}
}

@article{banerjee2024llms,
  title={Llms will always hallucinate, and we need to live with this},
  author={Banerjee, Sourav and Agarwal, Ayushi and Singla, Saloni},
  journal={arXiv preprint arXiv:2409.05746},
  year={2024}
}

@inproceedings{marchal2022establishing,
  title={Establishing annotation quality in multi-label annotations},
  author={Marchal, Marian and Scholman, Merel and Yung, Frances and Demberg, Vera},
  booktitle={Proceedings of the 29th international conference on computational linguistics},
  pages={3659--3668},
  year={2022}
}

@article{rogers2021primer,
  title={A primer in BERTology: What we know about how BERT works},
  author={Rogers, Anna and Kovaleva, Olga and Rumshisky, Anna},
  doi = {10.1162/tacl_a_00349},
  journal={TACL},
  volume={8},
  pages={842--866},
  year={2021},
}

@article{guo2023evaluating,
  title={Evaluating large language models: A comprehensive survey},
  author={Zishan Guo and Renren Jin and Chuang Liu and Yufei Huang and Dan Shi and Supryadi and Linhao Yu and Yan Liu and Jiaxuan Li and Bojian Xiong and Deyi Xiong},
  journal={arXiv preprint arXiv:2310.19736},
  year={2023}
}

@article{qi2023fine,
  title={Fine-tuning aligned language models compromises safety, even when users do not intend to!},
  author={Qi, Xiangyu and Zeng, Yi and Xie, Tinghao and Chen, Pin-Yu and Jia, Ruoxi and Mittal, Prateek and Henderson, Peter},
  journal={arXiv preprint arXiv:2310.03693},
  year={2023}
}

@article{tan2025equilibrate,
  title={Equilibrate RLHF: Towards Balancing Helpfulness-Safety Trade-off in Large Language Models},
  author={Tan, Yingshui and Jiang, Yilei and Li, Yanshi and Liu, Jiaheng and Bu, Xingyuan and Su, Wenbo and Yue, Xiangyu and Zhu, Xiaoyong and Zheng, Bo},
  journal={arXiv preprint arXiv:2502.11555},
  year={2025}
}

@article{wang2024comprehensive,
  title={A comprehensive survey of LLM alignment techniques: RLHF, RLAIF, PPO, DPO and more},
  author={Zhichao Wang and Bin Bi and Shiva Kumar Pentyala and Kiran Ramnath and Sougata Chaudhuri and Shubham Mehrotra and Zixu and Zhu and Xiang-Bo Mao and Sitaram Asur and Na and Cheng},
  journal={arXiv preprint arXiv:2407.16216},
  year={2024}
}

@article{zhang2024mm,
  title={Mm-llms: Recent advances in multimodal large language models},
  author={Zhang, Duzhen and Yu, Yahan and Dong, Jiahua and Li, Chenxing and Su, Dan and Chu, Chenhui and Yu, Dong},
  journal={arXiv preprint arXiv:2401.13601},
  year={2024}
}

@article{leiter2024nllg,
  title={NLLG Quarterly arXiv Report 09/24: What are the most influential current AI Papers?},
  author={Leiter, Christoph and Belouadi, Jonas and Chen, Yanran and Zhang, Ran and Larionov, Daniil and Kostikova, Aida and Eger, Steffen},
  journal={arXiv preprint arXiv:2412.12121},
  year={2024}
}

@article{movva2023topics,
  title={Topics, authors, and institutions in Large Language Model research: trends from 17K arXiv papers},
  author={Movva, Rajiv and Balachandar, Sidhika and Peng, Kenny and Agostini, Gabriel and Garg, Nikhil and Pierson, Emma},
  journal={arXiv preprint arXiv:2307.10700},
  year={2023}
}

@article{chang2024survey,
  title={A survey on evaluation of large language models},
  author={Yupeng Chang and Xu Wang and Jindong Wang and Yuan Wu and Linyi Yang and Kaijie Zhu and Hao Chen and Xiaoyuan Yi and Cunxiang Wang and Yidong Wang and Wei Ye and Yue Zhang and Yi Chang and Philip S. Yu and Qiang Yang and Xing Xie},
  journal={ACM TIST},
  volume={15},
  number={3},
  pages={1--45},
  year={2024},
  doi={10.1145/3641289}
}

@misc{chiang2024chatbotarena,
      title={Chatbot Arena: An Open Platform for Evaluating LLMs by Human Preference}, 
      author={Wei-Lin Chiang and Lianmin Zheng and Ying Sheng and Anastasios Nikolas Angelopoulos and Tianle Li and Dacheng Li and Hao Zhang and Banghua Zhu and Michael Jordan and Joseph E. Gonzalez and Ion Stoica},
      journal={arXiv preprint arXiv:2403.04132},
      year={2024},
}
